\pdfoutput=1
\documentclass{article}

\PassOptionsToPackage{numbers,compress}{natbib}

\usepackage[main,final]{neurips_2026}

\usepackage[utf8]{inputenc} %
\usepackage[T1]{fontenc} %
\usepackage{hyperref} %
\usepackage{url} %
\usepackage{booktabs} %
\usepackage{amsfonts} %
\usepackage{nicefrac} %
\usepackage{microtype} %
\usepackage[table]{xcolor} %
\usepackage{fontawesome5}
\usepackage{amsmath, amssymb, bm}
\usepackage{mathtools}
\usepackage{graphicx}
\usepackage{subcaption}
\usepackage{tikz}
\usepackage{pgfplots}
\pgfplotsset{compat=1.18}
\usetikzlibrary{positioning,arrows.meta,calc}
\usepackage{multirow}
\usepackage{enumitem}
\usepackage{float}
\usepackage{placeins}
\usepackage[most]{tcolorbox}
\usepackage{lipsum}
\usepackage{algorithm}
\usepackage{algpseudocode}
\usepackage{amsthm}
\usepackage[nameinlink]{cleveref}
\usepackage{array}
\usepackage{pgfplots}
\pgfplotsset{compat=1.18}
\hypersetup{
    colorlinks=true,
    linkcolor=blue,
    citecolor=blue,
    urlcolor=blue
}
\definecolor{githubpink}{HTML}{D63384}
\crefname{equation}{Eq.}{Eqs.}
\Crefname{equation}{Eq.}{Eqs.}
\crefname{figure}{Fig.}{Figs.}
\Crefname{figure}{Fig.}{Figs.}
\crefname{table}{Table}{Tables}
\Crefname{table}{Table}{Tables}
\crefname{algorithm}{Algorithm}{Algorithms}
\Crefname{algorithm}{Algorithm}{Algorithms}
\crefname{section}{Section}{Sections}
\Crefname{section}{Section}{Sections}
\crefname{subsection}{Section}{Sections}
\Crefname{subsection}{Section}{Sections}

\newtcolorbox{theorybox}[1][]{
    enhanced,
    breakable,
    colback=blue!4,
    colframe=blue!55!black,
    boxrule=0.7pt,
    arc=2.5mm,
    left=1.2mm,
    right=1.2mm,
    top=1mm,
    bottom=1mm,
    before skip=8pt,
    after skip=8pt,
    #1
}

\definecolor{findingfill}{HTML}{F9ECF7}
\definecolor{findingborder}{HTML}{716672}
\newsavebox{\qfsavebox}
\newenvironment{qfcard}[1]{%
  \par\smallskip\noindent\begin{lrbox}{\qfsavebox}%
  \begin{minipage}{\dimexpr\linewidth-20pt\relax}%
  \textbf{#1}\enspace\ignorespaces
}{%
  \end{minipage}\end{lrbox}%
  \noindent\begin{tikzpicture}
  \node[fill=findingfill,draw=findingborder,line width=.5pt,
        rounded corners=2pt,inner xsep=9pt,inner ysep=5pt]{\usebox{\qfsavebox}};
  \end{tikzpicture}\par\smallskip
}
\newenvironment{questionbox}{\begin{qfcard}{Question:}}{\end{qfcard}}
\newenvironment{findingbox}{\begin{qfcard}{Finding:}}{\end{qfcard}}

\title{Follow the Mean: Reference-Guided Flow Matching}

\author{%
  \makebox[0pt][c]{%
  Pedro M. P.~Curvo\textsuperscript{1,2}\thanks{pedro.pombeiro.curvo@student.uva.nl}%
  \hspace{0.7em}%
  Maksim Zhdanov\textsuperscript{1,2}\thanks{Equal contribution.}%
  \hspace{0.7em}%
  Floor Eijkelboom\textsuperscript{1,2}\footnotemark[\value{footnote}]%
  \hspace{0.7em}%
  Jan-Willem van de Meent\textsuperscript{1,2}}
  \\[0.5em]
  \textsuperscript{1}AMLab\quad
  \textsuperscript{2}University of Amsterdam
  \\[0.35em]
  \raisebox{-0.95em}[0pt][0pt]{\normalsize\href{https://github.com/pedrocurvo/follow-the-mean}{\textcolor{githubpink}{\faGithub\ https://github.com/pedrocurvo/follow-the-mean}}}
}

\begin{document}

\maketitle

\begin{abstract}
Existing approaches to controllable generation typically rely on fine-tuning, auxiliary networks, or test-time search. We show that flow matching admits a different control interface: adaptation through examples. For deterministic interpolants, the velocity field is solely governed by a conditional endpoint mean, so shifting this mean shifts the flow itself, and a pretrained model can be steered simply by changing the reference set it follows. We instantiate this idea in two forms. Reference-Mean Guidance is training-free. It computes a closed-form correction to the endpoint mean from a reference bank and applies it to a frozen FLUX.2-klein (4B) model, controlling color, identity, style, and structure while the prompt, seed, and weights stay fixed. Semi-Parametric Guidance builds the same principle into a trained model through an explicit mean anchor and a learned residual refiner, matching the quality of an unconditional DiT-B/4 on AFHQv2 while allowing the reference set to be swapped at inference time. Together, these results point to a broader direction: generative models that adapt through data rather than parameter updates.
\end{abstract}

\begin{figure}[H]
    \centering
    \includegraphics[width=0.82\textwidth]{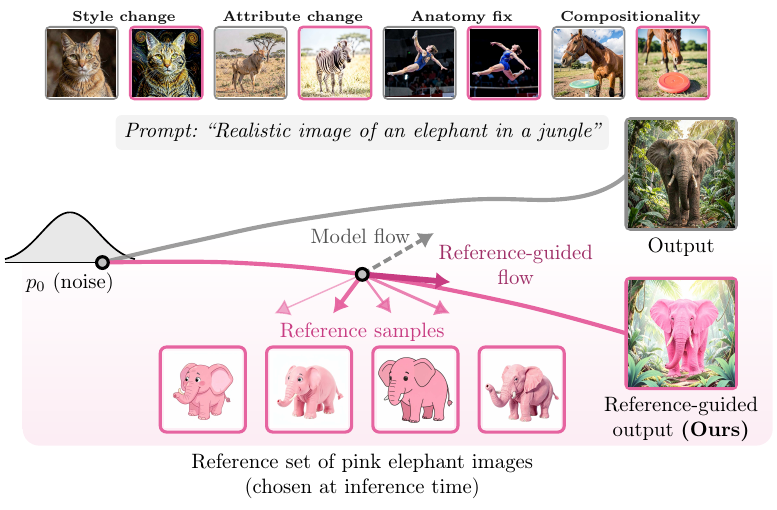}
    \caption{\textbf{Reference-guided flow matching steers a frozen model with examples.} A noisy state is matched against a reference set $\mathcal{R}$, which shifts the endpoint mean away from the pretrained prediction, so the flow takes on characteristics of the references without a classifier or reward.}
    \label{fig:method_overview}
\end{figure}

\section{Introduction}
\label{sec:introduction}

Flow matching~\citep{lipman2023} has emerged as a dominant paradigm for
training generative models, with recent approaches producing high-quality
samples across image, video, and scientific domains~\citep{lipman2023,
liu2023flow, albergo2023building, peebles2023, flux-2-2025}. Many downstream
applications, however, require control over the outputs of a pretrained model,
such as enforcing a specific attribute, concept, style, or target distribution
at generation time. Achieving such control without retraining the base model
remains a challenging problem.

Methods that address this problem fall into three broad families. \emph{Fine-tuning and adapter methods} update the parameters of the
model for each new target~\citep{ruiz2023dreambooth, hu2022lora}, or train a
conditioning module that accepts spatial or image conditions for a particular
modality and backbone~\citep{zhang2023adding, mou2024t2i, ye2023ip}.
\emph{Guidance methods} leave the generator unchanged and steer it with
auxiliary classifiers or reward signals~\citep{dhariwal2021diffusion,
feng2025on, potaptchik2025tilt}. \emph{Search-based methods} need no
additional training and instead sample repeatedly, filter candidates, or
optimize each prompt at inference
time~\citep{karthik2023dontsucceedtrytry,
wu2024practicalasymptoticallyexactconditional,
manas2024improvingtexttoimageconsistencyautomatic,
eyring2024renoenhancingonesteptexttoimage}. Each family thus obtains control
through additional training, an auxiliary network, or test-time search.

We propose a formulation that needs none of these, which we call
\emph{reference-guided flows}. The object we control is the endpoint mean, the mean of the posterior
distribution over data points given a noisy interpolant. In flow matching the
velocity field points toward this mean~\citep{lipman2023,
albergo2023building, eijkelboom2024variational}, so shifting the mean also
shifts the distribution of generated samples. Our key observation is that
this shift is easy to compute from a set of reference samples. Crucially,
these references need not represent the target distribution perfectly, as long as they move
the mean in the desired direction, and a reference set therefore provides
implicit guidance even when no reward or classifier is defined. In short:
\begin{center}
\small\textbf{\emph{``Guide with examples, not rewards.''}}
\end{center}

\Cref{fig:method_overview} illustrates the idea on a frozen text-to-image model. Prompted with ``an elephant in a jungle'', the model produces a photorealistic elephant, and conditioning on a small set of pink-elephant images turns the elephant pink.

\begin{theorybox}
\vspace{0.2em}
\textbf{Contributions.}
\vspace{0.3em}

\begin{enumerate}[leftmargin=2.8em,itemsep=0.55em,topsep=0.2em,labelsep=0.5em,rightmargin=1.8em]

    \item \textbf{Mean shift as a mechanism for guidance.} We derive guidance from a geometric interpolation between the probability paths induced by the training set and by a reference set. When the endpoint posteriors are Gaussian with a shared covariance, the guidance term reduces to the difference between the endpoint means of the reference set and the training distribution, and an exact arithmetic-mixture construction leads to the same rule.

    \item \textbf{Reference-mean guidance.} We show that the shift in the velocity field can be computed efficiently from the closed-form endpoint mean of the reference set. The result is a training-free method for test-time control of pretrained models that needs no additional training, auxiliary networks, gradients, or extra model evaluations.

    \item \textbf{Semi-parametric guidance.}
    We introduce a separately trained reference-conditioned architecture that combines a learned reference anchor with a residual refiner. Because it learns the similarity that decides which references to attend to, SPG keeps useful reference information and suppresses nuisance correlations such as shared backgrounds. At the same time, it matches unconditional generation quality and remains controllable through its reference set at inference time.
\end{enumerate}

\vspace{0.3em}
\end{theorybox}

\section{Background}
\label{sec:background}

\subsection{Flow Matching}

Flow matching (FM) learns a continuous-time transport model that maps a source distribution $p_0$ to a target distribution $p_1$ \cite{lipman2023,liu2023flow,albergo2023building}. To do so, it defines a time-dependent distribution $p_t$, known as the probability path, through an affine interpolant between a source sample $x_0 \sim p_0$ and a data sample $x_1 \sim p_1$,
\begin{equation}
x_t = \alpha_t x_0 + \beta_t x_1, \qquad t \in [0,1].
\label{eq:linear_bridge}
\end{equation}
Samples are generated by integrating the ordinary differential equation $\dot x = u_t(x)$ from $t=0$ to $t=1$. This equation transports samples from $p_0$ along the path $p_t$ when the velocity field satisfies the continuity equation $\partial_t p_t(x) + \nabla \cdot (p_t(x) u_t(x)) = 0$, which holds for the conditional expectation
\begin{align}
    u_t(x) := \mathbb{E}\big[\dot{\alpha}_t x_0 + \dot{\beta}_t x_1 \,\big|\, x_t=x \big].
\end{align}
This expectation takes a particularly simple form once it is written in terms of the data endpoint $x_1$.
\begin{theorybox}
For the common linear interpolant with $\alpha_t=(1-t)$ and $\beta_t=t$, the velocity field depends on the data only through the endpoint mean $\mu_t(x)$ (\Cref{app:proof_prop1} gives a derivation that also covers general affine interpolants),
\begin{align}
\label{eq:fm-linear-velocity}
u_t(x) &= \frac{\mu_t(x)-x}{1-t},
&
\mu_t(x) := \mathbb{E}[x_1 \mid x_t=x].
\end{align}
Intuitively, the velocity field moves a sample at $x$ toward $\mu_t(x)$, at a rate inversely proportional to the remaining time $1-t$.
\end{theorybox}

The identity in \eqref{eq:fm-linear-velocity} is invertible, so the endpoint mean $\mu_t(x) = x + (1-t) u_t(x)$ can equally be written in terms of the velocity field. A flow matching model can therefore be parameterized either by $u^\theta_t(x)$ or by $\mu_t^\theta(x)$. The objective can likewise be defined on the predicted velocity, through the standard flow matching loss \cite{lipman2023,liu2023flow,albergo2023building}, or on the mean $\mu_t^\theta(x)$ of a variational distribution $q^\theta_t(x_1 \mid x_t)$, through the variational flow matching loss \cite{eijkelboom2024variational},
\begin{align}
    \label{eq:fm-objectives}
    \mathcal{L}^\text{FM}(\theta)
    &=
    \mathbb{E}\Big[ \big\|(x_1 - x_0) - u^\theta_t(x_t) \big\|^2 \Big],
    &
    \mathcal{L}^\text{VFM}(\theta)
    &=
    -\mathbb{E}\Big[ \log q_t^\theta(x_1 \mid x_t) \Big],
\end{align}
where $x_0 \sim p_0$, $x_1 \sim p_1$ and $t \sim \text{Uniform}([0,1])$. Either parameterization can be combined with either loss, so every pretrained model specifies both a velocity field and an endpoint mean. An analogous observation holds for diffusion models \cite{gagneux2026training}.

\subsection{Closed Form of the Endpoint Mean}
\label{sec:closed-form-mean}

In practice, a flow matching model is trained on an empirical distribution $\hat{p}_1$ over a finite training set $\mathcal{D} = \{x^{(1)}, \dots, x^{(N)}\}$. The posterior over $x_1$ given $x_t$ is then a distribution over the $N$ training points, so the endpoint mean of \eqref{eq:fm-linear-velocity} can be computed exactly.

\begin{theorybox}
Applying Bayes' rule to the empirical distribution $\hat{p}_1(x_1)$ gives the closed form
\begin{align}
    \label{eq:empirical-endpoint-mean}
    \hat{\mu}_t(x) &= \sum_{n=1}^N w^{(n)}_t(x) \: x^{(n)},
    &
    w^{(n)}_t(x) &= \frac{p_t(x_t\!=\!x \mid x_1\!=\!x^{(n)})}{\sum_{m=1}^N p_t(x_t\!=\! x \mid x_1 = x^{(m)})}
    .
\end{align}
For a standard normal source $p_0 = \mathcal{N}(0,I)$, the weights become (\Cref{app:proof_prop2})
\begin{align}
    \label{eq:weights-softmax}
    w^{(n)}_t(x)
    = \text{Softmax}_n
    \left(
        - \frac{1}{2}
          \frac{\|x - t x^{(n)}\|^2}
               {(1-t)^2}
    \right)
    =
    \text{Softmax}_n
    \left(\frac{ t x^\top x^{(n)}  - \tfrac{1}{2} t^2 \|x^{(n)}\|^2 }{ (1-t)^2}
    \right).
\end{align}
\end{theorybox}
The learned endpoint mean $\mu^\theta_t(x)$ thus approximates $\hat{\mu}_t(x)$, a weighted sum over the training set. This observation has been made for flow matching \cite{bertrand2026on,gao2024flow} and for score matching \cite{niedoba2024nearest,scarvelis2025closedform} (\Cref{sec:related_work}), and to our knowledge it has not yet been used to design guidance methods.

\section{Reference-Guided Flows}
\label{sec:theory}

\subsection{Steering a Flow by Shifting the Endpoint Mean}
\label{sec:mean-shift}

Our method starts from a simple observation. Consider a pretrained flow
model that approximates the velocity $u_t(x)$ and endpoint mean $\mu_t(x)$ of
a training distribution $p_1$, and suppose that at test time we want to
generate from a different target distribution $\pi_1$. Let $\pi_t$ be its
path under the same bridge, with endpoint mean $\mu_t^\pi(x)$. Because both
flows share the source and the bridge, their velocity fields differ only
through their endpoint means,
\begin{equation}
\label{eq:velocity-difference}
u_t^\pi(x) - u_t(x) \;=\; \frac{\mu_t^\pi(x) - \mu_t(x)}{1-t}.
\end{equation}
Any target distribution $\pi_1$ can therefore be reached by approximating
the shift $\mu_t^\pi(x) - \mu_t(x)$ during generation, using the pretrained
estimate $\mu^\theta_t(x) = x + (1-t)\,u^\theta_t(x)$.

\subsection{Reference-Mean Guidance (RMG)}
\label{sec:rmg}

We now use a set of reference samples to specify $\mu^\pi_t(x)$ implicitly.
Let $\mathcal{R} = \{x^{(1)}, \dots, x^{(M)}\}$ be a reference set sampled
from a distribution $\rho_1$. Our goal is to shift the generated distribution
toward the endpoint mean $\mu^\rho_t(x)$ induced by the references while
preserving the diversity and quality of the pretrained model.

\paragraph{Geometric mixture at the endpoint level.}
For a constant $\lambda \in [0,1]$, consider the geometric mixture of the
training and reference endpoint distributions,
\[
\pi(x_1) \propto p_1(x_1)^{1-\lambda}\,\rho_1(x_1)^{\lambda},
\]
and let $\pi_t(x) = \int p_t(x \mid x_1)\,\pi(x_1)\,dx_1$ be its noisy
marginal under the same affine bridge, which is a valid bridge marginal by
construction. Because the bridge kernel is shared, the endpoint posterior of
$\pi$ is the geometric mixture of the two component posteriors. When both are
Gaussian, its mean is a precision-weighted average of their means, which
becomes a convex combination when the covariances agree
(\Cref{app:proof_thm3}).

\begin{theorybox}
\textbf{Proposition 3.3 (Reference-mean guided dynamics).}
Let $\lambda \in [0,1]$ and let $\pi \propto p_1^{1-\lambda}\rho_1^{\lambda}$.
If the endpoint posteriors of $p_1$ and $\rho_1$ given $x_t = x$ are Gaussian
with a common covariance, then
\begin{align}
\label{eq:guided_posterior_mean}
\mu_t^\pi(x) &\;=\; (1-\lambda)\,\mu_t(x) + \lambda\,\mu_t^\rho(x), \\[4pt]
\label{eq:retrieval_guided_velocity}
u_t^\pi(x) &\;=\; u_t(x)
+ \lambda\,\frac{\mu_t^\rho(x) - \mu_t(x)}{1-t}.
\end{align}
\end{theorybox}

\paragraph{Remark.} The arithmetic mixture
$\pi_\lambda = (1-\lambda)\,p_1 + \lambda\,\rho_1$ offers a second route. Its
noisy marginal $(1-\lambda)\,p_t + \lambda\,\rho_t$ is again a valid bridge
marginal, and Bayes' rule gives its posterior mean exactly,
\begin{equation}
\label{eq:arithmetic_mean}
\mu_t^\lambda(x)
= \bigl(1 - \omega_t^*(x)\bigr)\,\mu_t(x)
+ \omega_t^*(x)\,\mu_t^\rho(x),
\qquad
\omega_t^*(x)
= \frac{\lambda\,\rho_t(x)}{(1-\lambda)\,p_t(x) + \lambda\,\rho_t(x)}.
\end{equation}
This velocity field satisfies the continuity equation with no assumption on
$p_1$ or $\rho_1$ (\Cref{app:reference_sets}). The responsibility
$\omega_t^*(x)$ depends on the density $p_t(x)$, which a pretrained velocity
field does not provide, and replacing it with the scalar $\lambda$ recovers
Proposition~3.3. The two constructions thus arrive at the same guidance rule,
one through a distributional assumption and the other through a scalar
approximation of the responsibilities.

In practice we make two further approximations. We replace $\mu_t$ by the
pretrained estimate $\mu_t^\theta$, and $\mu_t^\rho$ by the empirical mean
$\hat{\mu}_t^\rho$ over a finite reference bank $\mathcal{R}$, computed in
closed form with \Cref{eq:empirical-endpoint-mean}. Such a small, curated
bank is rarely Gaussian, so we use Proposition~3.3 to motivate the
\emph{form} of the correction, a difference of endpoint means, rather than as
an exact sampling guarantee. Changing the composition of $\mathcal{R}$ then
directly changes the guided velocity field. We call the
resulting method \emph{reference-mean guidance} (RMG),
\begin{equation}
\label{eq:practical_rmg}
    u^{\mathrm{RMG}}_t(x) = u^\theta_t(x)
    + \gamma_t \frac{\hat{\mu}^\rho_t(x) - \mu^\theta_t(x)}{1-t},
\end{equation}
where the scalar schedule $\gamma_t$ takes the place of the mixture
coefficient.\footnote{With a time-varying $\gamma_t$, \Cref{eq:practical_rmg}
is a guidance rule rather than exact transport (\Cref{app:reference_sets}).} As $t \to 1$, the factor $(1-t)^{-1}$ diverges and the weights in
\Cref{eq:weights-softmax} collapse onto the nearest reference, so $\gamma_t$
decays toward the end of the trajectory and guidance is switched off shortly
before the endpoint, which makes the correction vanish at $t=1$
(\Cref{app:proof_thm3}).

\subsection{Semi-Parametric Guidance (SPG)}
\label{sec:spg}

SPG is a separately trained reference-conditioned model with its own
parameters $\phi$, distinct from the frozen model $\theta$ used by RMG. Rather
than implementing the mixtures above, it targets an implicit
reference-conditioned distribution whose mass moves toward whatever reference
set it receives at inference time. Its key difference from RMG is that it
learns what to transfer. In RMG, the closed-form weights of
\Cref{eq:weights-softmax} measure similarity by raw latent distance, so every
attribute that the references share is transferred, including unintended ones such as the light backgrounds of
the pink-elephant bank in \Cref{fig:flux_bank_swap}
(\Cref{app:spg_background_nocopy}). SPG instead learns the similarity between
the noisy state and the references over patchified latents, and with it
\emph{which} parts of a reference set to transfer.

\paragraph{Architecture.}
The model first computes an anchor $\bar{x}$ with a cross-attention pass
over the reference set, in which learned queries and keys replace the
closed-form weights and the values are the identity,
\begin{align}
\label{eq:spg_anchor}
    \bar{x}
    &=
    \sum_{m=1}^M a_m x^{(m)},
    &
    a
    =
    \text{Softmax}_m
    \left(
        \langle q^\phi(x_t), k^\phi(x^{(m)}) \rangle
    \right).
\end{align}
The anchor is not meant to reproduce the closed-form mean
$\hat{\mu}^\rho_t$, which \Cref{eq:empirical-endpoint-mean} already gives as
a cheap softmax. Instead, the learned similarity lets the model weight
references by the attributes it should transfer, while the identity values
keep $\bar{x}$ an average of the reference latents themselves. The bank
therefore enters the prediction directly, and swapping the bank changes the
behavior. Because the anchor is a convex combination of references, it
captures the content inside their hull, which in practice is shared, coarse
structure, and the refiner $f^\phi$ models everything beyond it, recovering
the details from $x_t$ given the coarse anchor. Finally, learned
time-dependent gates combine the noisy state, the anchor, and the residual
into the final endpoint prediction,
\begin{equation}
\label{eq:spg_prediction}
    \mu^\phi_t(x_t, \mathcal{R})
    =
    (1 - g_t) \cdot x_t
    + g_t \cdot \bar{x}
    + \eta_t \cdot f^\phi\!\bigl(\bar{x},\, x_t,\, t\bigr),
\end{equation}
with $g_t, \eta_t \in [0,1]$. Late in the trajectory $x_t$ is almost the
final image and should dominate, whereas early on the anchor carries most of
the signal (\Cref{app:spg_training}).

\paragraph{Training.}
During training, the reference set $\mathcal{R}$ is sampled from the training
set. For each of its elements we form an interpolant $x^{(m)}_t$ and
condition on the remaining references
$\mathcal{R}^{\setminus \{m\}} := \mathcal{R} \setminus \{x^{(m)}\}$, which
gives the batch-level endpoint-prediction objective
\begin{equation}
    \mathcal{L}_{\mu}(\phi)
    =
    \mathbb{E}
    \left[
    \sum_{m=1}^M
    \frac{1}{(1-t)^2}
    \left\|
    x^{(m)}
    -
    \mu_t^\phi \big( x_t^{(m)}, \mathcal{R}^{\setminus \{m\}} \big)
    \right\|^2
    \right].
\end{equation}
Leaving one reference out prevents $x_t^{(m)}$ from attending to its own
endpoint, so the model cannot simply retrieve it. The anchor is already a
strong predictor, which leaves the refiner little gradient signal from
$\mathcal{L}_\mu$, so we also train the refiner on the residual between the
target and the anchor, with gradients stopped through the anchor,
\begin{equation}
    \mathcal{L}_{\mathrm{ref}}(\phi)
    =
    \mathbb{E}
    \left[
    \sum_{m=1}^M
    \left\|
    \mathrm{sg}\!\left[
        x^{(m)} - \bar{x}^{(m)}
    \right]
    -
    f^\phi\!\left(
        \mathrm{sg}\!\left[\bar{x}^{(m)}\right],
        x_t^{(m)},
        t
    \right)
    \right\|^2
    \right],
\end{equation}
where $\bar{x}^{(m)}$ is the anchor computed from $x_t^{(m)}$ and
$\mathcal{R}^{\setminus \{m\}}$, and $\mathrm{sg}[\cdot]$ denotes
stop-gradient.

Reference sets are sampled independently of the target during training,
so $\mathbb{E}[x^{(m)} \mid x^{(m)}_t, \mathcal{R}^{\setminus\{m\}}]
= \mathbb{E}[x^{(m)} \mid x^{(m)}_t]$ and the objective alone does not
require the model to use $\mathcal{R}$. The incentive comes instead from the
architecture, since the anchor is built from the references themselves and
the stop-gradient keeps the refiner from taking over its role. As a result,
the trained model does rely on its references in practice
(\Cref{sec:spg_results,app:spg_training}). For example, swapping banks changes the class
and even the internal attributes of the output, and a bank of pure noise
generates noise. The anchor on its own, by contrast, yields only coarse
structure, so the fine detail comes from the refiner.

\section{Experiments}
\label{sec:results}

Reference-guided flows replace rewards and auxiliary networks with a set of
examples. We first verify this mechanism on toy data, where the endpoint mean
is exact, and then evaluate its two instantiations, RMG on a frozen 4B
text-to-image model and SPG as a trained reference-conditioned model. Four
questions guide the evaluation.

\begin{enumerate}[leftmargin=2.4em,itemsep=1pt,topsep=2pt,parsep=0pt]
  \item[\textbf{Q1.}] Does the reference set steer the flow through the
  endpoint mean?
  \item[\textbf{Q2.}] Can swapping the reference set control a frozen
  text-to-image model?
  \item[\textbf{Q3.}] How does a small reference bank compare with search-
  and gradient-based control?
  \item[\textbf{Q4.}] Can a trained model learn to use reference sets
  without losing quality?
\end{enumerate}

\subsection{Experimental Setup}
\label{sec:experimental_setup}

\paragraph{Models and data.}
Q1 uses the two-dimensional two-moons distribution, for which the endpoint
mean can be computed exactly. Q2 and Q3 apply RMG (\Cref{sec:rmg}) to a
frozen \mbox{FLUX.2-klein} (4B) model~\citep{flux-2-2025} at
$768\times768$ resolution. FLUX.2 is a latent rectified flow with endpoint
estimate $\mu_t^\theta(x) = x + (1-t)u_t^\theta(x)$, and the references are
encoded with the same frozen VAE, so every correction acts in the latent space
of the pretrained model. For Q4 we train SPG (\Cref{sec:spg}) and a DiT-B/4
baseline on AFHQv2~\citep{choi2020stargan} at $256\times256$ resolution, in the latent space of a
frozen REPA-E VAE~\citep{leng2025repae} (\Cref{app:spg_training,app:spg_setup}).

\paragraph{Guidance.}
RMG uses the schedule $\gamma_t = \gamma_0(1-t)^2$, under which the
correction scales as $\gamma_0(1-t)$, and guidance is switched off for
$t \geq 0.85$. In latent space, the softmax of \Cref{eq:weights-softmax} is
evaluated at temperature $\sqrt{d}$, where $d$ is the latent dimension. The
strength $\gamma_0$ of each experiment is listed in \Cref{app:flux_sampling},
and \Cref{app:ablations} ablates the schedule, its strength, and the number
of sampling steps.

\paragraph{Reference banks.}
Unless stated otherwise, a bank holds 20 images that share a target
attribute, such as a color, an object, or a style (\Cref{app:flux_banks}).
When we swap banks, the prompt, noise seed, and model weights stay fixed, so
any change in the output comes from the bank.

\paragraph{Baselines and metrics.}
For Q3 we compare RMG with prompt
optimization~\citep{manas2024improvingtexttoimageconsistencyautomatic},
Best-of-4~\citep{karthik2023dontsucceedtrytry},
SMC~\citep{wu2024practicalasymptoticallyexactconditional}, and
ReNO~\citep{eyring2024renoenhancingonesteptexttoimage} on
GenEval~\citep{ghosh2023geneval}. All methods share the backbone, the 20-step
sampler, the prompts, and the seeds. We report GenEval accuracy for Q3, and
FID, KID, and IS together with CLIP-labeled class frequencies for Q4.
Prompts and the remaining metrics are given in \Cref{app:experiments}.

\subsection{Q1: The Endpoint Mean Controls the Flow}
\label{sec:pgd}
\label{sec:toy_experiments}

\begin{questionbox}
Does the reference set steer the flow through the endpoint mean?
\end{questionbox}

\begin{figure}[H]
\centering

\begin{tabular}{@{}c@{\hspace{0.02\linewidth}}c@{}}
\textbf{15\% class 1} & \textbf{85\% class 1} \\[0.3em]

\begin{tabular}{@{}c@{\hspace{0.01\linewidth}}c@{}}
\small Flow & \small Trajectory \\
\includegraphics[width=0.19\linewidth]{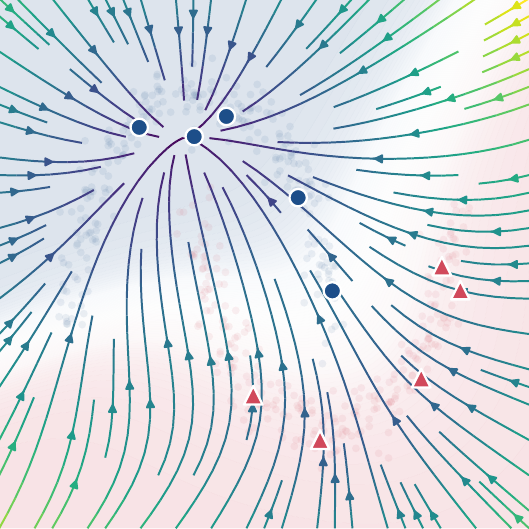} &
\includegraphics[width=0.19\linewidth]{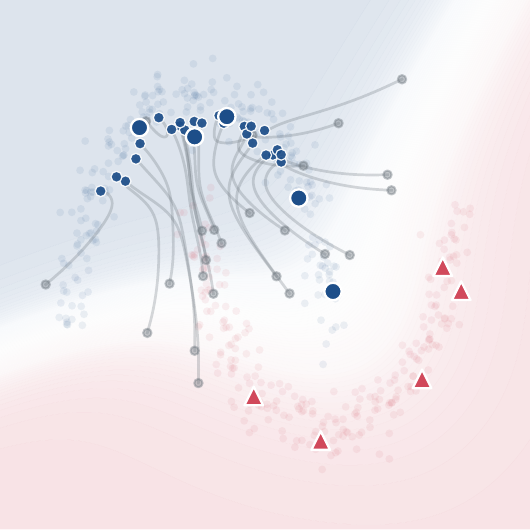}
\end{tabular}
&
\begin{tabular}{@{}c@{\hspace{0.01\linewidth}}c@{}}
\small Flow & \small Trajectory \\
\includegraphics[width=0.19\linewidth]{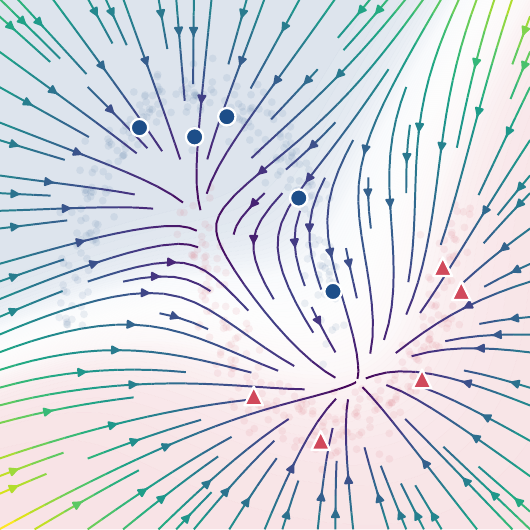} &
\includegraphics[width=0.19\linewidth]{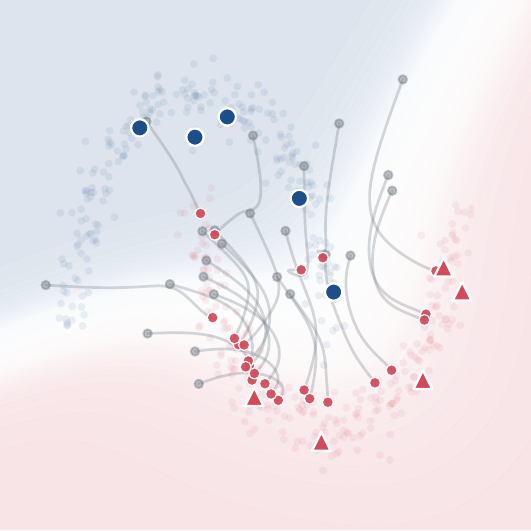}
\end{tabular}
\end{tabular}
\caption{\textbf{The composition of the reference set alone decides where the
flow goes.} Flow field and trajectories on two moons for reference sets with
15\% (left) and 85\% (right) class~1, with the model fixed. With few class-1
references the flow converges onto the upper moon (class~0, blue), and
raising their share moves the attractor, and the samples, to the lower moon
(class~1, red).}
\label{fig:twomoons_flow}
\end{figure}

We draw $N=500$ points from the two-moons distribution and withhold their
labels from the model. A small labeled reference set enters only at inference
time, through the soft posterior weights. Varying only its composition, with
the model and all other settings fixed, moves the flow field and its final
attractor (\Cref{fig:twomoons_flow}), which isolates the causal role of the
endpoint mean predicted by \Cref{sec:theory}.
\Cref{app:mechanistic_validation} extends the analysis, showing that the
posterior concentrates on the class structure as $t \to 1$
(\Cref{fig:twomoons_combined}), that as few as $M=5$ references come close to
the hard-filter upper bound (\Cref{fig:hard_soft_twomoons}), and that the same
mechanism steers generation in pixel space on MNIST
(\Cref{fig:mnist_results}).

\begin{findingbox}
The reference set is a control knob for the flow itself. Changing only the
examples that the endpoint mean averages over moves the attractor, with the
model untouched.
\end{findingbox}

\subsection{Q2: Training-Free Control of a Frozen Model}
\label{sec:flux_results}

\begin{questionbox}
Can swapping the reference set control a frozen text-to-image model?
\end{questionbox}

\begin{figure}[ht]
\centering
\setlength{\tabcolsep}{0pt}
\renewcommand{\arraystretch}{0.8}
\begin{tabular}{@{}r@{\hspace{0.01\linewidth}}c@{\hspace{0.01\linewidth}}c@{\hspace{0.01\linewidth}}c@{\hspace{0.01\linewidth}}c@{}}
& \makebox[0.20\linewidth][c]{\small\textit{elephant in a jungle}}
& \makebox[0.20\linewidth][c]{\small\textit{a cat}}
& \makebox[0.20\linewidth][c]{\small\textit{a house in a forest}}
& \makebox[0.20\linewidth][c]{\small\textit{animal in a savanna}} \\[0.3em]

\rotatebox{90}{\small\textbf{Baseline}}
& \includegraphics[width=0.20\linewidth]{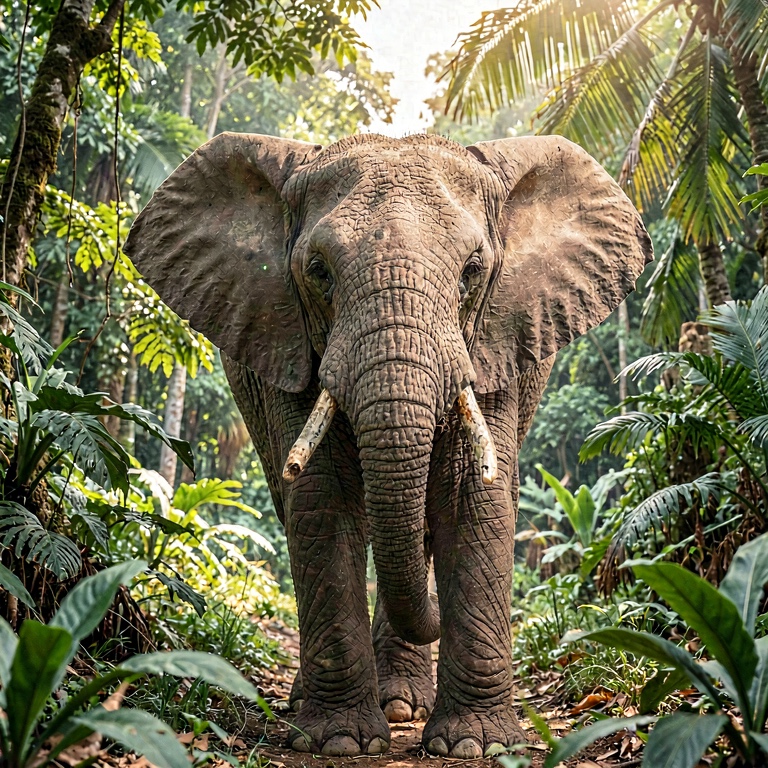}
& \includegraphics[width=0.20\linewidth]{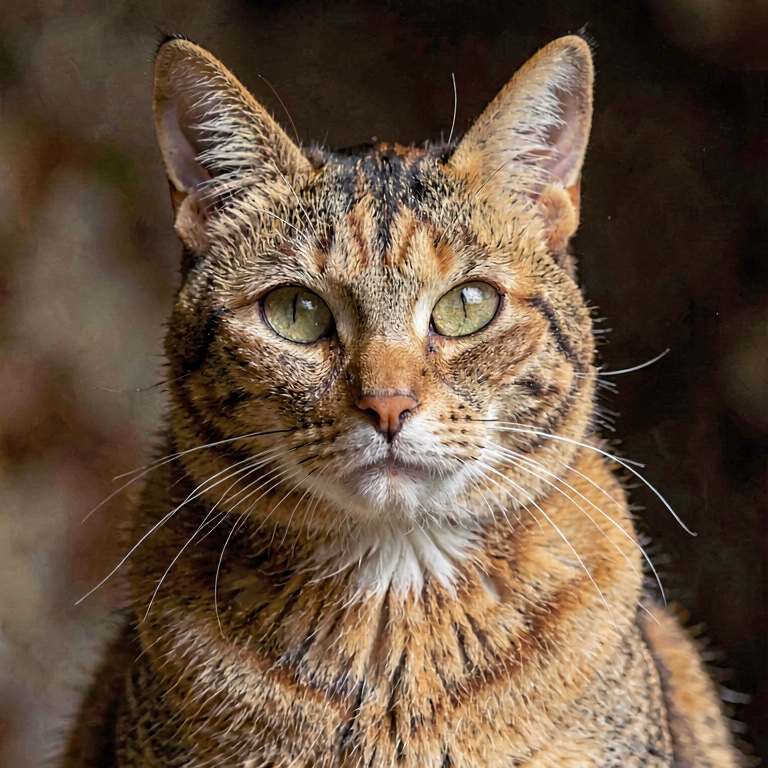}
& \includegraphics[width=0.20\linewidth]{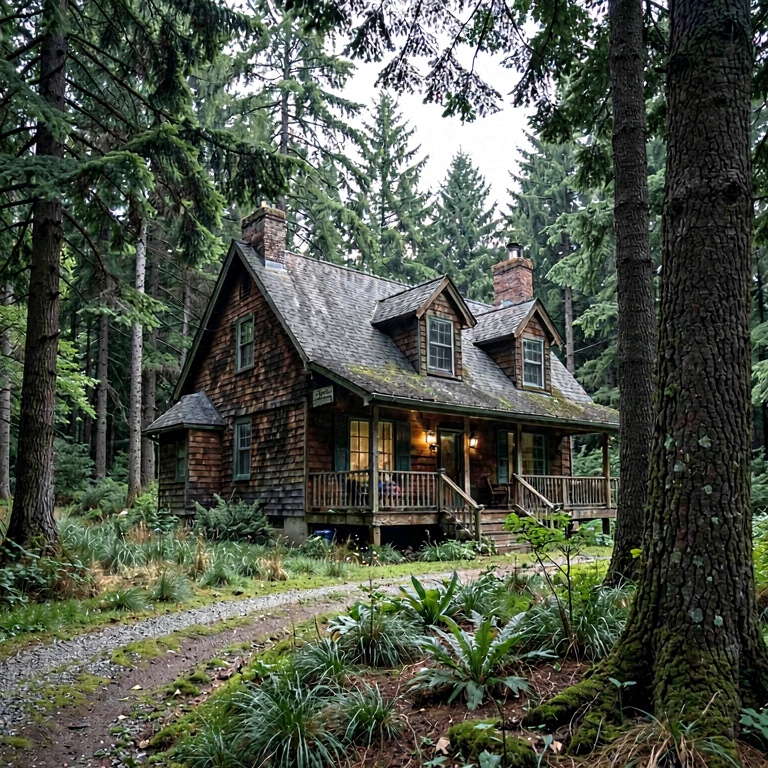}
& \includegraphics[width=0.20\linewidth]{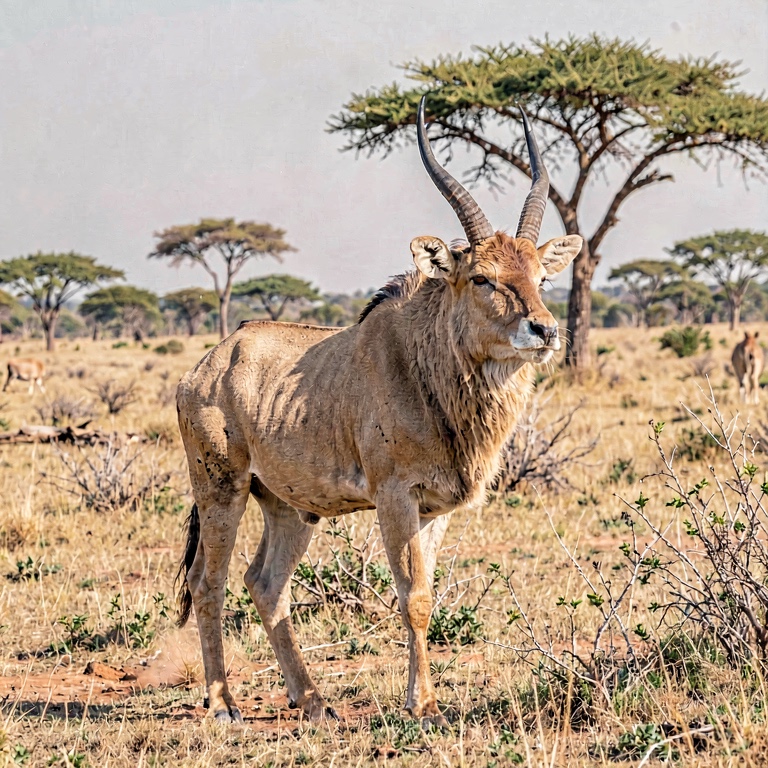}\\[0.25em]

\rotatebox{90}{\small\textbf{Reference A}}
& \includegraphics[width=0.20\linewidth]{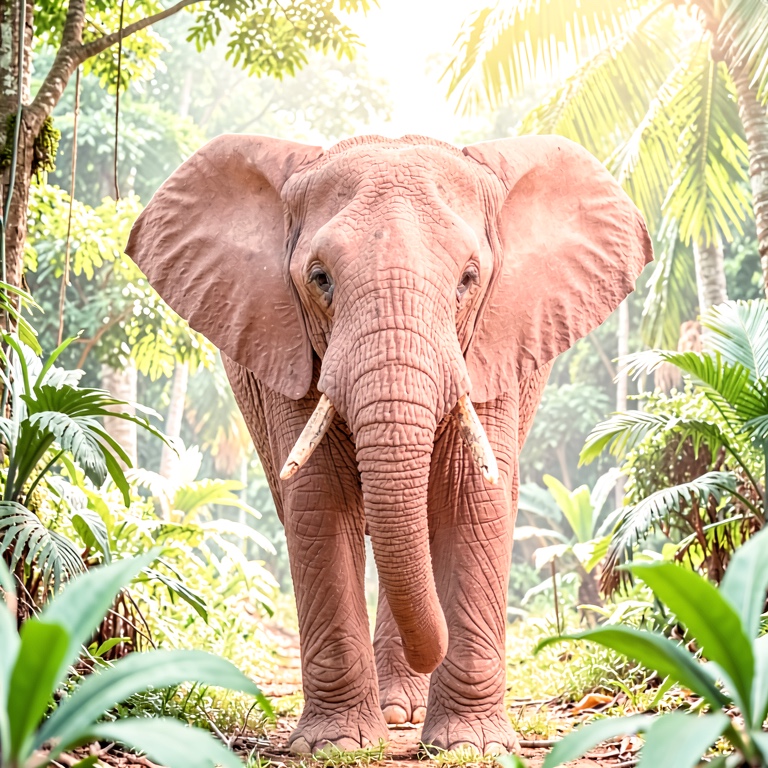}
& \includegraphics[width=0.20\linewidth]{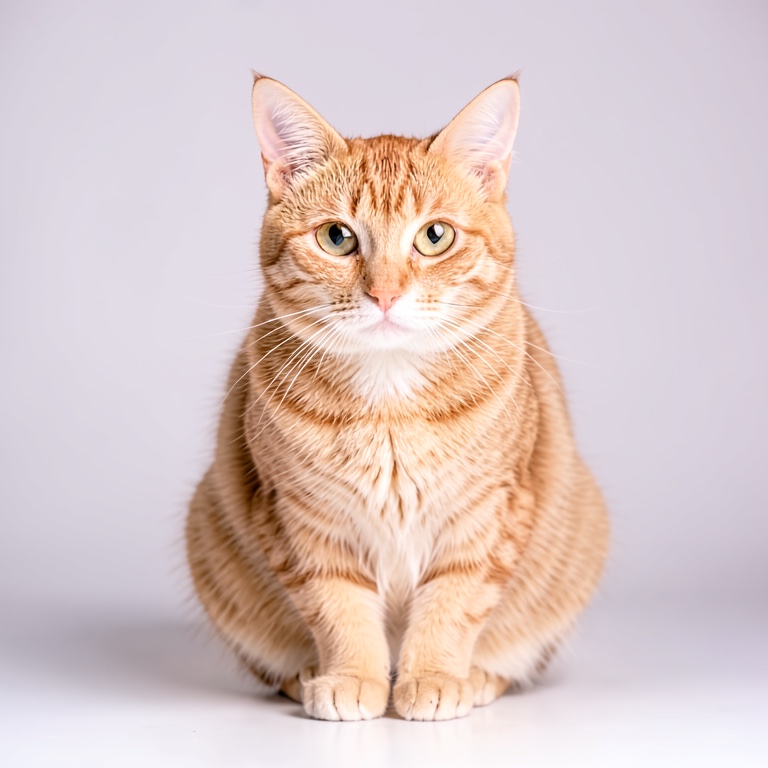}
& \includegraphics[width=0.20\linewidth]{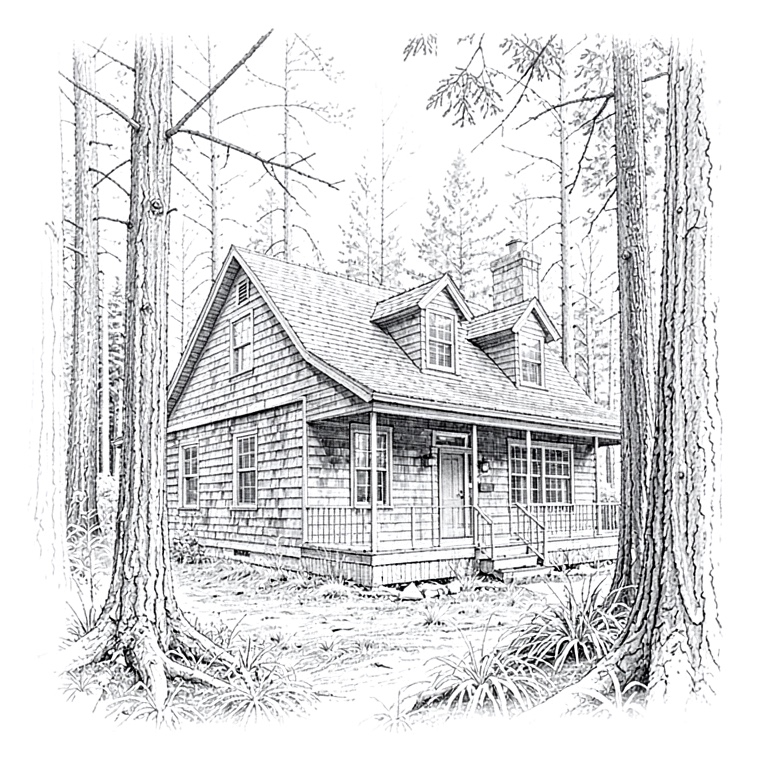}
& \includegraphics[width=0.20\linewidth]{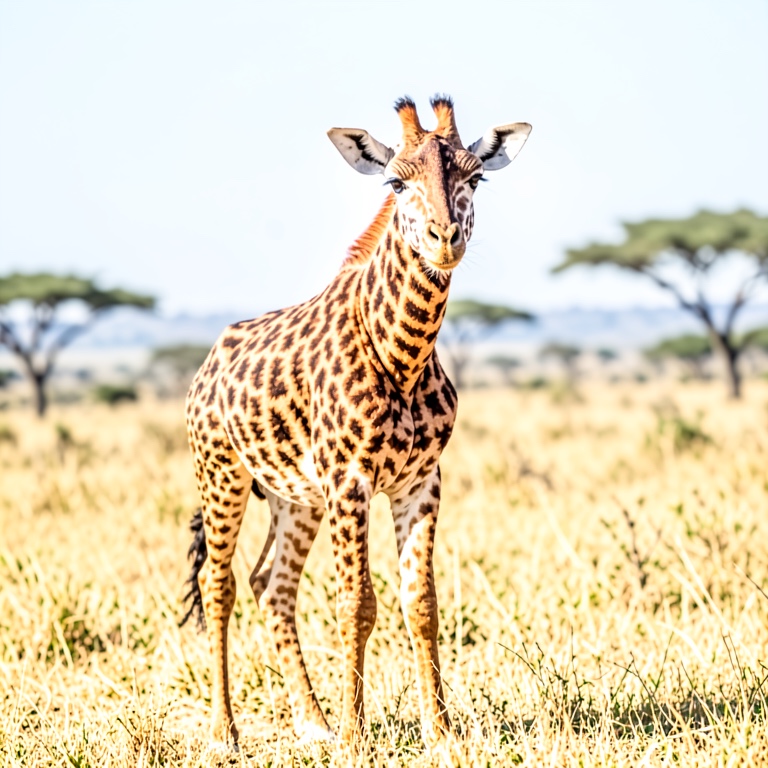} \\
& \small pink elephants & \small studio photos & \small sketches & \small giraffes \\[0.25em]

\rotatebox{90}{\small\textbf{Reference B}}
& \includegraphics[width=0.20\linewidth]{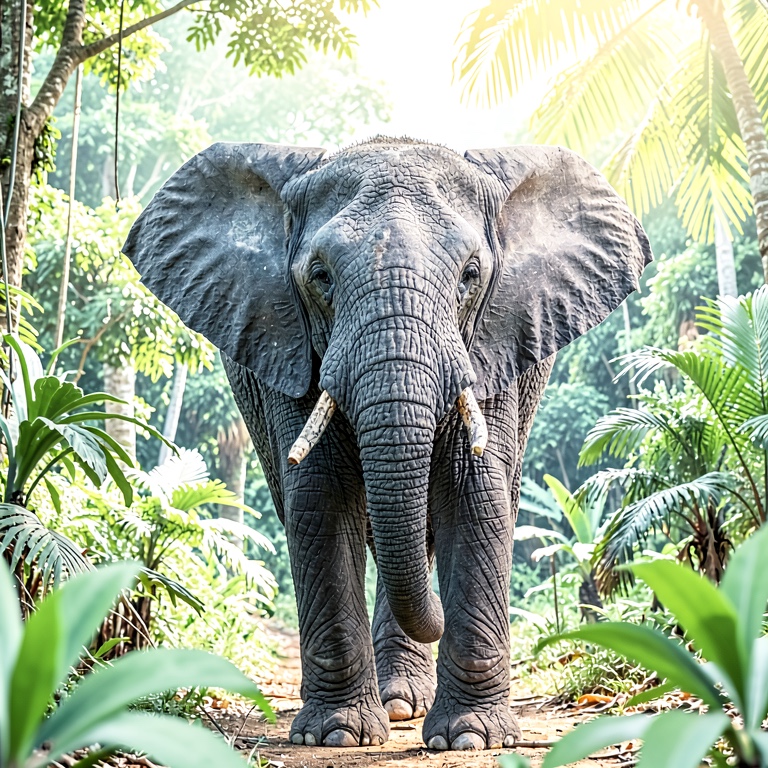}
& \includegraphics[width=0.20\linewidth]{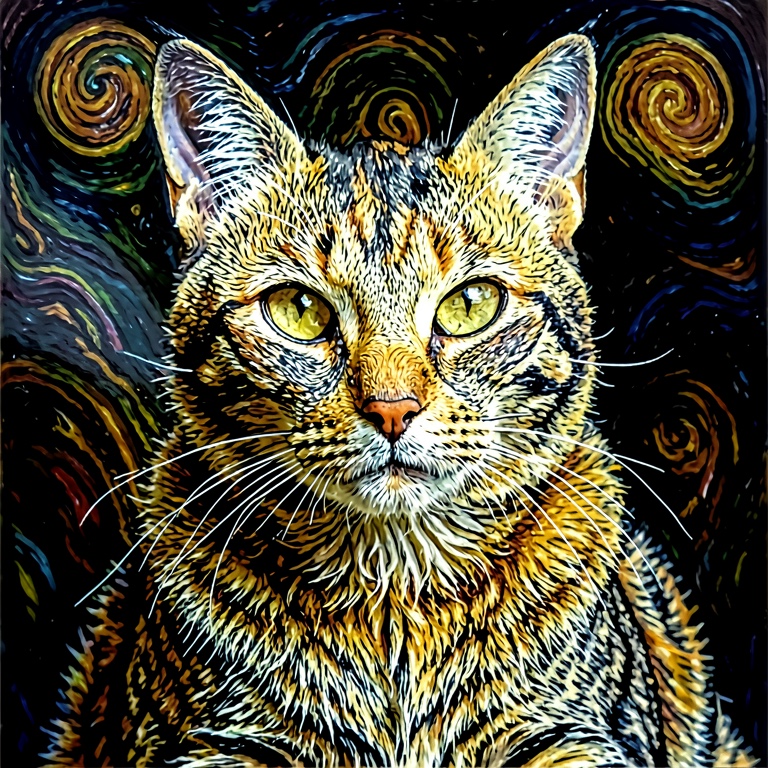}
& \includegraphics[width=0.20\linewidth]{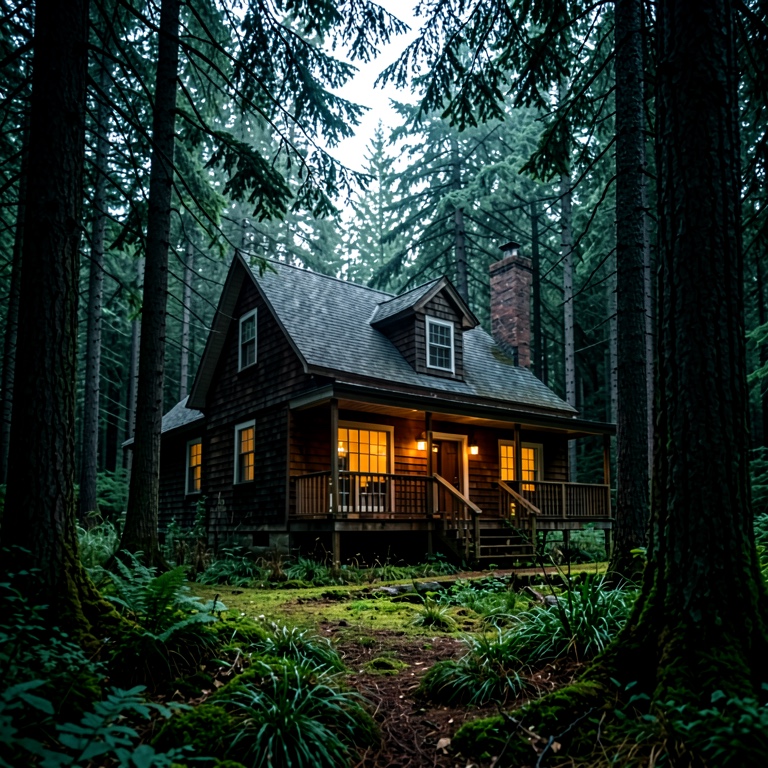}
& \includegraphics[width=0.20\linewidth]{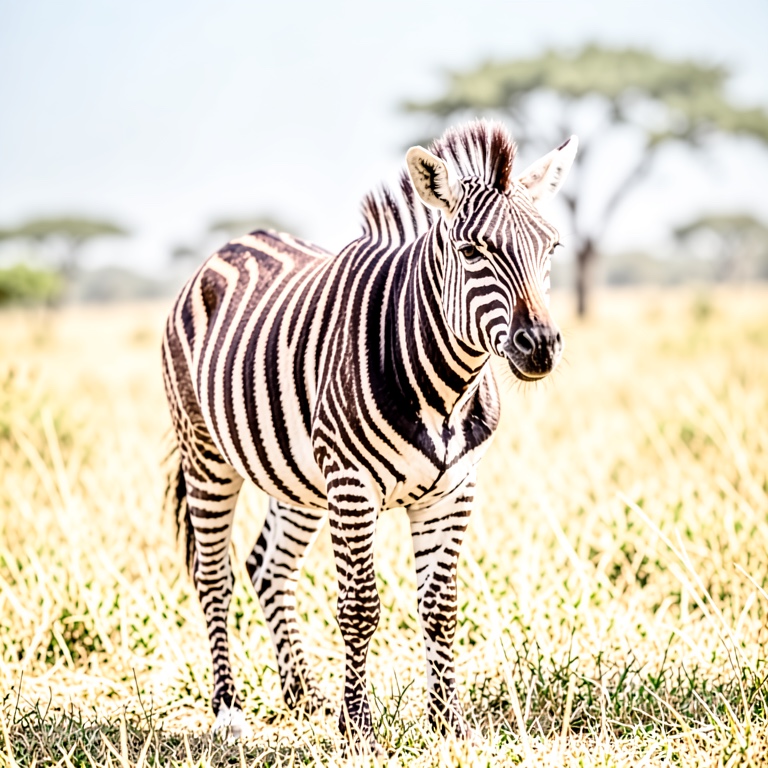} \\
& \small blue elephants & \small Van Gogh & \small cinematic & \small zebras \\
\end{tabular}
\caption{\textbf{Swapping the reference set changes color, identity, and
style in a frozen model.} Outputs of \mbox{FLUX.2-klein} without guidance
(top) and with two different reference sets (middle and bottom). The prompt
and noise seed are fixed within each column.}
\label{fig:flux_bank_swap}
\end{figure}

\paragraph{Attributes.}
\Cref{fig:flux_bank_swap} shows four prompts, each paired with two banks that
encode different attributes. In every case the output follows the bank in
color, object identity, or style, and the endpoint mean induced by the bank
thus acts as a control signal for the frozen model.

\begin{figure}[ht]
\centering
\setlength{\tabcolsep}{0pt}
\makebox[\linewidth][c]{%
\begin{tabular}{@{}c@{\hspace{0.012\linewidth}}c@{\hspace{0.012\linewidth}}c@{}}
\makebox[0.21\linewidth][c]{\small\textbf{Baseline}} &
\makebox[0.21\linewidth][c]{\begin{tabular}{@{}c@{}}\small\textbf{Reference-guided}\\[-0.1em]\small\textbf{control}\end{tabular}} &
\makebox[0.21\linewidth][c]{\begin{tabular}{@{}c@{}}\small\textbf{Reference-set}\\[-0.1em]\small\textbf{nearest neighbor}\end{tabular}} \\[0.3em]
\includegraphics[width=0.21\linewidth]{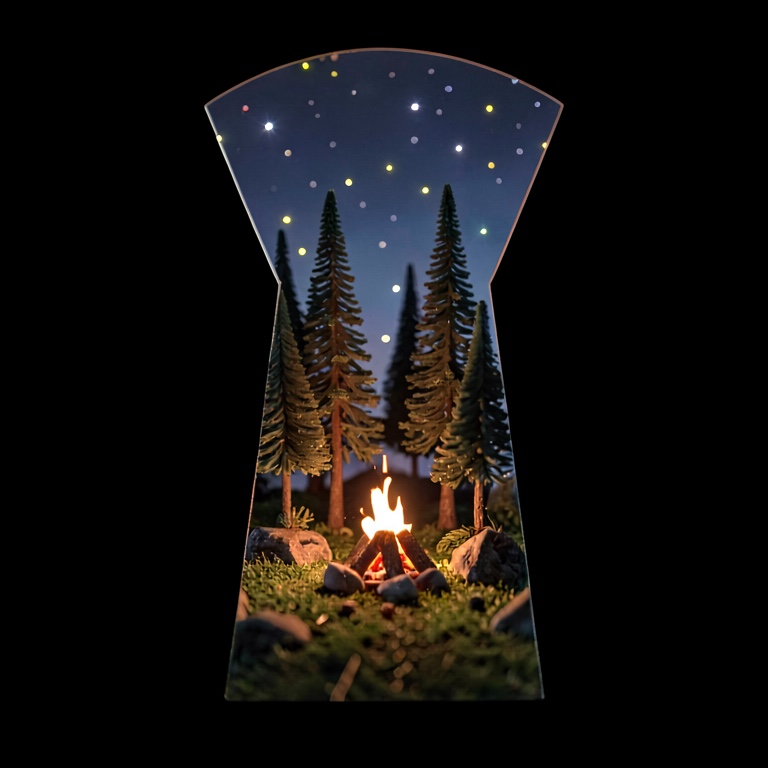} &
\includegraphics[width=0.21\linewidth]{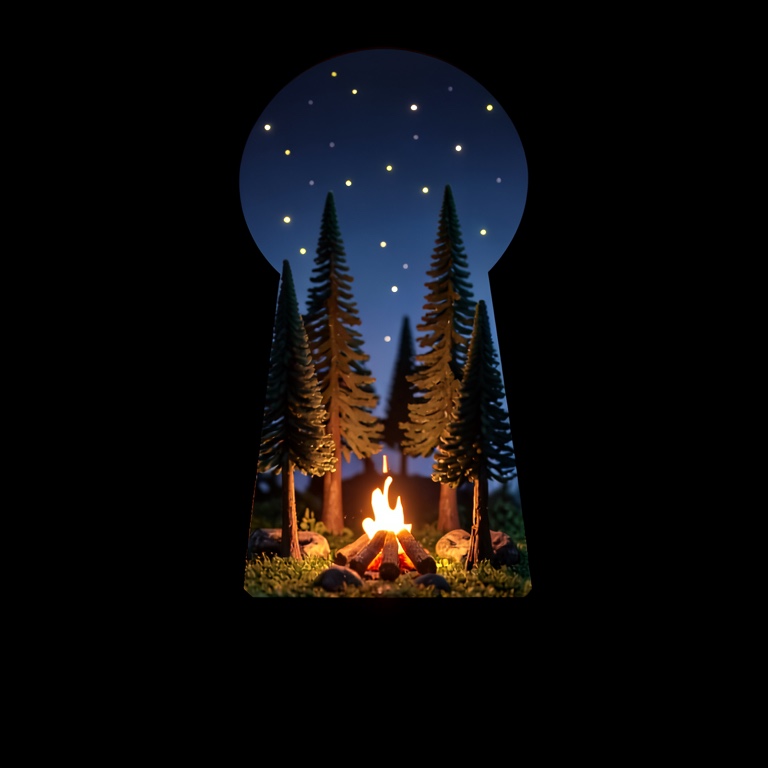} &
\includegraphics[width=0.21\linewidth]{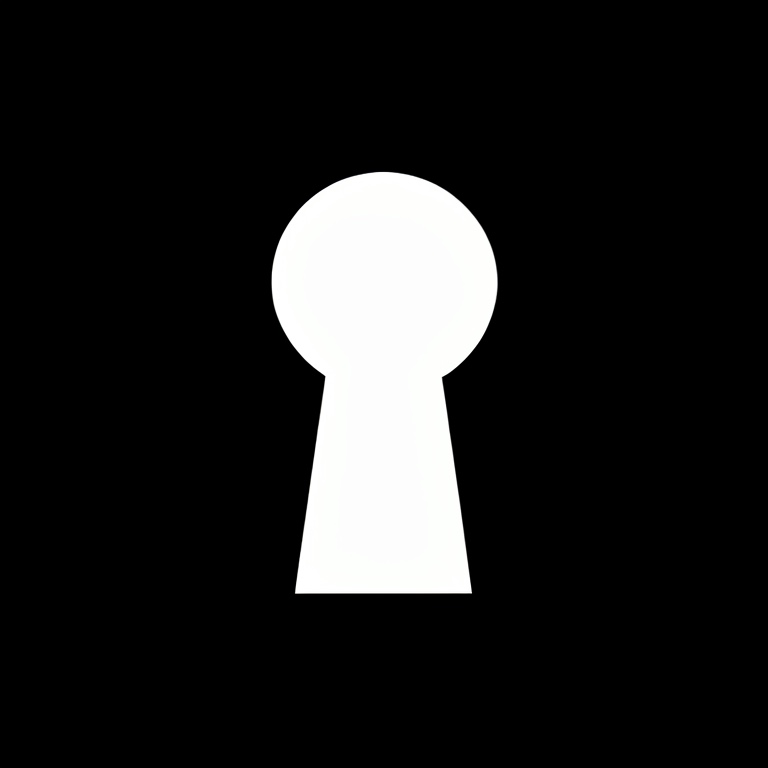} \\
\multicolumn{3}{c}{\small\textbf{Keyhole shape control}} \\
\multicolumn{3}{c}{
\parbox{0.67\linewidth}{\centering\small\textit{
``a miniature forest with tall pine trees, a glowing campfire, and fireflies drifting in the night sky, all inside a keyhole on a black background''}}
} \\[0.6em]
\includegraphics[width=0.21\linewidth]{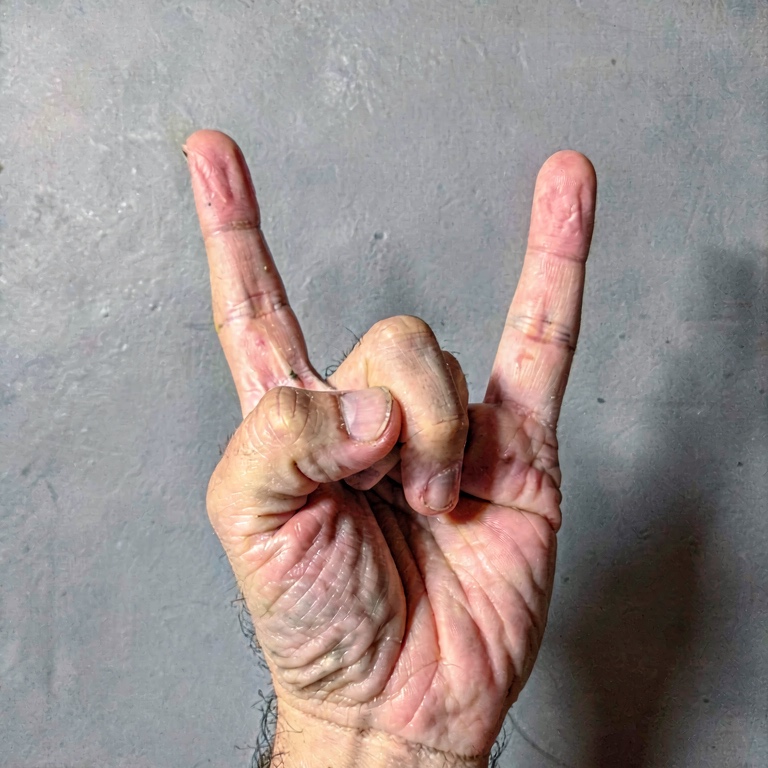} &
\includegraphics[width=0.21\linewidth]{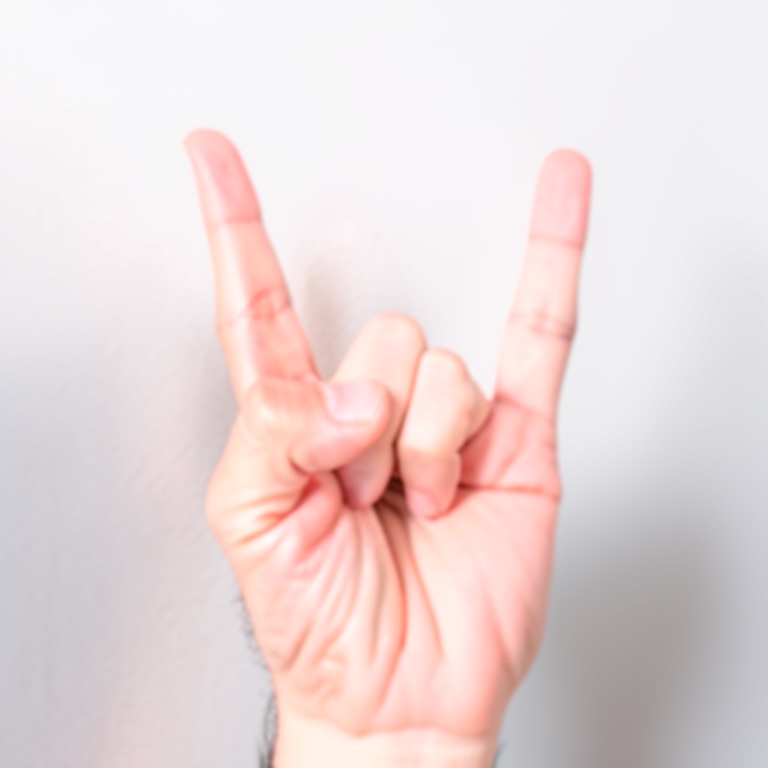} &
\includegraphics[width=0.21\linewidth]{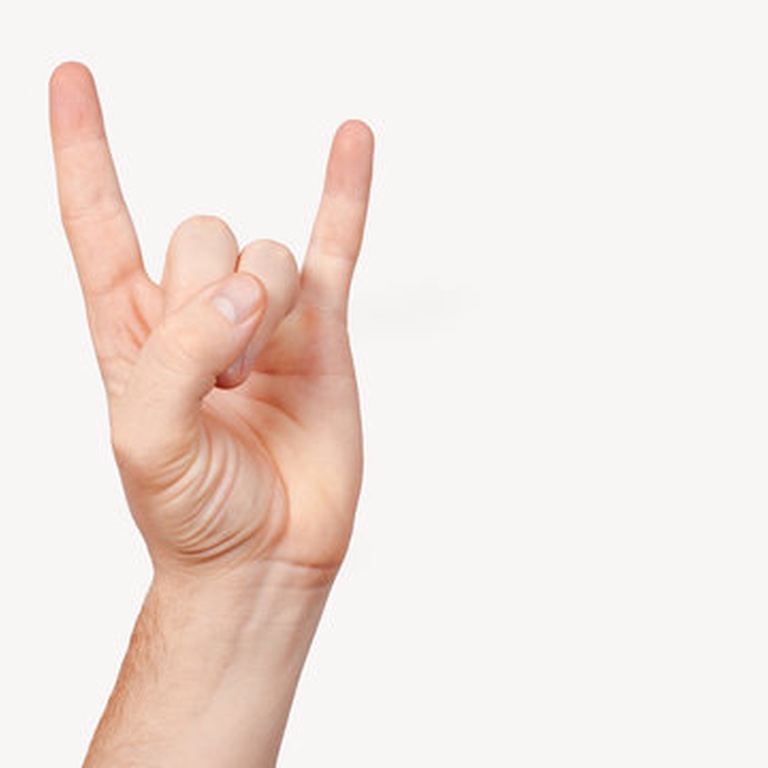} \\
\multicolumn{3}{c}{\small\textbf{Hand pose control}} \\ \multicolumn{3}{c}{
\parbox{0.67\linewidth}{\centering\small\textit{
``a hand doing the sign of the horns''}}
} \\[0.6em]
\includegraphics[width=0.21\linewidth]{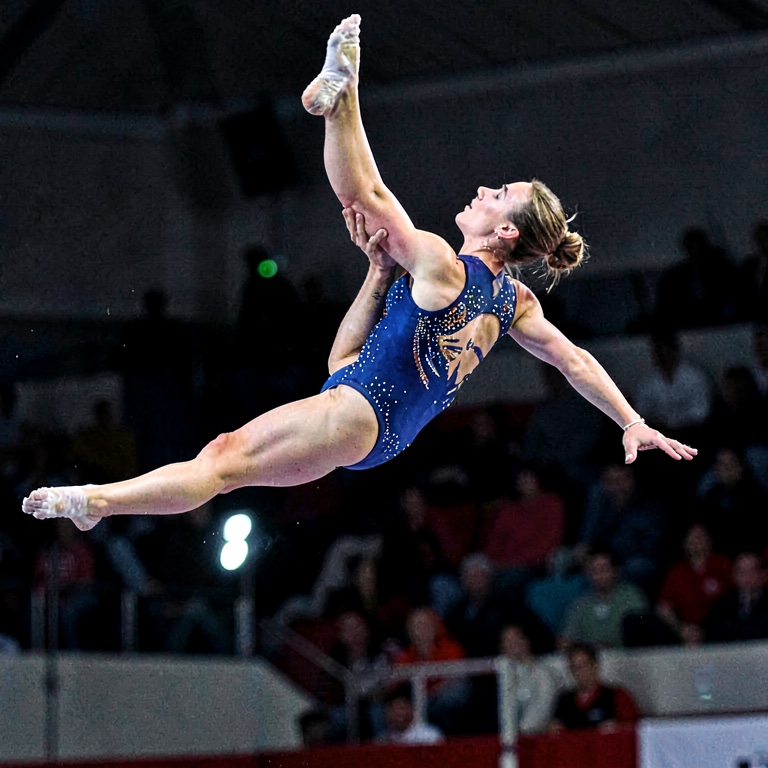} &
\includegraphics[width=0.21\linewidth]{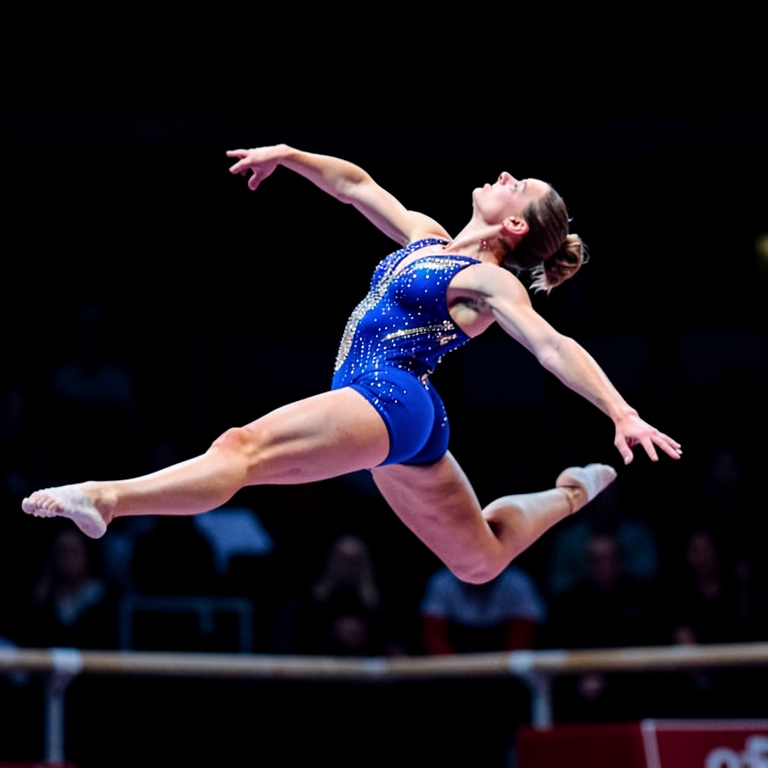} &
\includegraphics[width=0.21\linewidth]{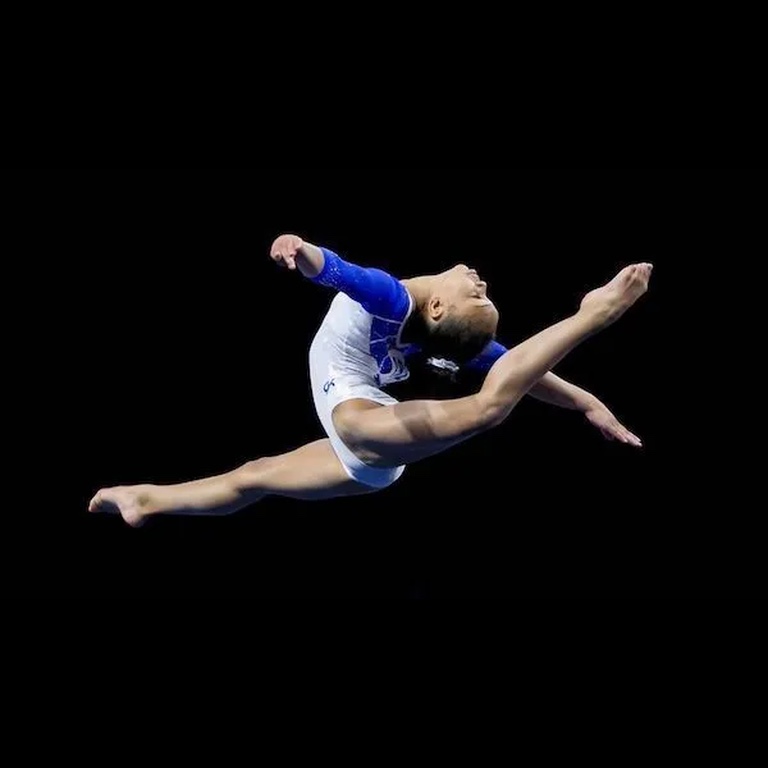} \\
\multicolumn{3}{c}{\small\textbf{Gymnastics pose control}} \\ \multicolumn{3}{c}{
\parbox{0.67\linewidth}{\centering\small\textit{
``a gymnast performing a ring leap, full body visible, airborne, one leg extended forward, the back leg bent high behind the head, arched back, pointed toes, arms extended, dynamic sports photograph''}}
} \\
\end{tabular}
}
\caption{\textbf{Small reference sets transfer coarse structure such as shape
and pose.} The closest reference (right) separates transferred structure from
copied content. The keyhole correction reshapes the silhouette and preserves
the interior scene.}
\label{fig:flux_structural_control}
\end{figure}

\paragraph{Structure.}
Structure is harder to steer with rewards or gradients because, unlike color
and style, it has no simple scalar proxy. Indeed, even strong vision-language
models struggle to judge whether a silhouette matches a shape, a hand is
correctly oriented, or limbs are ordered correctly in
depth~\citep{WolDan_Your_MICCAI2025}. A reference set sidesteps this problem,
because it specifies the target structure by example rather than through a
score. We test this in three selected cases, a keyhole silhouette, a hand
making the sign of the horns, and a gymnast performing a ring leap. In each
case RMG moves the output toward the target structure
(\Cref{fig:flux_structural_control}), with a small bank supplying the
structural prior without gradients, retraining, or extra model evaluations.

\begin{findingbox}
A frozen text-to-image model can be steered by examples alone. Swapping a
small image bank changes color, identity, style, and coarse structure without
retraining.
\end{findingbox}

\subsection{Q3: Comparison with Search- and Gradient-Based Control}
\label{sec:geneval_comparison}

\begin{questionbox}
How does a small reference bank compare with search- and gradient-based
control?
\end{questionbox}

We generate four images for each of the 553 prompts of GenEval, which covers
single objects, two objects, counting, colors, positions, and color
attribution. Each baseline expresses the constraint through its native
interface, whereas RMG receives a 20-image bank for each prompt. For single
objects and colors, this bank holds direct examples of the target. For the
compositional categories, it is instead assembled from simpler components,
and the position and counting banks are tiled into synthetic layouts
(\Cref{app:geneval_banks}). RMG therefore has access to visual information
that the baselines lack, so the comparison measures what a small bank buys
rather than which algorithm makes better use of the same information.

\begin{table}[t]
\centering
\footnotesize
\caption{\textbf{A single guided trajectory improves compositional alignment
more than search- or gradient-based control.} Time is wall-clock runtime
relative to the base model, excluding bank preparation for RMG
(\Cref{app:runtime_benchmark}). NFE is normalized by one generation, and
Aux.\ counts external-model calls (C~=~classifier, L~=~LLM). Bold marks the
best text-only method. RMG also receives a visual bank and is listed
separately.}
\label{tab:flux_comparison}
\setlength{\tabcolsep}{3.2pt}
\begin{tabular}{@{}lcccccccccc@{}}
\toprule
& \multicolumn{3}{c}{Cost $\downarrow$} & \multicolumn{7}{c}{GenEval accuracy $\uparrow$} \\
\cmidrule(lr){2-4}\cmidrule(l){5-11}
Method & Time & NFE & Aux. & Mean & Single & Two & Count & Colors & Pos. & Attr. \\
\midrule
\mbox{FLUX.2-klein} (4B) & $1.00\times$ & $\mathbf{1\times}$ & \textbf{--} & 80.10 & 99.69 & 91.41 & 80.62 & 84.84 & 65.25 & 58.75 \\
\multicolumn{11}{@{}l}{\textit{Search-based}} \\
\hspace{0.6em}+ Prompt Opt.~\citep{manas2024improvingtexttoimageconsistencyautomatic} & $7.87\times$ & $8\times$ & 8C+2L & 84.18 & \textbf{100.00} & 95.45 & \textbf{88.12} & 87.77 & 69.75 & 64.00 \\
\hspace{0.6em}+ Best-of-4~\citep{karthik2023dontsucceedtrytry} & $4.07\times$ & $4\times$ & 4C & 83.35 & 99.69 & 95.96 & 83.44 & 88.03 & 67.75 & 65.25 \\
\hspace{0.6em}+ SMC~\citep{wu2024practicalasymptoticallyexactconditional} & $6.17\times$ & $4\times$ & 81C & 80.28 & 99.69 & 95.71 & 81.88 & 85.37 & 61.75 & 57.25 \\
\multicolumn{11}{@{}l}{\textit{Gradient-based}} \\
\hspace{0.6em}+ ReNO~\citep{eyring2024renoenhancingonesteptexttoimage} & $19.44\times$ & $4\times$ & 4C & 83.46 & 99.69 & 93.18 & 87.50 & 90.16 & 65.50 & 64.75 \\
\midrule
\multicolumn{11}{@{}l}{\textit{Reference-based}} \\
\rowcolor{green!12}
\hspace{0.6em}+ RMG (ours) & $\mathbf{1.02\times}$ & $\mathbf{1\times}$ & \textbf{--} & \textbf{91.17} & \textbf{100.00} & \textbf{99.49} & \textbf{88.12} & \textbf{90.16} & \textbf{94.00} & \textbf{75.25} \\
\bottomrule
\end{tabular}
\end{table}

Without a
classifier, reward model, LLM, gradient, or candidate selection, RMG raises
the mean GenEval score from 80.10 to 91.17 at $1.02\times$ the runtime of the
unguided model, whereas the baselines reach at most 84.18 at $4\times$ to
$19\times$ the runtime (\Cref{tab:flux_comparison}). The gains are largest on
position ($+28.75$) and two-object prompts ($+8.08$), precisely the
categories in which the base model follows the text least reliably.

\begin{findingbox}
A handful of examples can replace search and gradients as a control signal.
One guided trajectory with a small bank improves alignment more than methods
that cost four to nineteen times as much.
\end{findingbox}

\subsection{Q4: Learning to Use the Reference Set}
\label{sec:spfm}
\label{sec:spg_results}

\begin{questionbox}
Can a trained model learn to use reference sets without losing quality?
\end{questionbox}

\begin{figure}[ht]
\centering

\begin{minipage}[t]{0.495\linewidth}
    \centering
    {\small \textbf{Generated (full reference set)}}

    \vspace{0.2em}

    \includegraphics[width=0.19\linewidth]{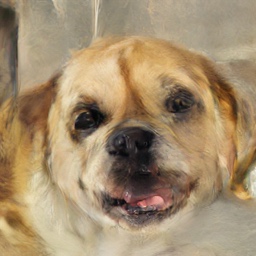}%
    \includegraphics[width=0.19\linewidth]{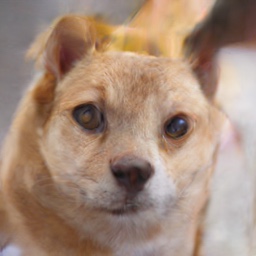}%
    \includegraphics[width=0.19\linewidth]{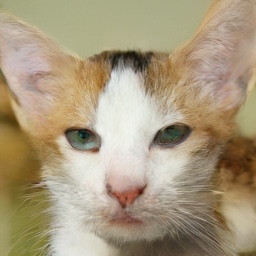}%
    \includegraphics[width=0.19\linewidth]{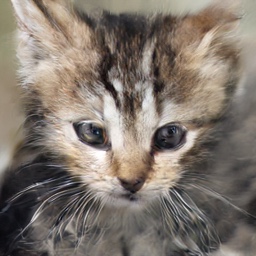}%
    \includegraphics[width=0.19\linewidth]{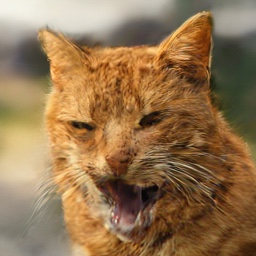}
\end{minipage}%
\hfill%
\begin{minipage}[t]{0.495\linewidth}
    \centering
    {\small \textbf{Cat-only reference set}}

    \vspace{0.2em}

    \includegraphics[width=0.19\linewidth]{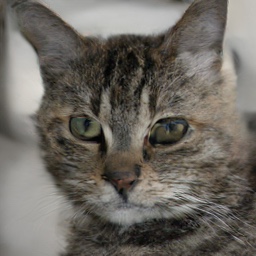}%
    \includegraphics[width=0.19\linewidth]{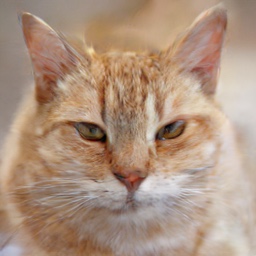}%
    \includegraphics[width=0.19\linewidth]{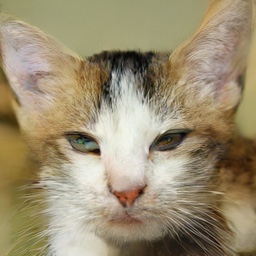}%
    \includegraphics[width=0.19\linewidth]{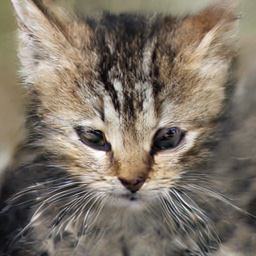}%
    \includegraphics[width=0.19\linewidth]{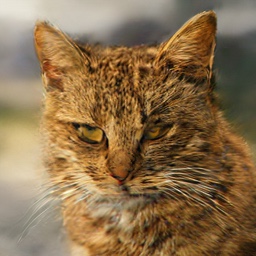}
\end{minipage}

\vspace{0.55em}

\begin{minipage}[t]{0.495\linewidth}
    \centering
    {\small \textbf{Nearest neighbors (latent space)}}

    \vspace{0.2em}

    \includegraphics[width=0.19\linewidth]{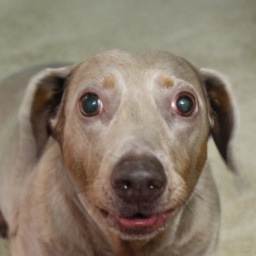}%
    \includegraphics[width=0.19\linewidth]{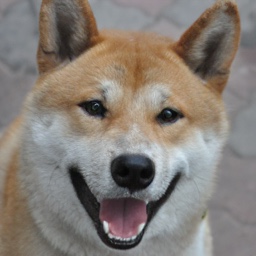}%
    \includegraphics[width=0.19\linewidth]{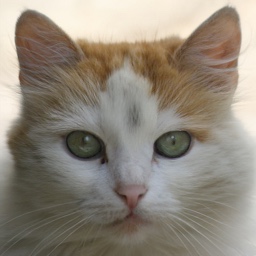}%
    \includegraphics[width=0.19\linewidth]{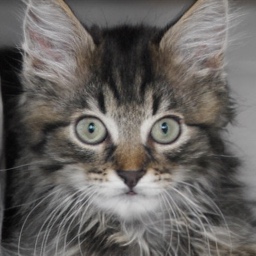}%
    \includegraphics[width=0.19\linewidth]{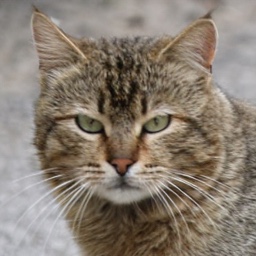}
\end{minipage}%
\hfill%
\begin{minipage}[t]{0.495\linewidth}
    \centering
    {\small \textbf{Dog-only reference set}}

    \vspace{0.2em}

    \includegraphics[width=0.19\linewidth]{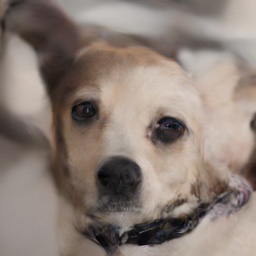}%
    \includegraphics[width=0.19\linewidth]{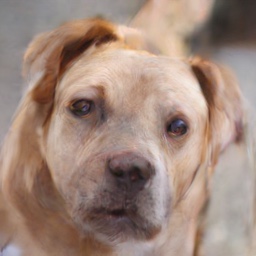}%
    \includegraphics[width=0.19\linewidth]{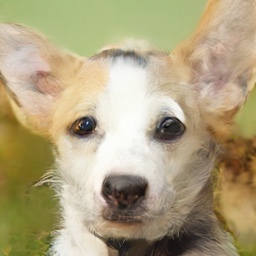}%
    \includegraphics[width=0.19\linewidth]{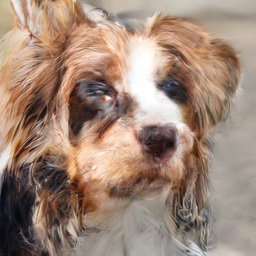}%
    \includegraphics[width=0.19\linewidth]{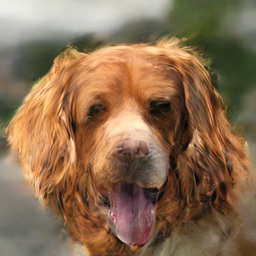}
\end{minipage}

\caption{\textbf{SPG samples resemble their references without copying them,
and swapping the bank changes the output.} \textbf{Left:} samples with the
full reference set and their nearest neighbors in latent space.
\textbf{Right:} the same noise seeds with a cat-only and a dog-only bank.}
\label{fig:spg_qual}
\vspace{-1em}
\end{figure}

We first check that conditioning on a reference set does not cost generation
quality. Against a DiT-B/4 trained on the same data, SPG reaches an FID of
23.26 on AFHQv2, compared with 23.11 for the baseline (\Cref{fig:main_block}a).
Its samples are semantically close to their nearest references in latent
space yet visibly distinct from them (\Cref{fig:spg_qual}), so the reference
set acts as a soft conditioning signal rather than as retrieval. We then test
whether the reference set controls what SPG generates. With the same noise
seeds, a cat-only bank yields cats and a dog-only bank yields dogs
(\Cref{fig:spg_qual}, right), and over 10{,}000 CLIP-labeled samples the
generated class proportions follow the composition of the bank for every
reference size from $M=10$ to $M=5{,}000$ (\Cref{fig:main_block}b).

\begin{figure}[ht]
\centering

\setlength{\tabcolsep}{4pt} %
\renewcommand{\arraystretch}{1.0}

\begin{tabular}{@{}cc@{}}

\begin{minipage}{0.42\linewidth}
\centering
\begin{tabular}{lccc}
\toprule
Model & FID $\downarrow$ & KID $\downarrow$ & IS $\uparrow$ \\
\midrule
DiT-B/4 & 23.111 & 0.012 & 6.554 \\
SPG     & 23.256 & 0.013 & 6.227 \\
\bottomrule
\end{tabular}
\end{minipage}
&
\begin{minipage}{0.45\linewidth}
\centering
\begin{tikzpicture}
\begin{axis}[
  width=0.78\linewidth,
  height=0.72\linewidth,
  xmin=-2, xmax=102,
  ymin=-2, ymax=102,
  xtick={0,20,...,100},
  ytick={0,20,...,100},
  tick label style={font=\small},
  xlabel={Cat \% in database},
  ylabel={Generated cat \%},
  xlabel style={font=\small, yshift=4pt},
  ylabel style={font=\small, yshift=-4pt},
  grid=both,
  grid style={dashed, gray!40},
  colormap/viridis,
  point meta min=0,
  point meta max=1,
  colorbar,
  colorbar style={
    width=6pt,
    xshift=-8pt,
    ytick={0, 0.1667, 0.3333, 0.5, 0.6667, 0.8333, 1.0},
    yticklabels={10, 50, 100, 200, 500, 1000, 5000},
    ylabel={Reference size $M$},
    ylabel style={font=\small},
    tick label style={font=\small},
  },
]
\addplot[dashed, line width=1.5pt, color=gray!60, forget plot]
  coordinates {(0,0)(100,100)}
  node[pos=0.55, below right, font=\small, text=gray!60] {$y=x$};
\addplot[color={rgb,255:red,68;green,1;blue,84}, line width=2pt, mark=none, forget plot]
  coordinates {(100,100.00)(90,97.80)(80,97.30)(70,92.10)(60,90.80)(50,73.70)(40,56.50)(30,33.10)(20,17.10)(10,16.40)(0,16.20)};
\addplot[color={rgb,255:red,68;green,57;blue,131}, line width=2pt, mark=none, forget plot]
  coordinates {(100,98.40)(90,97.20)(80,91.80)(70,80.10)(60,70.90)(50,60.40)(40,43.50)(30,31.80)(20,23.50)(10,19.60)(0,15.00)};
\addplot[color={rgb,255:red,49;green,104;blue,142}, line width=2pt, mark=none, forget plot]
  coordinates {(100,96.90)(90,94.20)(80,86.80)(70,75.20)(60,65.30)(50,56.40)(40,43.70)(30,40.10)(20,32.10)(10,23.40)(0,17.60)};
\addplot[color={rgb,255:red,33;green,145;blue,140}, line width=2pt, mark=none, forget plot]
  coordinates {(100,97.10)(90,94.30)(80,88.10)(70,82.60)(60,73.80)(50,59.30)(40,49.00)(30,42.40)(20,27.10)(10,21.00)(0,13.40)};
\addplot[color={rgb,255:red,53;green,183;blue,121}, line width=2pt, mark=none, forget plot]
  coordinates {(100,98.00)(90,94.00)(80,89.50)(70,81.00)(60,72.70)(50,64.10)(40,52.90)(30,41.20)(20,28.70)(10,21.20)(0,14.90)};
\addplot[color={rgb,255:red,144;green,215;blue,67}, line width=2pt, mark=none, forget plot]
  coordinates {(100,98.00)(90,94.80)(80,89.10)(70,82.50)(60,72.20)(50,60.90)(40,52.20)(30,40.50)(20,30.90)(10,21.40)(0,14.90)};
\addplot[color={rgb,255:red,253;green,231;blue,37}, line width=2pt, mark=none, forget plot]
  coordinates {(100,98.60)(90,95.70)(80,89.40)(70,81.00)(60,73.10)(50,63.60)(40,51.20)(30,40.70)(20,32.40)(10,23.60)(0,16.30)};
\end{axis}
\end{tikzpicture}
\end{minipage}

\\[-0.2em]

{\small (a) Unconditional generation quality}
&
{\small (b) Control via reference-set composition}

\end{tabular}

\caption{\textbf{SPG keeps unconditional quality, and the composition of the
bank sets the output distribution.} \textbf{(a)} Generation quality on
AFHQv2. \textbf{(b)} Generated against reference cat percentage for several
reference sizes $M$, each point from 10{,}000 CLIP-labeled samples.}
\label{fig:main_block}

\end{figure}

\begin{findingbox}
Reference conditioning can be learned without giving up quality. SPG
matches an unconditional DiT, yet its bank sets what it generates.
\end{findingbox}

\paragraph{Further results.}
\Cref{app:additional_experiments} reports additional experiments for each
question, and \Cref{app:ablations} ablates the schedule, its strength, the
reference-set size, the bank quality, and the number of steps.

\section{Related Work}
\label{sec:related_work}

Pretrained generative models are commonly controlled by
fine-tuning~\citep{ruiz2023dreambooth,hu2022lora}, by inference-time guidance
from auxiliary models or reward
signals~\citep{dhariwal2021diffusion,eyring2024renoenhancingonesteptexttoimage},
or by search over
samples~\citep{karthik2023dontsucceedtrytry,wu2024practicalasymptoticallyexactconditional,manas2024improvingtexttoimageconsistencyautomatic},
which trades efficiency for quality. Recent work on endpoint
posteriors~\citep{potaptchik2025tilt,sabour2025testtime,holderrieth2026diamond}
shares our view that endpoint information governs controllable generation,
and steers it through scalar rewards that require training or repeated
evaluation. Retrieval-augmented
methods~\citep{lewis2020retrieval,borgeaud2022improving,blattmann2022retrieval}
condition generation on external data and treat the retrieved content as
auxiliary context, whereas we use it as the control signal itself. Closest to
our setting, ControlNet, T2I-Adapter, and IP-Adapter accept spatial or image
conditions through modules trained for a particular modality and
backbone~\citep{zhang2023adding,mou2024t2i,ye2023ip}. RMG needs no trained
module, because the reference set defines the endpoint posterior mean
directly and yields a closed-form drift correction. The two approaches are
complementary, since a trained module suits explicit spatial conditions and
RMG, which changes only the sampling velocity, could be combined with one. Under a Gaussian bridge, the endpoint posterior
mean is exactly a softmax-weighted aggregation over reference points, which
grounds attention as a conditional expectation and connects our approach to
non-parametric score
estimation~\citep{niedoba2024nearest,scarvelis2025closedform}.
\Cref{app:related_work} gives more detail.

\section{Limitations}
\label{sec:limitations}

Reference-mean guidance inherits the quality of its reference set. Whatever
the references have in common is transferred, and degraded or irrelevant
banks reduce diversity or give no control (\Cref{app:bank_quality}). On
GenEval, RMG retains 71\% of the within-prompt diversity of the unguided
model and lowers the mean ImageReward from 1.214 to 1.164, a change
concentrated in the two categories whose banks were built from tiled layouts
(\Cref{app:geneval_tradeoffs}). The scope of our evaluation is also limited
in several ways. The GenEval baselines do not receive the visual bank, the
structural results are selected qualitative examples of coarse control, all
large-model experiments use a single FLUX backbone, and SPG is evaluated on
AFHQv2. \Cref{app:limitations} discusses these points, the assumptions behind
Proposition~3.3, and possible mitigations. As with any controllability
method, reference sets should be curated responsibly to prevent harmful or
misleading generation.

\section{Conclusion}
\label{sec:conclusion}

We have shown that control can be framed as a problem of shifting endpoint means. This view offers a simple alternative to fine-tuning, auxiliary guidance, and search, in which generation is steered by changing the reference set over which the model aggregates, implicitly or explicitly. Reference-Mean Guidance shows that the principle already works in frozen pretrained models given a well-chosen reference set, and Semi-Parametric Guidance builds the same mechanism into a trained architecture that learns which parts of the references to transfer while keeping its generation quality. Together, they suggest a path toward generative models that adapt through data rather than through parameter updates.

\clearpage
\begin{ack}
This project was supported by the ELLIS Unit Amsterdam, by the Bosch Center for Artificial Intelligence and carried out using the Dutch national e-infrastructure,
with the support of SURF through the use of the Snellius
supercomputer. MZ acknowledges support from Microsoft Research AI4Science. JWvdM acknowledges support from the European Union Horizon Framework Programme (Grant agreement ID: 101120237)
\end{ack}

\medskip

{
\small
\bibliographystyle{unsrtnat}
\bibliography{references}
}

\appendix
\makeatletter
\setlength{\@fptop}{0pt}
\setlength{\@fpsep}{14pt plus 4pt}
\setlength{\@fpbot}{0pt plus 1fil}
\makeatother
\let\appendixsection\section
\renewcommand{\section}{\clearpage\suppressfloats[t]\appendixsection}
\let\appendixsubsection\subsection
\renewcommand{\subsection}{\FloatBarrier\appendixsubsection}
\section*{Appendix Index}
\addcontentsline{toc}{section}{Appendix Index}

\noindent
\hyperref[app:proofs]{\ref*{app:proofs}\quad Proofs and Derivations}\par
\hspace*{1.5em}\hyperref[app:proof_prop1]{\ref*{app:proof_prop1}\quad Proof of Proposition 3.1}\par
\hspace*{1.5em}\hyperref[app:proof_prop2]{\ref*{app:proof_prop2}\quad Proof of Proposition 3.2}\par
\hspace*{1.5em}\hyperref[app:proof_thm3]{\ref*{app:proof_thm3}\quad Proof of Proposition 3.3}\par
\hspace*{1.5em}\hyperref[app:reference_sets]{\ref*{app:reference_sets}\quad Reference-Set Formalism}\par

\hyperref[app:related_work]{\ref*{app:related_work}\quad Related Work}\par

\hyperref[app:experiments]{\ref*{app:experiments}\quad Experimental Details and Additional Results}\par
\hspace*{1.5em}\hyperref[app:mechanistic_validation]{\ref*{app:mechanistic_validation}\quad Q1: Mechanistic Validation}\par
\hspace*{1.5em}\hyperref[app:flux_experiments]{\ref*{app:flux_experiments}\quad Q2: Training-Free Control of a Frozen Model}\par
\hspace*{3em}\hyperref[app:flux_sampling]{\ref*{app:flux_sampling}\quad Guidance Setup}\par
\hspace*{3em}\hyperref[app:prompt_reference_interaction]{\ref*{app:prompt_reference_interaction}\quad Prompt--Reference Interaction}\par
\hspace*{3em}\hyperref[app:reference_composition]{\ref*{app:reference_composition}\quad Reference Composition}\par
\hspace*{1.5em}\hyperref[app:geneval]{\ref*{app:geneval}\quad Q3: GenEval Comparison}\par
\hspace*{3em}\hyperref[app:geneval_banks]{\ref*{app:geneval_banks}\quad Bank Construction}\par
\hspace*{3em}\hyperref[app:geneval_evaluation_protocol]{\ref*{app:geneval_evaluation_protocol}\quad Evaluation Metrics and Runtime}\par
\hspace*{3em}\hyperref[app:geneval_tradeoffs]{\ref*{app:geneval_tradeoffs}\quad Quality, Diversity, and Reference Similarity}\par
\hspace*{3em}\hyperref[app:quality_categories]{\ref*{app:quality_categories}\quad ImageReward by Category}\par
\hspace*{1.5em}\hyperref[app:spg]{\ref*{app:spg}\quad Q4: Semi-Parametric Guidance}\par
\hspace*{3em}\hyperref[app:spg_training]{\ref*{app:spg_training}\quad Architecture and Training}\par
\hspace*{3em}\hyperref[app:spg_setup]{\ref*{app:spg_setup}\quad AFHQv2 Setup}\par
\hspace*{3em}\hyperref[app:spg_diversity]{\ref*{app:spg_diversity}\quad Diversity and Reference-Set Size}\par
\hspace*{3em}\hyperref[app:spg_background_nocopy]{\ref*{app:spg_background_nocopy}\quad Suppressing Nuisance Attributes}\par

\hyperref[app:ablations]{\ref*{app:ablations}\quad Ablations}\par
\hspace*{1.5em}\hyperref[app:schedule_ablation]{\ref*{app:schedule_ablation}\quad Guidance Schedule}\par
\hspace*{1.5em}\hyperref[app:beta0_sweep]{\ref*{app:beta0_sweep}\quad Guidance Strength and Schedule Shape}\par
\hspace*{1.5em}\hyperref[app:dataset_size_ablation]{\ref*{app:dataset_size_ablation}\quad Reference-Set Size}\par
\hspace*{1.5em}\hyperref[app:bank_quality]{\ref*{app:bank_quality}\quad Reference-Bank Quality}\par
\hspace*{1.5em}\hyperref[app:nfe_ablation]{\ref*{app:nfe_ablation}\quad Number of Function Evaluations}\par

\hyperref[app:limitations]{\ref*{app:limitations}\quad Extended Discussion of Limitations}\par

\hyperref[app:flux_banks]{\ref*{app:flux_banks}\quad Reference Banks}\par

\section{Proofs and Derivations}
\label{app:proofs}

This appendix gives full derivations of the three results in \Cref{sec:theory}. The main text works with the linear bridge for readability. Here we treat the general affine bridge
\[
I_t(x_0,x_1) := \alpha_t x_0 + \beta_t x_1,
\]
where $\alpha_t$ and $\beta_t$ are differentiable scalar schedules with $\alpha_0=1$, $\beta_0=0$, $\alpha_1=0$, $\beta_1=1$, and $\alpha_t>0$ for $t\in[0,1)$. We write
\[
a_t := \frac{\dot{\alpha}_t}{\alpha_t},
\qquad
c_t := \dot{\beta}_t - \beta_t\frac{\dot{\alpha}_t}{\alpha_t}.
\]
The linear bridge of the main text is the special case $\alpha_t=1-t$ and $\beta_t=t$, for which $a_t=-1/(1-t)$ and $c_t=1/(1-t)$. To keep $\alpha_t$ and $\beta_t$ free for the bridge coefficients, we denote the guidance schedule by $\gamma_t$ throughout.

\subsection{Proof of Proposition 3.1 (Optimal Drift)}
\label{app:proof_prop1}

We derive the optimal flow-matching velocity under the affine bridge, with $x_0 \sim p_0$ and $x_1 \sim p_1$ drawn independently.

\paragraph{Step 1: The bridge velocity.}
Differentiating the bridge with respect to $t$ gives
\[
\dot{x}_t = \dot{\alpha}_t x_0 + \dot{\beta}_t x_1,
\]
which for the linear bridge is simply the displacement $x_1-x_0$.

\paragraph{Step 2: The marginal velocity field.}
The flow-matching objective regresses a velocity field $u_t^\theta(x)$ onto the conditional velocity $\dot{x}_t$. Many endpoint pairs $(x_0, x_1)$ pass through the same intermediate state, so the minimizer of this regression is the conditional expectation
\begin{equation}
\label{app:eq:conditional_velocity}
u_t(x) = \mathbb{E}[\dot{x}_t \mid x_t = x] = \mathbb{E}[\dot{\alpha}_t x_0 + \dot{\beta}_t x_1 \mid x_t = x].
\end{equation}

\paragraph{Step 3: Eliminating $x_0$.}
For $t<1$ the bridge can be solved for the source point,
\[
x_0 = \frac{x_t - \beta_t x_1}{\alpha_t},
\]
and substituting into \Cref{app:eq:conditional_velocity} gives
\begin{align*}
u_t(x)
&= \mathbb{E}\!\left[\dot{\alpha}_t\frac{x_t-\beta_t x_1}{\alpha_t} + \dot{\beta}_t x_1 \;\middle|\; x_t = x\right] \\
&= \mathbb{E}\!\left[\frac{\dot{\alpha}_t}{\alpha_t}x_t
 + \left(\dot{\beta}_t - \beta_t\frac{\dot{\alpha}_t}{\alpha_t}\right)x_1
 \;\middle|\; x_t = x\right].
\end{align*}

\paragraph{Step 4: The conditional endpoint mean.}
Conditioning on $x_t = x$ fixes the first term, and the second becomes the endpoint mean $\mu_t(x) := \mathbb{E}[x_1 \mid x_t = x]$. Hence
\[
u_t(x)
= \frac{\dot{\alpha}_t}{\alpha_t}x
+ \left(\dot{\beta}_t - \beta_t\frac{\dot{\alpha}_t}{\alpha_t}\right)\mu_t(x)
= a_t x + c_t\mu_t(x),
\]
which for the linear bridge reads $u_t(x)=(\mu_t(x)-x)/(1-t)$. \hfill$\square$

\paragraph{Remark.} The coefficients $a_t$ and $c_t$ diverge as $\alpha_t\to0$ at $t=1$. For the linear bridge this is the familiar factor $(1-t)^{-1}$. Numerical integration stops short of $t=1$, so the singularity is never evaluated in practice.

\subsection{Proof of Proposition 3.2 (Posterior Mean as Cross-Attention)}
\label{app:proof_prop2}

We now take a standard normal source $p_0 = \mathcal{N}(0,I)$ and the empirical target $\hat{p}_1$ over the training set $\mathcal{D} = \{x^{(n)}\}_{n=1}^N$, derive the endpoint mean in closed form, and show that it takes the form of cross-attention over the dataset.

\paragraph{Step 1: The bridge conditional.}
With $x_0 \sim \mathcal{N}(0,I)$ and the endpoint fixed at $x^{(n)}$, the state $x_t = \alpha_t x_0 + \beta_t x^{(n)}$ is an affine function of a Gaussian and is therefore Gaussian itself,
\[
x_t \mid x^{(n)} \sim \mathcal{N}\!\bigl(\beta_t x^{(n)},\, \alpha_t^2 I\bigr).
\]
Its density at $x_t=x$ is
\begin{equation}
\label{app:eq:bridge_density}
p_t(x \mid x^{(n)})
= \frac{1}{(2\pi)^{d/2}\alpha_t^d}
\exp\!\left(-\frac{\|x - \beta_t x^{(n)}\|^2}{2\alpha_t^2}\right).
\end{equation}

\paragraph{Step 2: Posterior weights.}
Under $\hat{p}_1$ every data point has prior probability $1/N$. Bayes' rule gives the posterior probability that a trajectory passing through $x$ at time $t$ ends at $x^{(n)}$,
\[
w_t^{(n)}(x)
:= p\!\left(x_1 = x^{(n)} \mid x_t=x\right)
= \frac{p_t(x \mid x^{(n)}) \cdot \frac{1}{N}}{\sum_{j=1}^N p_t(x \mid x^{(j)}) \cdot \frac{1}{N}}
= \frac{p_t(x \mid x^{(n)})}{\sum_{j=1}^N p_t(x \mid x^{(j)})}.
\]
The uniform prior cancels, and so does the normalizing constant $(2\pi)^{d/2}\alpha_t^d$ of \Cref{app:eq:bridge_density}, leaving
\begin{equation}
\label{app:eq:raw_weights}
w_t^{(n)}(x)
=
\frac{\exp\!\left(-\dfrac{\|x - \beta_t x^{(n)}\|^2}{2\alpha_t^2}\right)}
{\displaystyle\sum_{j=1}^N \exp\!\left(-\dfrac{\|x - \beta_t x^{(j)}\|^2}{2\alpha_t^2}\right)}.
\end{equation}

\paragraph{Step 3: The endpoint mean.}
The conditional endpoint mean is the posterior expectation of $x_1$,
\begin{equation}
\label{app:eq:posterior_mean}
\hat{\mu}_t(x) = \mathbb{E}_{\hat{p}_1}[x_1 \mid x_t=x] = \sum_{n=1}^N w_t^{(n)}(x)\, x^{(n)}.
\end{equation}

\paragraph{Step 4: Expanding the exponent.}
Expanding the squared norm in the exponent gives
\[
\|x - \beta_t x^{(n)}\|^2
= \|x\|^2 - 2\beta_t\langle x, x^{(n)}\rangle + \beta_t^2\|x^{(n)}\|^2.
\]
The term $\|x\|^2$ does not depend on $n$, so it scales every term of the softmax by the same factor and cancels:
\[
\exp\!\left(-\frac{\|x - \beta_t x^{(n)}\|^2}{2\alpha_t^2}\right)
\propto
\exp\!\left(\frac{\beta_t\langle x, x^{(n)}\rangle}{\alpha_t^2} - \frac{\beta_t^2\|x^{(n)}\|^2}{2\alpha_t^2}\right),
\]
where $\propto$ hides a factor independent of $n$. The weights therefore become
\begin{equation}
\label{app:eq:simplified_weights}
w_t^{(n)}(x)
=
\frac{\exp\!\left(\dfrac{\beta_t}{\alpha_t^2}\langle x, x^{(n)}\rangle - \dfrac{\beta_t^2}{2\alpha_t^2}\|x^{(n)}\|^2\right)}
{\displaystyle\sum_{j=1}^N \exp\!\left(\dfrac{\beta_t}{\alpha_t^2}\langle x, x^{(j)}\rangle - \dfrac{\beta_t^2}{2\alpha_t^2}\|x^{(j)}\|^2\right)}.
\end{equation}

\paragraph{Step 5: Cross-attention.}
Define the query, keys, values, and biases
\[
q := \frac{\beta_t}{\alpha_t^2}x,
\qquad
k_n := x^{(n)},
\qquad
v_n := x^{(n)},
\qquad
b_n := -\frac{\beta_t^2}{2\alpha_t^2}\|x^{(n)}\|^2.
\]
The exponent in \Cref{app:eq:simplified_weights} is then $q^\top k_n + b_n$, so that
\[
w_t^{(n)}(x) = \frac{\exp(q^\top k_n + b_n)}{\sum_{j=1}^N \exp(q^\top k_j + b_j)} =: \alpha_n(q),
\qquad
\hat{\mu}_t(x) = \sum_{n=1}^N \alpha_n(q)\, v_n.
\]
This is cross-attention with query $q$, keys $\{k_n\}$, values $\{v_n\}$, and per-key biases $\{b_n\}$. \hfill$\square$

\paragraph{Remark on the bias term.} The bias $b_n$ penalizes data points with large norm. Standard dot-product attention has no such term. Here it comes from the Gaussian geometry of the bridge and acts as a length normalization on the keys. Setting $\alpha_t=1-t$ and $\beta_t=t$ recovers the weights of \Cref{eq:weights-softmax}.

\subsection{Proof of Proposition~3.3 (Reference-Mean Guided Dynamics)}
\label{app:proof_thm3}

We derive the guided velocity field induced by a geometric mixture of endpoint distributions. Under a shared-covariance Gaussian assumption on the endpoint posteriors, the correction depends only on a difference of posterior means. We then show that interpolating scores at the level of the noisy marginals leads to the same formula, and compare the assumptions that the two routes require.

\paragraph{Setup.}
Let $p_1$ and $\rho_1$ be the training and reference endpoint distributions, both with densities with respect to a common measure, and fix $\lambda \in [0,1]$. We assume that the endpoint-level geometric mixture
\[
\pi(x_1) = Z_\lambda^{-1}\, p_1(x_1)^{1-\lambda}\,\rho_1(x_1)^{\lambda}
\]
has a finite, nonzero normalizer $Z_\lambda$. Under the affine bridge with $x_0 \sim \mathcal{N}(0,I)$, every endpoint distribution induces a noisy marginal through the same Gaussian kernel,
\[
\rho_t(x) = \int p_t(x \mid x_1)\,\rho_1(x_1)\,dx_1,
\qquad
p_t(x \mid x_1) = \mathcal{N}\!\bigl(x;\, \beta_t x_1,\, \alpha_t^2 I\bigr),
\]
and likewise for $p_t$ and $\pi_t$. By \Cref{app:proof_prop1}, the three velocity fields are $u_t(x) = a_t x + c_t\mu_t(x)$, $u_t^\rho(x) = a_t x + c_t\mu_t^\rho(x)$, and $u_t^\pi(x) = a_t x + c_t\mu_t^\pi(x)$, where $\mu_t$, $\mu_t^\rho$, and $\mu_t^\pi$ are the corresponding endpoint posterior means given $x_t = x$.

\paragraph{Step 1: Posterior of the geometric mixture.}
Bayes' rule gives the endpoint posteriors of the two components,
\[
p(x_1 \mid x_t = x) = \frac{p_t(x \mid x_1)\,p_1(x_1)}{p_t(x)},
\qquad
\rho(x_1 \mid x_t = x) = \frac{p_t(x \mid x_1)\,\rho_1(x_1)}{\rho_t(x)}.
\]
Because the two share the kernel and the exponents sum to one,
\[
p(x_1 \mid x)^{1-\lambda}\,\rho(x_1 \mid x)^{\lambda}
\;\propto\; p_t(x \mid x_1)\,p_1(x_1)^{1-\lambda}\rho_1(x_1)^{\lambda}
\;\propto\; \pi(x_1 \mid x_t = x).
\]
The posterior of the geometric mixture is thus the normalized geometric mixture of the component posteriors.

\paragraph{Step 2: Gaussian posteriors.}
Suppose now that the component posteriors are Gaussian, $p(x_1 \mid x) = \mathcal{N}(\mu_t(x), \Sigma_{p,t}(x))$ and $\rho(x_1 \mid x) = \mathcal{N}(\mu_t^\rho(x), \Sigma_{\rho,t}(x))$, with positive-definite covariances. A geometric mixture of Gaussians is again Gaussian, and completing the square gives
\begin{align}
\label{app:eq:geometric_covariance}
\Sigma_{\pi,t}^{-1}
&= (1-\lambda)\,\Sigma_{p,t}^{-1} + \lambda\,\Sigma_{\rho,t}^{-1}, \\
\label{app:eq:geometric_precision_mean}
\mu_t^\pi
&= \Sigma_{\pi,t}\bigl[(1-\lambda)\,\Sigma_{p,t}^{-1}\mu_t + \lambda\,\Sigma_{\rho,t}^{-1}\mu_t^\rho\bigr],
\end{align}
where every quantity may depend on $x$. The posterior mean of the mixture is a precision-weighted average of the component means.

\paragraph{Step 3: Shared covariance.}
If $\Sigma_{p,t} = \Sigma_{\rho,t}$, the precision weights cancel and \Cref{app:eq:geometric_precision_mean} reduces to the convex combination
\begin{equation}
\label{app:eq:geometric_mean}
\mu_t^\pi(x) = (1-\lambda)\,\mu_t(x) + \lambda\,\mu_t^\rho(x).
\end{equation}
This is the only step that needs the shared-covariance assumption. Gaussianity alone yields \Cref{app:eq:geometric_precision_mean}, not \Cref{app:eq:geometric_mean}.

\paragraph{Step 4: Guided velocity field.}
Substituting \Cref{app:eq:geometric_mean} into the affine-bridge parameterization gives
\begin{align}
u_t^\pi(x)
&= a_t x + c_t\bigl[(1-\lambda)\mu_t(x) + \lambda\mu_t^\rho(x)\bigr] \notag \\
&= u_t(x) + \lambda\, c_t\bigl(\mu_t^\rho(x) - \mu_t(x)\bigr).
\label{app:eq:guided_velocity}
\end{align}
For the linear bridge $c_t = (1-t)^{-1}$, and we recover \Cref{eq:retrieval_guided_velocity}.

\paragraph{Step 5: Empirical reference set.}
When $\rho_1$ is the empirical distribution $\hat{\rho}_1 = \frac{1}{M}\sum_{m=1}^M \delta_{x^{(m)}}$, the derivation of \Cref{app:proof_prop2} applies verbatim and the reference posterior mean is
\[
\hat{\mu}_t^\rho(x)
= \sum_{m=1}^M w_t^{(m)}(x)\,x^{(m)},
\]
with $w_t^{(m)}(x)$ the softmax weights of \Cref{app:eq:simplified_weights} computed over $\mathcal{R} = \{x^{(m)}\}_{m=1}^M$. Substituting into \Cref{app:eq:guided_velocity} gives the guided velocity in closed form. \hfill$\square$

\paragraph{Remark on the assumptions.}
The shared-covariance condition is strong, and it does not follow from simpler assumptions. Gaussian endpoint distributions $p_1$ and $\rho_1$ with different covariances generally violate it, and working in a VAE latent space does not help, since a small curated reference set is a biased subset of the data rather than a Gaussian-like sample. When the reference measure is empirical and the training distribution is continuous, the two are mutually singular and the geometric mixture $\pi$ is not even defined, although the finite-bank update remains well defined. We therefore read the endpoint-level construction as motivating the form of the correction, not as an exact description of RMG. The arithmetic construction of \Cref{app:reference_sets} removes these assumptions at the price of an intractable weight. Both derivations hold for any affine interpolant, whose forward kernel stays Gaussian. For non-affine interpolants the kernel, and with it the guidance rule, is usually not available in closed form.

\paragraph{Remark on marginal-level score interpolation.}
The same guided velocity also follows algebraically from interpolating the scores of the noisy marginals. Differentiating $\rho_t$ under the integral sign gives
\begin{equation}
\label{app:eq:score_identity}
\nabla_x \log \rho_t(x)
= \mathbb{E}_{\rho_1}\!\left[-\frac{x - \beta_t x_1}{\alpha_t^2} \;\middle|\; x_t = x\right]
= \frac{\beta_t\mu_t^\rho(x) - x}{\alpha_t^2},
\end{equation}
so that, for $\beta_t \neq 0$,
\begin{equation}
\label{app:eq:mean_from_score}
\mu_t^\rho(x) = \frac{x + \alpha_t^2 \nabla_x \log \rho_t(x)}{\beta_t}.
\end{equation}
The log-linear density $q_t(x) \propto p_t(x)^{1-\gamma_t}\rho_t(x)^{\gamma_t}$ has score
\begin{equation}
\label{app:eq:guided_score}
\nabla_x \log q_t(x)
= (1-\gamma_t)\nabla_x \log p_t(x) + \gamma_t \nabla_x \log \rho_t(x),
\end{equation}
and since \Cref{app:eq:mean_from_score} is linear in the score,
\begin{equation}
\label{app:eq:guided_mean}
\widetilde{\mu}_t(x) := \frac{x + \alpha_t^2 \nabla_x \log q_t(x)}{\beta_t}
= (1-\gamma_t)\,\mu_t(x) + \gamma_t\,\mu_t^\rho(x).
\end{equation}
This yields the velocity of \Cref{app:eq:guided_velocity} with $\lambda$ replaced by $\gamma_t$. The identities are exact and, unlike the endpoint construction, allow a time-dependent coefficient. The price is that $q_t$ is in general not the marginal of any endpoint distribution under the bridge. The quantity $\widetilde{\mu}_t$ is then a score-implied surrogate rather than a posterior mean, and the resulting field need not satisfy the continuity equation for $q_t$, even when $\gamma_t$ is constant. For this reason we present the guided velocity as a score-motivated guidance rule rather than as an exact probability-flow velocity.

\paragraph{Remark on late-time instability.}
The correction $\gamma_t c_t(\mu_t^\rho(x) - \mu_t(x))$ can grow without bound as $t \to 1$, where $c_t$ diverges. At the same time, the bandwidth $\alpha_t^2$ in \Cref{app:eq:raw_weights} shrinks to zero and the reference posterior collapses onto the nearest reference point. For the linear bridge $c_t=1/(1-t)$, which motivates schedules $\gamma_t = \gamma_0(1-t)^r$ with $r \geq 1$. These cancel the divergence and keep the correction bounded along the whole trajectory, and for $r=2$ the correction vanishes at the endpoint. In the text-to-image experiments we additionally set $\gamma_t = 0$ for $t \geq 0.85$. Schedules that remain large late in the trajectory apply a strong correction exactly where the weights have collapsed onto a single reference, which is why they tend to produce outputs close to individual reference images (\Cref{app:beta0_sweep}).

\subsection{Reference-Set Formalism}
\label{app:reference_sets}

We close with a unified view of flow matching in terms of reference sets,
which places standard flow matching, RMG, and SPG in a single framework. It
also contains two results used in the main text, an arithmetic-mixture
construction that is exact without distributional assumptions, and a
continuity-equation calculation that shows what a time-dependent schedule
does.

\paragraph{Setup.}
Let $\{x^{(n)}\}_{n=1}^N$ be a training set drawn i.i.d.\ from $p_1$, and let
$\mathcal{R} = \{x^{(m)}\}_{m=1}^M$ be a reference set drawn from a possibly
different distribution $\rho_1$, with noisy marginals $p_t$ and $\rho_t$
under the affine bridge as in \Cref{app:proof_thm3}. For the empirical
distributions
\[
\hat{p}_1 = \frac{1}{N}\sum_{n=1}^N \delta_{x^{(n)}},
\qquad
\hat{\rho}_1 = \frac{1}{M}\sum_{m=1}^M \delta_{x^{(m)}},
\]
the posterior means $\hat{\mu}_t$ and $\hat{\mu}_t^\rho$ are the
Nadaraya--Watson estimators of \Cref{app:proof_prop2} over $\mathcal{D}$ and
$\mathcal{R}$, and by \Cref{app:proof_prop1} each defines a velocity field,
$u_t(x) = a_t x + c_t\mu_t(x)$ and $u_t^\rho(x) = a_t x + c_t\mu_t^\rho(x)$.

\paragraph{Arithmetic mixture.}
An alternative to the geometric mixture of \Cref{app:proof_thm3} is the
arithmetic mixture
\[
\pi_\lambda = (1-\lambda)\,p_1 + \lambda\,\rho_1,
\qquad \lambda \in [0,1],
\]
whose noisy marginal is $\pi_t^\lambda(x) = (1-\lambda)\,p_t(x) + \lambda\,\rho_t(x)$. This construction applies equally to empirical measures. Bayes' rule gives the posterior mean
\begin{equation}
\label{app:eq:arithmetic_mean}
\mu_t^\lambda(x)
=
(1-\omega_t^*(x))\,\mu_t(x) + \omega_t^*(x)\,\mu_t^\rho(x),
\qquad
\omega_t^*(x)
=
\frac{\lambda \rho_t(x)}{(1-\lambda)p_t(x) + \lambda \rho_t(x)},
\end{equation}
and the corresponding velocity is the responsibility-weighted combination
\[
u_t^\lambda(x) = (1-\omega_t^*(x))\,u_t(x) + \omega_t^*(x)\,u_t^\rho(x).
\]
This holds with no distributional assumption, and the resulting field generates the mixture exactly. To see why, note that the continuity equation is linear in the flux. Multiplying by the mixture density gives
\[
\pi_t^\lambda u_t^\lambda = (1-\lambda)\,p_t u_t + \lambda\,\rho_t u_t^\rho,
\]
so the component equations $\partial_t p_t + \nabla\cdot(p_t u_t) = 0$ and $\partial_t \rho_t + \nabla\cdot(\rho_t u_t^\rho) = 0$ add up to
\[
\partial_t \pi_t^\lambda + \nabla\cdot\bigl(\pi_t^\lambda u_t^\lambda\bigr) = 0.
\]
Nothing is assumed about $p_1$ and $\rho_1$, and any affine interpolant will do. With exact responsibilities and a constant $\lambda$, integrating $u_t^\lambda$ from the common source therefore samples the arithmetic mixture. The difficulty is that $\omega_t^*(x)$ requires the scalar density $p_t(x)$, which a pretrained velocity field does not provide. Replacing $\omega_t^*(x)$ with the constant $\lambda$ recovers the guided velocity of the geometric construction, so both routes lead to the same practical rule. In the special case where both distributions are empirical and $\lambda = \frac{M}{N+M}$, $\pi_\lambda$ is the uniform distribution over the union of training and reference points, and $\mu_t^\lambda$ is the Nadaraya--Watson estimator over both sets.

\paragraph{Time-dependent schedules.}
In practice the guidance coefficient varies with $t$. The induced path is then no longer the bridge of a fixed endpoint mixture, but the arithmetic construction still tells us exactly what the scheduled dynamics do. For a differentiable $\lambda : [0,1] \to [0,1]$, let $\tilde{\pi}_t = (1-\lambda(t))\,p_t + \lambda(t)\,\rho_t$, and let $u_t^{\lambda(t)}$ be the responsibility-weighted field of \Cref{app:eq:arithmetic_mean} evaluated at $\lambda(t)$, with responsibility $\omega_t^{\lambda(t)}(x)$. The same flux calculation now gives
\begin{equation}
\label{app:eq:scheduled_source}
\partial_t \tilde{\pi}_t + \nabla\cdot\bigl(\tilde{\pi}_t\, u_t^{\lambda(t)}\bigr) = \lambda'(t)\,\bigl(\rho_t - p_t\bigr).
\end{equation}
The schedule therefore transports the interpolated family up to a source term proportional to $\lambda'(t)$. This is why we present $\gamma_t$ as a time-dependent, score-motivated guidance rule, in the spirit of scheduled classifier-free guidance. If the responsibility is further replaced by the scalar $\gamma_t$, so that $v_t = u_t + \gamma_t(u_t^\rho - u_t)$, a second term appears,
\begin{align}
\partial_t \tilde{\pi}_t + \nabla\cdot(\tilde{\pi}_t v_t)
={}& \lambda'(t)\,(\rho_t - p_t)
+ \nabla\cdot\!\Bigl[\tilde{\pi}_t\bigl(\gamma_t - \omega_t^{\lambda(t)}\bigr)\bigl(u_t^\rho - u_t\bigr)\Bigr],
\end{align}
which accounts for the gap between the scalar schedule and the state-dependent responsibilities. Together with the model and finite-bank approximations, these two terms are what separate RMG from exact mixture transport.

\paragraph{Relation to RMG and SPG.}
The formalism places the methods of the main paper side by side. Standard
flow matching is \emph{self-referencing}. A network is trained so that
$u_t^\theta \approx u_t$, which amounts to learning $\mu_t$ and encoding the
training distribution in the model parameters, the degenerate case
$\hat{\rho}_1 = \hat{p}_1$. RMG cross-references at test time, replacing
$\mu_t$ in the guided velocity of \Cref{app:eq:guided_velocity} by the
pretrained estimate $\mu_t^\theta$ and computing $\hat{\mu}_t^\rho$ in
closed form from $\mathcal{R}$. SPG learns to cross-reference. Its anchor is
a learned aggregation of the references rather than an estimate of
$\mu_t^\rho$, and it targets neither of the mixtures above. What it shares
with RMG is that the endpoint prediction, and therefore the flow, is governed
by the reference set. In every case guidance arises from the posterior mean
that a reference set defines, and reference-guided methods exploit the
freedom to choose that set at test time.

\section{Related Work}
\label{app:related_work}

\subsection{Flow Matching, Diffusion Models, and Endpoint Posteriors}

Flow matching learns a velocity field along a prescribed probability
path~\citep{lipman2023,albergo2023building,liu2023flow}, and diffusion and
score-based models learn the corresponding reverse-time dynamics or score
fields~\citep{ho2020denoising,song2021score}. Conditional flow matching
extends the framework to conditional paths~\citep{tong2024improving},
rectified flow favors straight paths~\citep{liu2023flow}, stochastic
interpolants unify deterministic and stochastic
dynamics~\citep{albergo2023building}, and scalable interpolant transformers
pair these objectives with modern transformer backbones~\citep{ma2024sit}.

Our analysis starts from the observation that, under a linear bridge, the
optimal velocity field is fully determined by the conditional endpoint mean.
Variational flow matching formalizes this posterior
view~\citep{eijkelboom2024variational}, and closed-form analyses show that
finite-sample flows and scores can be written as kernel-weighted
aggregations over training
examples~\citep{scarvelis2025closedform,niedoba2024nearest,bertrand2026on,gao2024flow},
exposing the non-parametric structure implicit in trained generative models.
These works use that structure to analyze generation. We use it to control
generation, by changing at test time the reference set over which the
posterior mean is taken.

\subsection{Guidance and Controllable Generation}

Classifier guidance steers sampling by adding the gradients of an external
classifier to the score~\citep{dhariwal2021diffusion}, and classifier-free
guidance interpolates conditional and unconditional predictions to
strengthen prompt adherence~\citep{ho2022classifier}. Recent work extends
guidance theory to flow matching~\citep{feng2025on}. Our correction has a
different source. It is a difference between two endpoint means, so it
needs neither an auxiliary classifier nor a reward, and computing it
requires no additional model or gradient evaluation.

A second family augments pretrained generators with dedicated conditioning
modules. ControlNet~\citep{zhang2023adding} and
T2I-Adapter~\citep{mou2024t2i} add trainable branches for spatial conditions
such as edges, depth, or pose. IP-Adapter~\citep{ye2023ip},
BLIP-Diffusion~\citep{li2023blip}, and InstantID~\citep{wang2024instantid}
condition on reference images through dedicated encoders and attention
pathways, and Prompt-to-Prompt~\citep{hertz2022prompt} and
MasaCtrl~\citep{cao2023masactrl} manipulate attention maps to preserve
structure during editing. RMG uses no conditioning module at all. The
reference set acts on generation only through the posterior mean it induces.

A third family spends additional computation at inference time.
Best-of-$N$~\citep{karthik2023dontsucceedtrytry} selects among several
candidates, SMC~\citep{wu2024practicalasymptoticallyexactconditional}
resamples trajectories under an external score, prompt
optimization~\citep{manas2024improvingtexttoimageconsistencyautomatic}
searches for better prompts, and
ReNO~\citep{eyring2024renoenhancingonesteptexttoimage} optimizes the initial
noise with reward gradients. These methods avoid fine-tuning, and their cost
grows with the number of candidates, scores, or optimization steps. RMG
corrects a single trajectory in closed form, without a reward model, a
classifier, or a search over candidates.

Closest in spirit are methods based on reward tilting and flow maps. Tilt
Matching~\citep{potaptchik2025tilt} derives velocity fields for
reward-tilted endpoint distributions by regression,
FMTT~\citep{sabour2025testtime} uses a flow-map lookahead to steer
trajectories toward high-reward endpoints, and Diamond
Maps~\citep{holderrieth2026diamond} learn stochastic flow maps for scalable
reward alignment through SMC and guidance. Like ours, these methods treat
endpoint information as the key to controllable generation. They express the
target through a scalar reward, however, and require training or repeated
evaluation. RMG specifies the target with an empirical reference set, which
defines the endpoint posterior mean directly and yields a data-defined drift
correction without reward gradients, value estimates, or Monte Carlo
rollouts.

\subsection{Retrieval-Augmented and Semi-Parametric Generation}

kNN-LM interpolates the predictions of a language model with
nearest-neighbor statistics from a datastore~\citep{khandelwal2020generalization},
RAG conditions generation on retrieved documents~\citep{lewis2020retrieval},
and RETRO scales the idea to very large corpora~\citep{borgeaud2022improving}.
In vision, retrieval-augmented diffusion models condition image synthesis on
retrieved neighbors~\citep{blattmann2022retrieval}. All of these methods
feed retrieved content to the model as additional context. We use the
reference set differently, to define a posterior statistic of the generative
path itself. RMG compresses the set into a closed-form endpoint mean. SPG
reads it through a learned attention anchor and a residual refiner, which
places it between purely non-parametric guidance and fully parametric
conditional generation.

\subsection{Personalization and Low-Data Adaptation}

Textual Inversion~\citep{gal2022image},
DreamBooth~\citep{ruiz2023dreambooth}, Custom
Diffusion~\citep{kumari2023multi}, and LoRA~\citep{hu2022lora} adapt model
parameters to new concepts from a few examples. Because they write each
concept into the weights, adding, removing, or recombining concepts requires
further optimization. C-LoRA addresses this continual-customization
setting~\citep{smith2024continual}, and test-time adaptation methods handle
distribution shift by updating parameters or normalization statistics at
deployment~\citep{sun2020test,wang2021tent}. Reference-guided flows offer a
different interface. The model stays fixed, and adaptation becomes a data
operation, carried out by editing the reference set rather than by
optimization.

\subsection{Attention as Posterior Aggregation}

Under a Gaussian bridge with an empirical target, the endpoint posterior
mean is a softmax-weighted average of data points. This is algebraically
identical to cross-attention with the noisy state as query and the reference
examples as keys and values, and similar structures appear in finite-sample
analyses of diffusion and score
estimation~\citep{niedoba2024nearest,scarvelis2025closedform}. The identity
motivates the reference-attention module of SPG, which replaces the fixed
Gaussian kernel with learned queries and keys (\Cref{sec:spg}).

\section{Experimental Details and Additional Results}
\label{app:experiments}
\label{app:additional_experiments}

This appendix follows the four questions of \Cref{sec:results}. For each
question we give the setup behind the results in the main text and then
report additional experiments.

\subsection{Q1: Mechanistic Validation}
\label{app:mechanistic_validation}

In both settings below the posterior mean can be computed exactly, so the
effect of the reference set is isolated from modeling error.

\begin{figure}[!htbp]
\centering
\begin{tabular}{@{}c@{\hspace{0.02\linewidth}}c@{}}
\begin{tabular}{@{}c@{\hspace{0.01\linewidth}}c@{}}
\small $t=0.02$ (near noise) & \small $t=0.25$ \\
\includegraphics[width=0.22\linewidth]{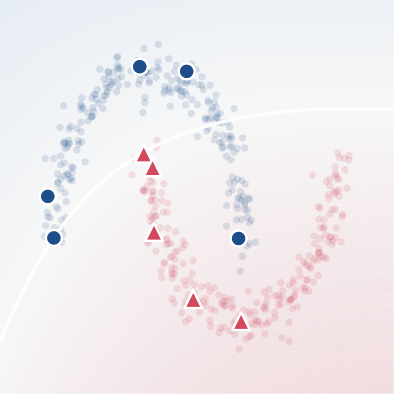} &
\includegraphics[width=0.22\linewidth]{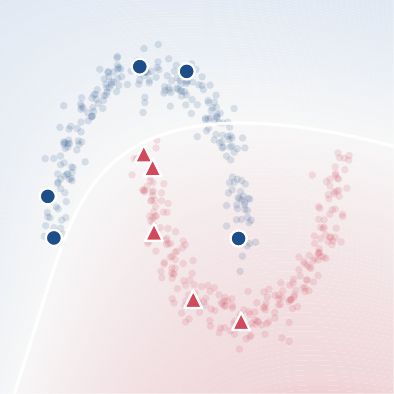}
\end{tabular}
&
\begin{tabular}{@{}c@{\hspace{0.01\linewidth}}c@{}}
\small $t=0.55$ & \small $t=0.92$ (near data) \\
\includegraphics[width=0.22\linewidth]{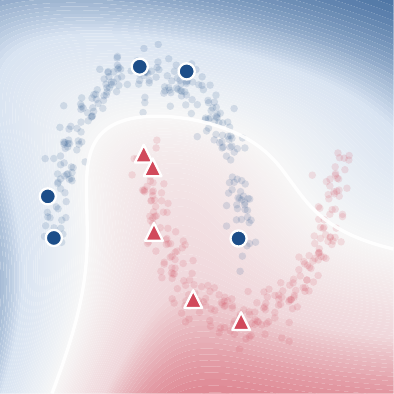} &
\includegraphics[width=0.22\linewidth]{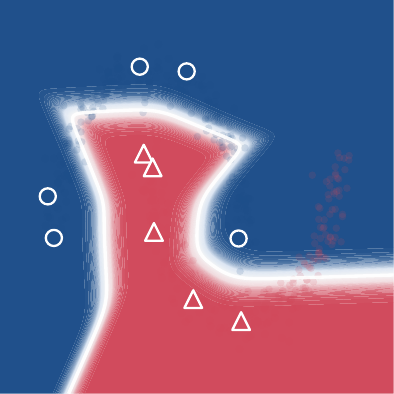}
\end{tabular}
\end{tabular}
\caption{\textbf{The posterior sharpens onto the class structure.} Posterior
weights over the two-moons dataset at increasing $t$, with the reference set
fixed.}
\label{fig:twomoons_combined}
\end{figure}

\paragraph{Two moons.}
With the setup of \Cref{sec:toy_experiments}, the posterior is diffuse near
noise and concentrates on the class structure as $t \to 1$
(\Cref{fig:twomoons_combined}). This sharpening is what lets the
composition of the reference set redirect the flow in
\Cref{fig:twomoons_flow}.

\begin{figure}[!htbp]
\centering
\begin{tabular}{ccc}
\small \textbf{Unconditional} & \small \textbf{Soft labels ($M=5$)} & \small \textbf{Hard filter} \\
\includegraphics[width=0.22\linewidth]{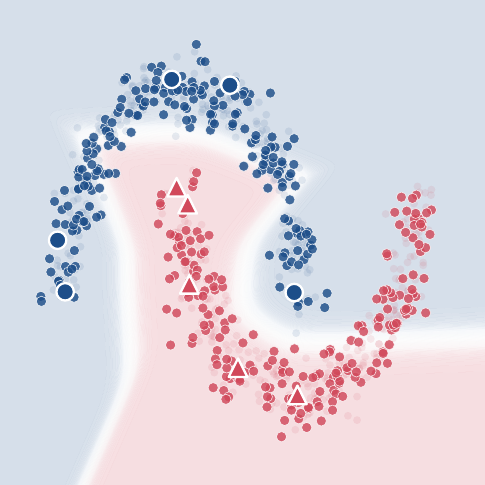} &
\includegraphics[width=0.22\linewidth]{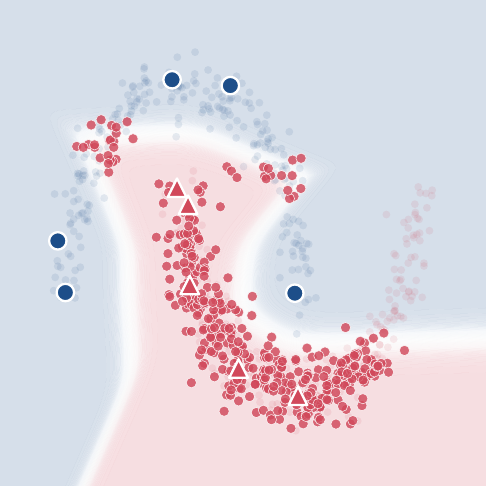} &
\includegraphics[width=0.22\linewidth]{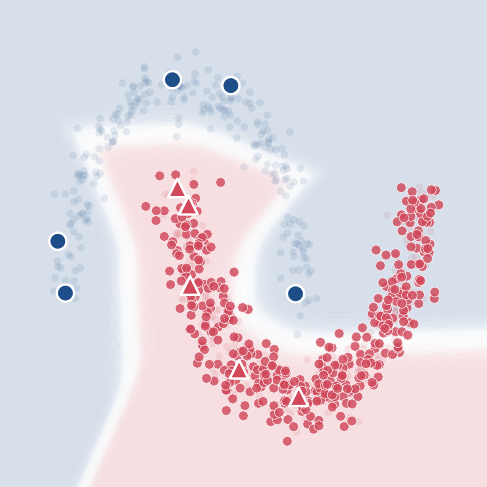}
\end{tabular}
\caption{\textbf{Five soft-labeled references come close to hard
filtering.} Inference-time conditioning on two moons, with dataset and model
fixed: unconditional weights, soft labels from $M=5$ labeled references,
and hard filtering to a single class.}
\label{fig:hard_soft_twomoons}
\end{figure}

We also compare three inference-time conditions. \emph{Unconditional
weighting} uses the standard posterior weights over the full dataset.
\emph{Soft posterior reweighting} uses a small labeled reference set to
assign soft class probabilities to the dataset, biasing the weights toward a
target class. \emph{Hard filtering} restricts the distribution to a single
class. With as few as $M=5$ labeled points, soft reweighting already steers
strongly toward the target class and approaches the hard-filter upper bound
(\Cref{fig:hard_soft_twomoons}).

\begin{figure}[!htbp]
\centering
\begin{minipage}[c]{0.56\linewidth}
\centering
\textbf{\small Generated ones}\\[0.2em]
\includegraphics[width=\linewidth, trim={0 0 0 1cm}, clip]{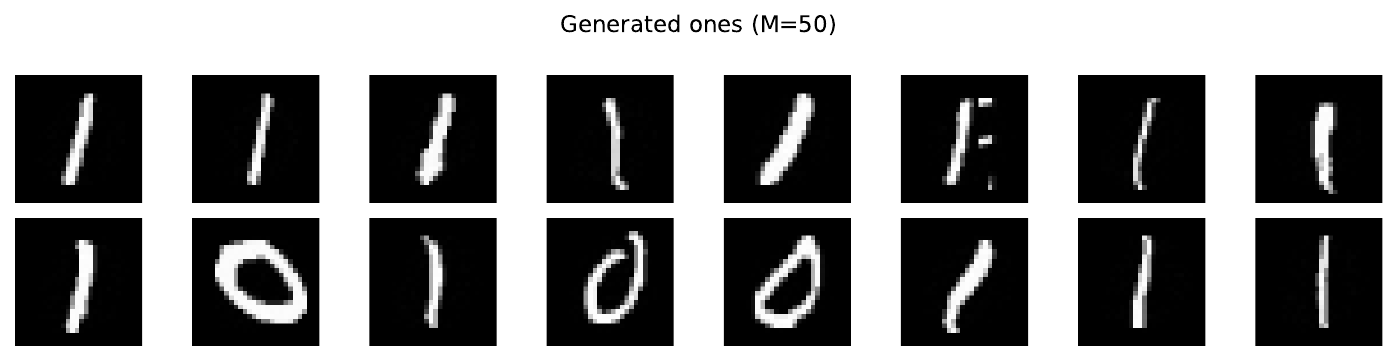}\\[0.5em]
\textbf{\small Generated zeros}\\[0.2em]
\includegraphics[width=\linewidth, trim={0 0 0 1cm}, clip]{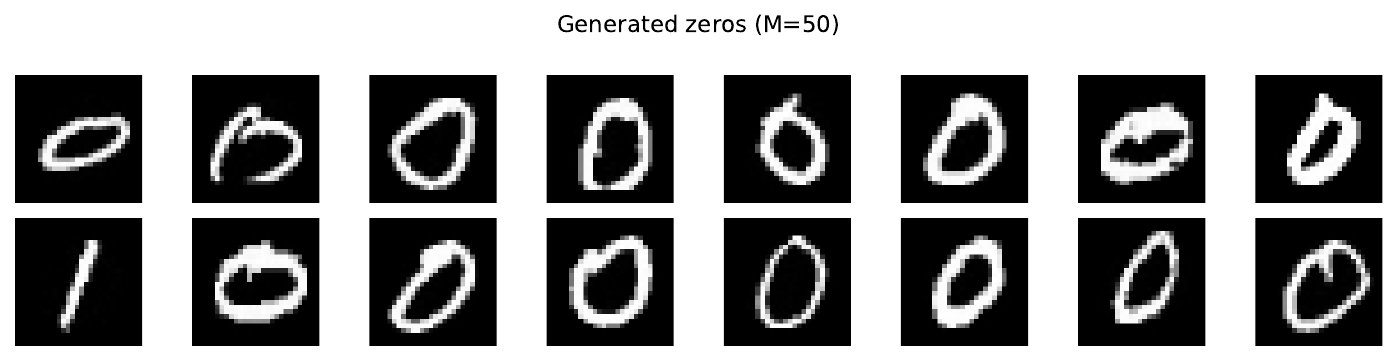}
\end{minipage}\hfill
\begin{minipage}[c]{0.42\linewidth}
\centering
\includegraphics[width=\linewidth]{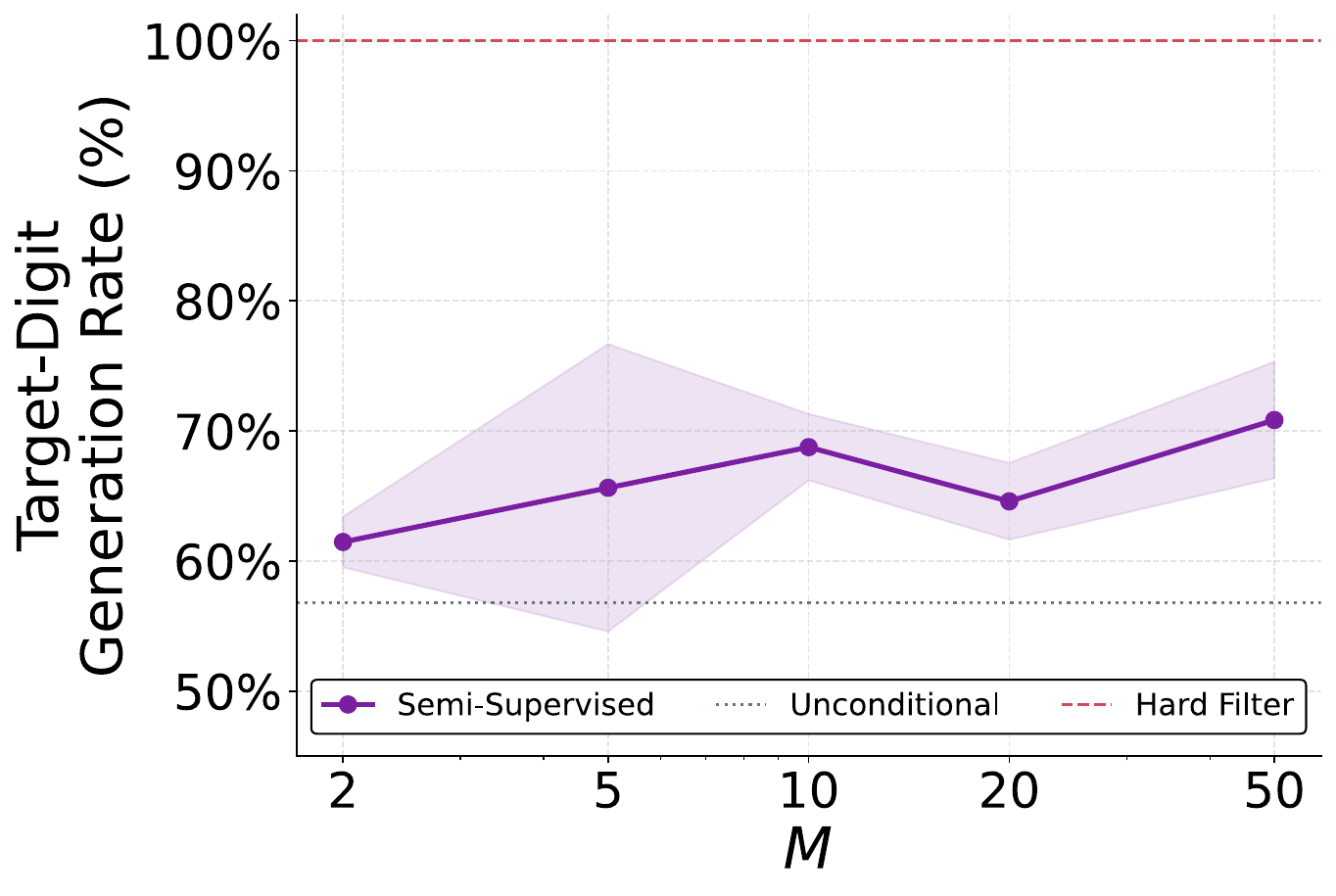}
\end{minipage}
\caption{\textbf{A few soft-labeled references steer generation in pixel
space.} \textbf{Left:} with model and noise fixed, the same model generates
ones or zeros depending on the reference set. \textbf{Right:} target-digit generation rate as a
function of the number of references $M$, compared with the unconditional
model and the hard-filter upper bound.}
\label{fig:mnist_results}
\end{figure}

\paragraph{MNIST.}
We repeat the analysis on MNIST digits 0 and 1, where the reference set now
acts in image space. With model and noise fixed, the same model generates
ones or zeros depending only on the reference set (\Cref{fig:mnist_results},
top). The target-digit rate exceeds that of the unconditional model already
at $M=2$ and rises to roughly 70\% at $M=50$ (\Cref{fig:mnist_results},
bottom). A handful of soft-labeled references therefore steers generation in
image space through the same mechanism as in two dimensions, although it
remains well short of the hard-filter upper bound.

\subsection{Q2: Training-Free Control of a Frozen Model}
\label{app:flux_experiments}

We first complete the guidance setup of \Cref{sec:experimental_setup}, then
separate the roles of prompt and references, and finally show that the
composition of a bank gives graded control.

\subsubsection{Guidance Setup}
\label{app:flux_sampling}

All text-to-image experiments apply the update of \Cref{eq:practical_rmg} to
\mbox{FLUX.2-klein} (4B). Reference images are encoded with the frozen VAE
into the same packed latent representation as the sampler state, so the
reference mean $\hat{\mu}_t^\rho(x)$ and the correction live in the
coordinate system of the pretrained model. The quadratic schedule and the
cutoff at $t \geq 0.85$ guard against the late-time instability discussed in
\Cref{app:proof_thm3}. In latent space, the reference softmax of
\Cref{eq:weights-softmax} is evaluated with temperature $\tau=\sqrt{d}$,
\[
w_t^{(m)}(x)
=
\mathrm{Softmax}_m\!\left(
-\frac{1}{2\tau}
\frac{\|x-tx^{(m)}\|^2}{(1-t)^2}
\right),
\]
where $d$ is the latent dimensionality. The dominant term of the squared
distance is an inner product between latent vectors, whose variance grows
with $d$. Dividing by $\sqrt{d}$ follows the same reasoning as scaled
dot-product attention and keeps the softmax from collapsing to a hard
nearest-neighbor assignment in high dimension. \Cref{tab:hyperparams} lists
the prompt and strength of each experiment.

\begin{table}[!htbp]
\centering
\caption{\textbf{Prompts and guidance strengths of the text-to-image experiments.}
    All experiments use \mbox{FLUX.2-klein} (4B) at $768\times768$
    resolution, with the quadratic schedule and softmax temperature of
    \Cref{app:flux_sampling}.}
\label{tab:hyperparams}
\setlength{\tabcolsep}{5pt}
\renewcommand{\arraystretch}{1.2}
\small
\begin{tabular}{p{0.24\linewidth}p{0.60\linewidth}c}
\toprule
\textbf{Experiment} & \textbf{Prompt} & \textbf{$\gamma_0$} \\
\midrule
\multirow{4}{*}{Bank swaps}
  & \textit{an elephant in a jungle} & 0.5 \\
  & \textit{a house in a forest}     & 1.0 \\
  & \textit{a cat}                   & 1.0 \\
  & \textit{an animal in a savanna}  & 1.0 \\
\midrule
\multirow{2}{*}{Prompt--reference}
  & \textit{an elephant in a jungle}     & \multirow{2}{*}{0.5} \\
  & \textit{a pink elephant in a jungle} & \\
\midrule
Geometric control
  & \textit{a miniature forest with tall pine trees, a glowing campfire, and fireflies drifting in the night sky, all inside a keyhole on a black background} & 0.2 \\
\midrule
Hand pose control
  & \textit{a hand doing the sign of the horns} & 0.8 \\
\midrule
Gymnastics pose control
  & \textit{a gymnast performing a ring leap, full body visible, airborne, one leg extended forward, the back leg bent high behind the head, arched back, pointed toes, arms extended, dynamic sports photograph} & 0.2 \\
\midrule
\multirow{2}{*}{Controllability}
  & \textit{an animal in a savanna}   & \multirow{2}{*}{1.0} \\
  & \textit{an elephant in a jungle}  & \\
\bottomrule
\end{tabular}
\end{table}

\subsubsection{Prompt--Reference Interaction}
\label{app:prompt_reference_interaction}

\begin{figure}[!htbp]
\centering
\setlength{\tabcolsep}{3pt}
\begin{tabular}{@{}>{\centering\arraybackslash}m{0.13\linewidth}ccc@{}}
& \small\textbf{No reference set} & \small\textbf{Elephant references} & \small\textbf{Pink-elephant references} \\
\scriptsize\textit{``an elephant in a jungle''}
& \parbox[c]{0.2\linewidth}{\centering\includegraphics[width=\linewidth]{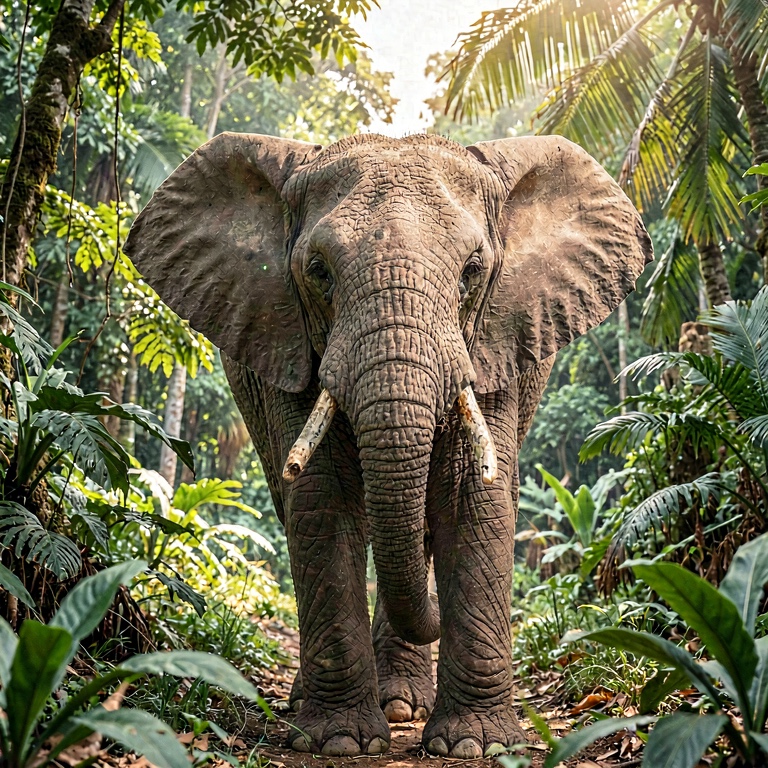}} & \parbox[c]{0.2\linewidth}{\centering\includegraphics[width=\linewidth]{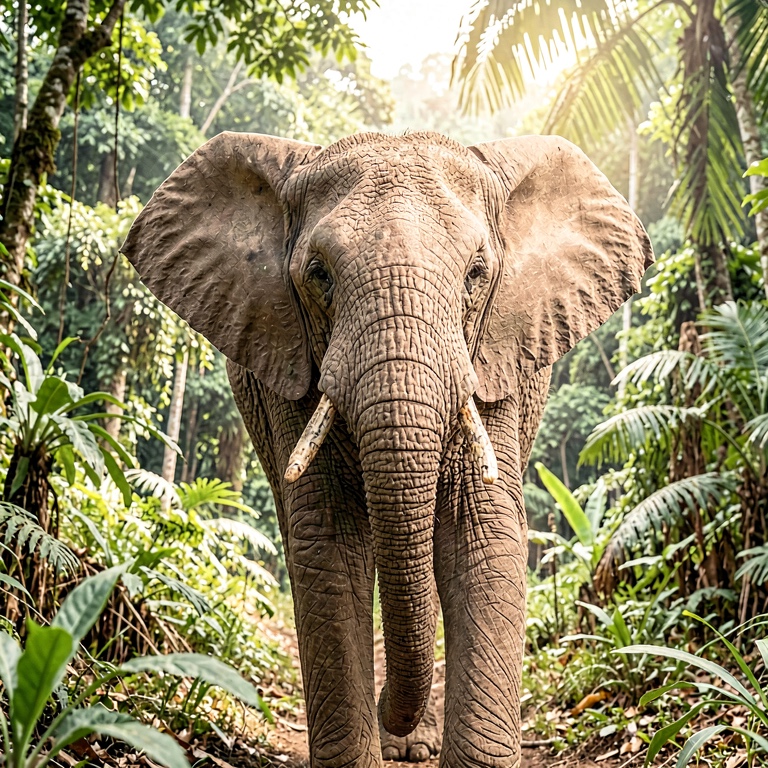}} & \parbox[c]{0.2\linewidth}{\centering\includegraphics[width=\linewidth]{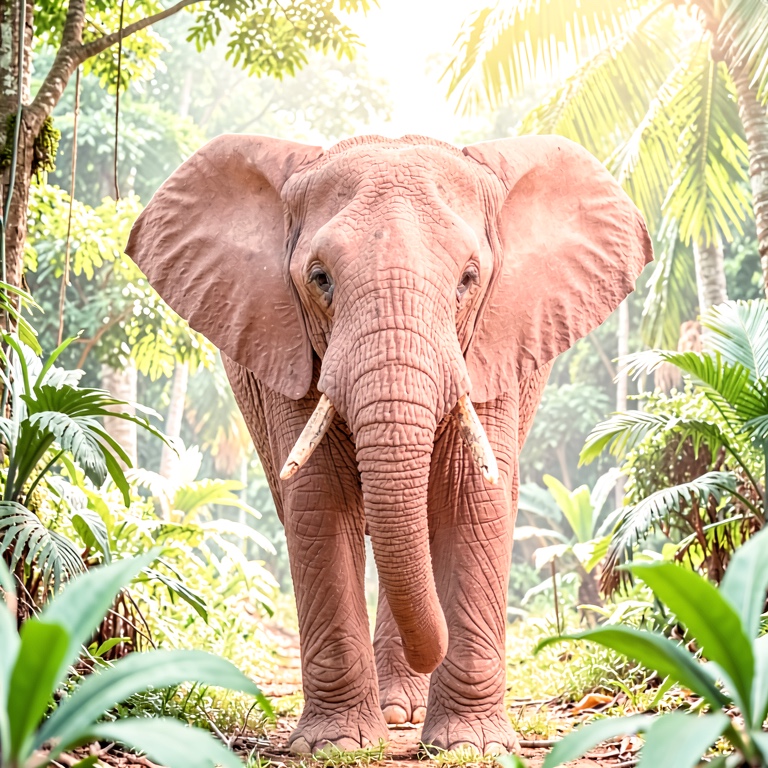}} \\
& \small $s_{\mathrm{pink}}=-0.029$ & \small $s_{\mathrm{pink}}=-0.026$ & \small $s_{\mathrm{pink}}=-0.015$ \\[0.4em]
\scriptsize\textit{``a pink elephant in a jungle''}
& \parbox[c]{0.2\linewidth}{\centering\includegraphics[width=\linewidth]{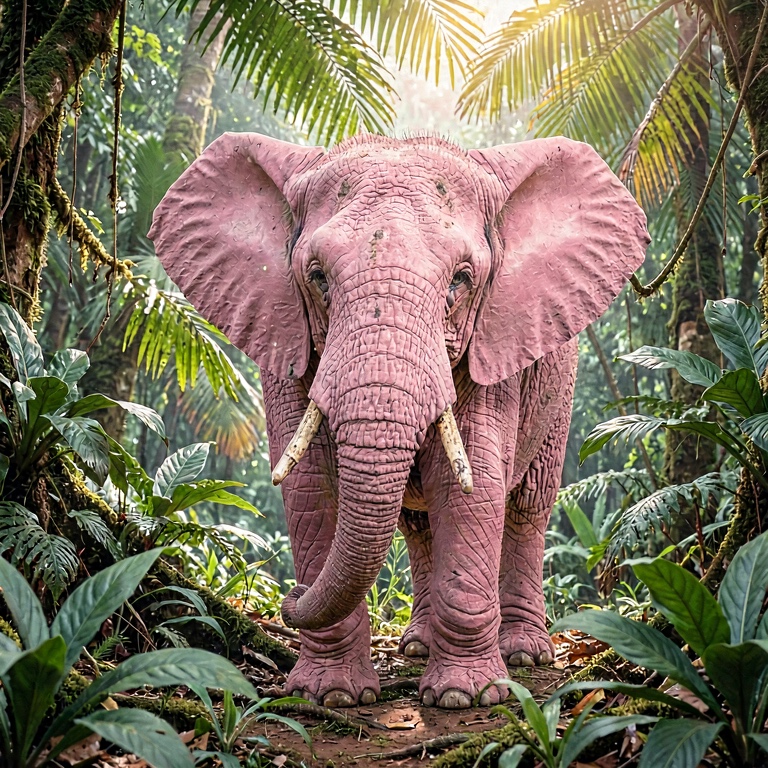}} & \parbox[c]{0.2\linewidth}{\centering\includegraphics[width=\linewidth]{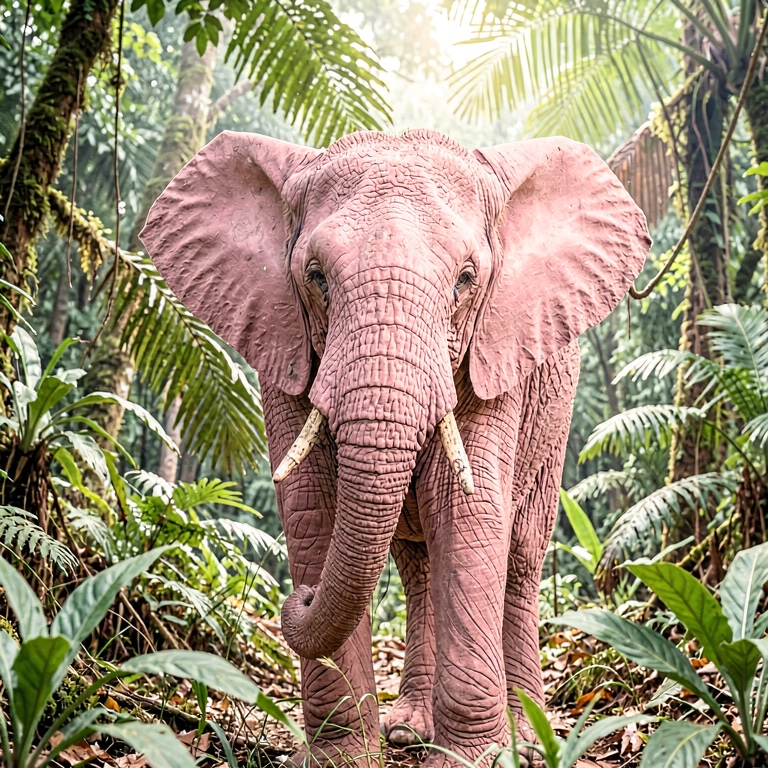}} & \parbox[c]{0.2\linewidth}{\centering\includegraphics[width=\linewidth]{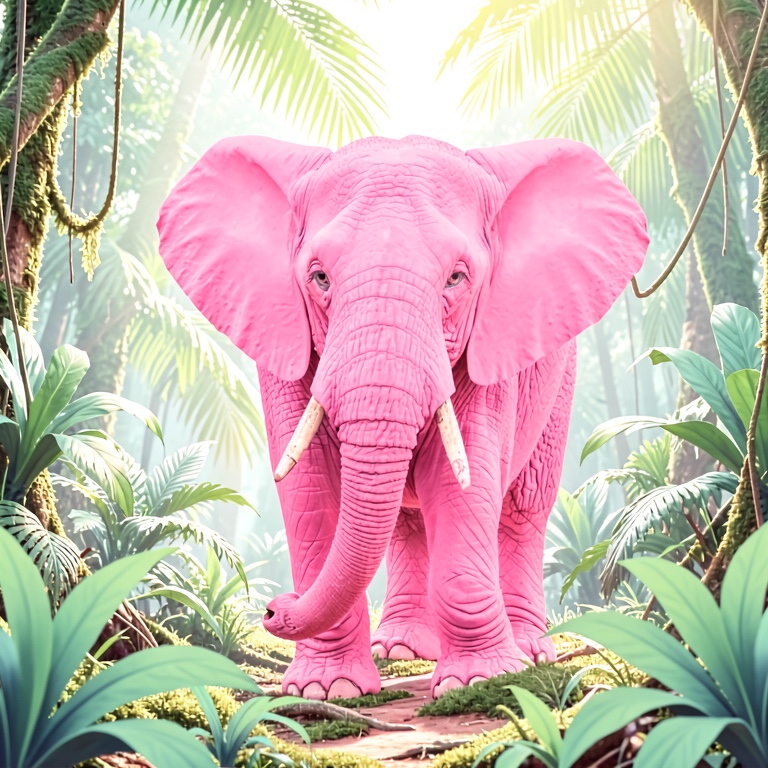}} \\
& \small $s_{\mathrm{pink}}=0.064$ & \small $s_{\mathrm{pink}}=0.024$ & \small $s_{\mathrm{pink}}=0.084$
\end{tabular}
\caption{\textbf{Prompt and reference set act as separate control axes.}
Outputs with a fixed noise seed. Rows vary
the prompt and columns the reference set. Values are the directional CLIP
pinkness score of \Cref{app:clip_attribute_score}.}
\label{fig:flux_prompt_reference}
\end{figure}

To separate the effect of the reference set from that of the prompt, we vary
the two independently. A neutral prompt and an attribute-specific prompt are
each crossed with three reference conditions: no reference set, a set of
ordinary elephants, and a set of pink elephants
(\Cref{fig:flux_prompt_reference}).

\paragraph{Pinkness score.}
\label{app:clip_attribute_score}
We measure pinkness with a directional CLIP score~\citep{radford2021learning}.
With $f_{\mathrm{img}}(x)$ and $f_{\mathrm{text}}(p)$ the normalized CLIP
embeddings of an image $x$ and a prompt $p$,
\[
s_{\mathrm{pink}}(x)
= \cos\!\bigl(f_{\mathrm{img}}(x), f_{\mathrm{text}}(p_{\mathrm{pink}})\bigr)
- \cos\!\bigl(f_{\mathrm{img}}(x), f_{\mathrm{text}}(p_{\mathrm{gray}})\bigr),
\]
where $p_{\mathrm{pink}} =$ \textit{``a pink elephant in a jungle''} and
$p_{\mathrm{gray}} =$ \textit{``a gray elephant in a jungle''}. Positive
values mean that the image is closer to the pink prompt than to the gray
one, and larger values indicate a stronger pink attribute. We use the score
only as a continuous proxy for the attribute, not as a calibrated
classifier.

\paragraph{Results.}
Without a reference set, the output follows the prompt. The ordinary
elephant references suppress the pink attribute even when the prompt asks
for it, lowering $s_{\mathrm{pink}}$ from 0.064 to 0.024. The pink
references push the neutral prompt toward pink ($s_{\mathrm{pink}}$ from
$-0.029$ to $-0.015$) and amplify the attribute when prompt and references
agree, raising the score to 0.084. Prompt and reference set thus act as
separate control axes that can reinforce or counteract each other.

\subsubsection{Reference Composition}
\label{app:reference_composition}

Reference-mean guidance also allows graded control through the composition
of the reference set. We mix two attribute-specific banks, keep the prompt
and sampler fixed, and vary the share of the target attribute over
$\{0, 25, 50, 75, 100\}\%$, with all other settings as in
\Cref{tab:hyperparams}.

\paragraph{Metrics.}
We measure the composition of the outputs in two ways. The first uses the
vision-language model Qwen2-VL-7B~\citep{Qwen2-VL} as a
zero-shot classifier. For each image we ask a binary question tailored to
the task, for example
\begin{quote}
\small
\textit{``Which animal does the main animal in this image most resemble:
zebra or giraffe? Answer with exactly one word: zebra or giraffe.''}
\end{quote}
and map the answer to one of the two attributes. The second assigns an image
to the target attribute when its CLIP similarity to the target prompt
exceeds its similarity to the distractor prompt. Both yield the percentage of
outputs with the target attribute. As a continuous complement, we also
report the mean CLIP similarity of the outputs to the target and distractor
prompts.

\begin{figure}[!htbp]
    \centering
    \includegraphics[width=0.48\linewidth]{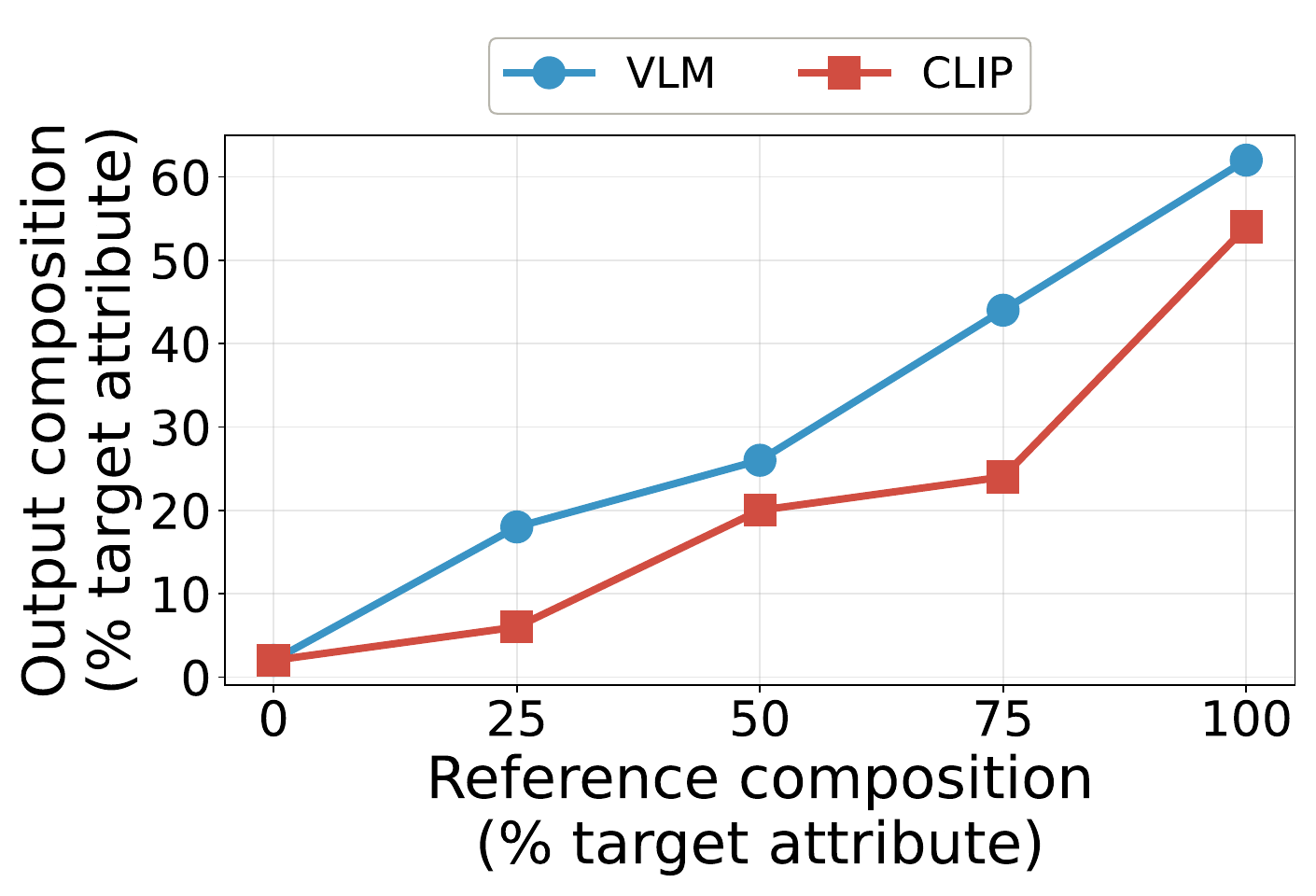}\hfill
    \includegraphics[width=0.48\linewidth]{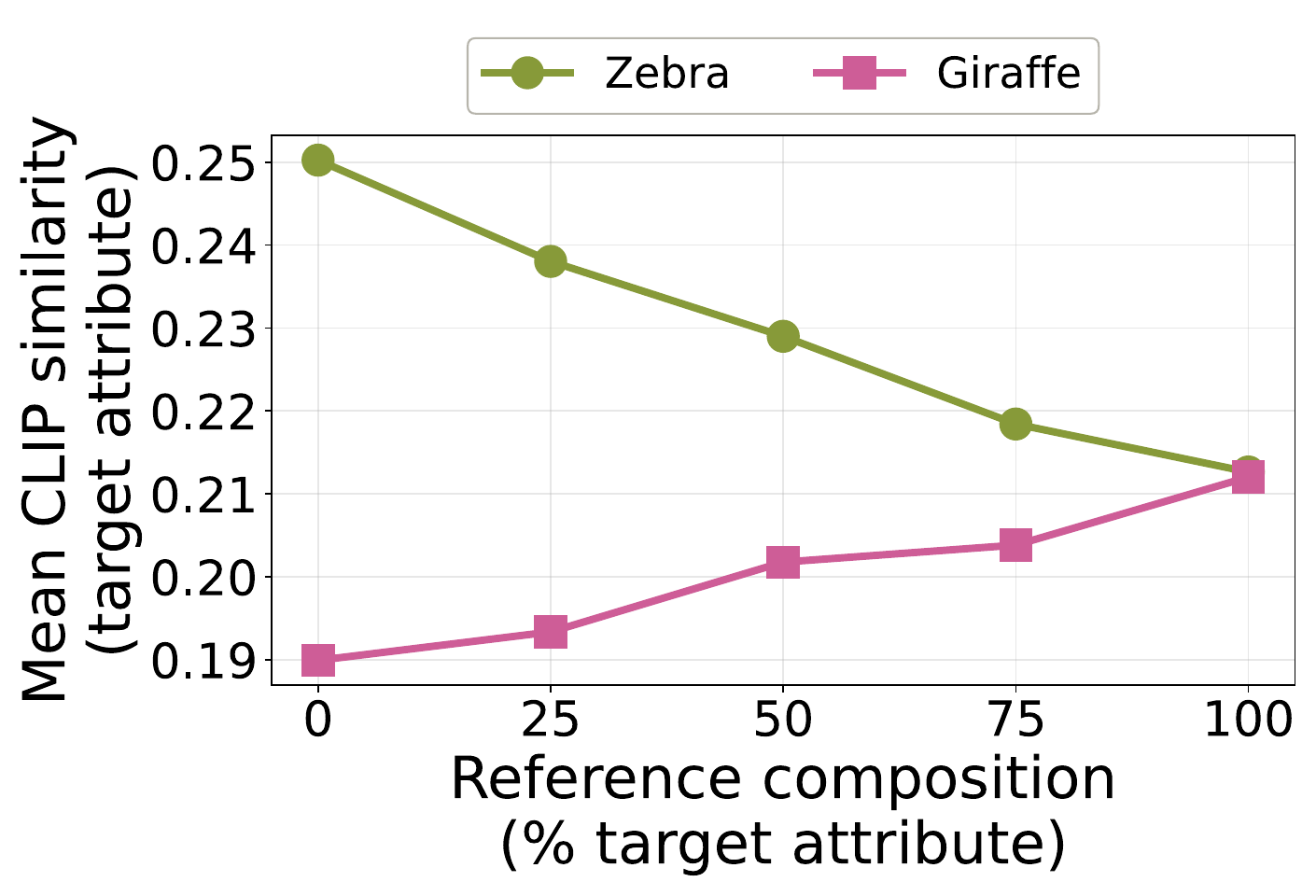}
    \caption{\textbf{The output composition follows the composition of the
    reference set.} Results for \textit{``an animal in a savanna''}.
    \textbf{Left:} fraction of outputs classified as giraffes by Qwen2-VL-7B
    and by CLIP. \textbf{Right:} mean CLIP similarity to the zebra and giraffe
    prompts. Both move steadily toward giraffes as their share of the
    reference set grows.}
    \label{fig:appendix_reference_composition_quant}
\end{figure}

\begin{figure}[!htbp]
\centering
\setlength{\tabcolsep}{2pt}
\begin{tabular}{@{}ccc@{}}
\small Unguided & \small 0\% giraffes & \small 25\% giraffes \\
\includegraphics[width=0.26\linewidth]{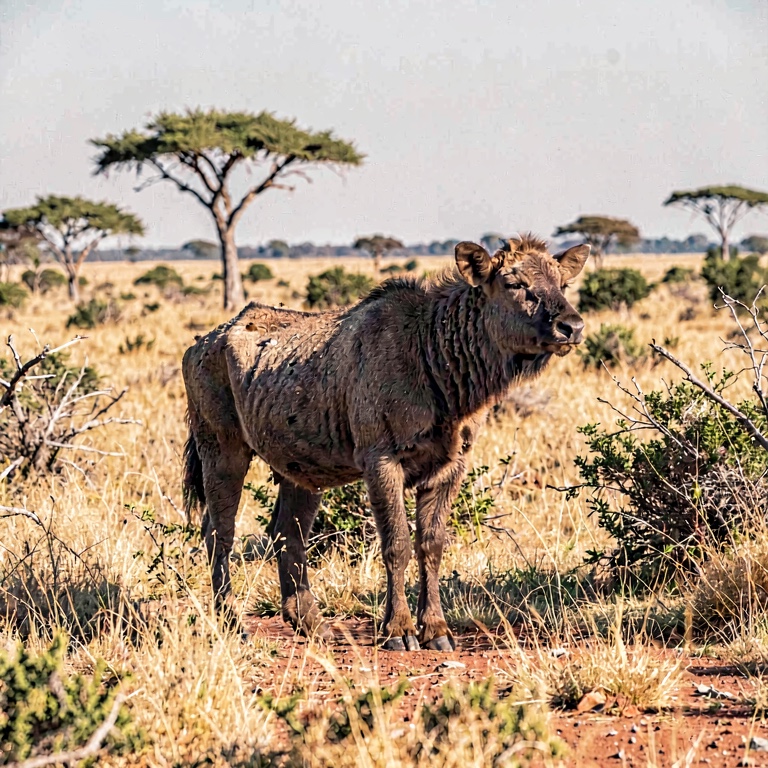} &
\includegraphics[width=0.26\linewidth]{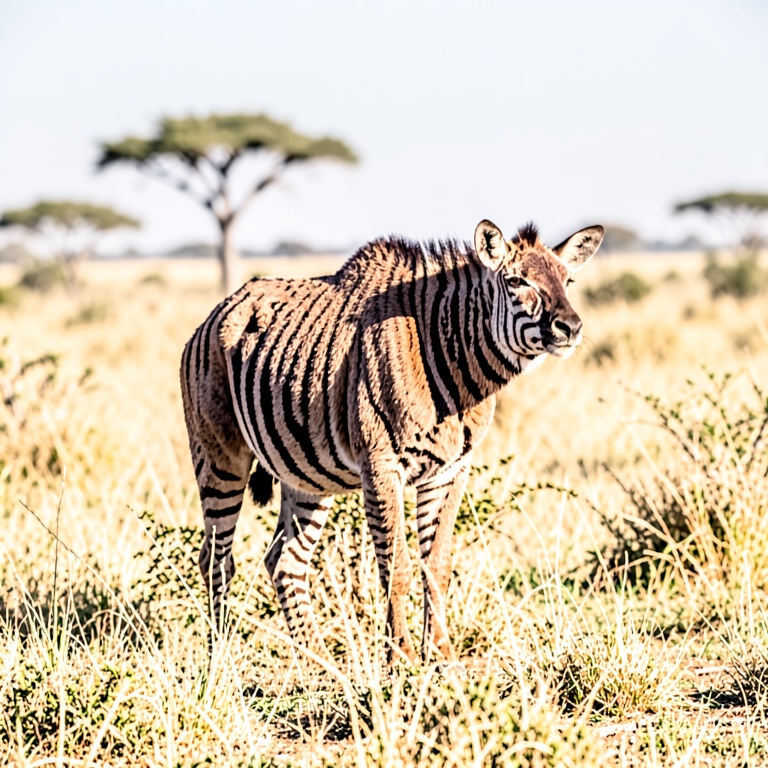} &
\includegraphics[width=0.26\linewidth]{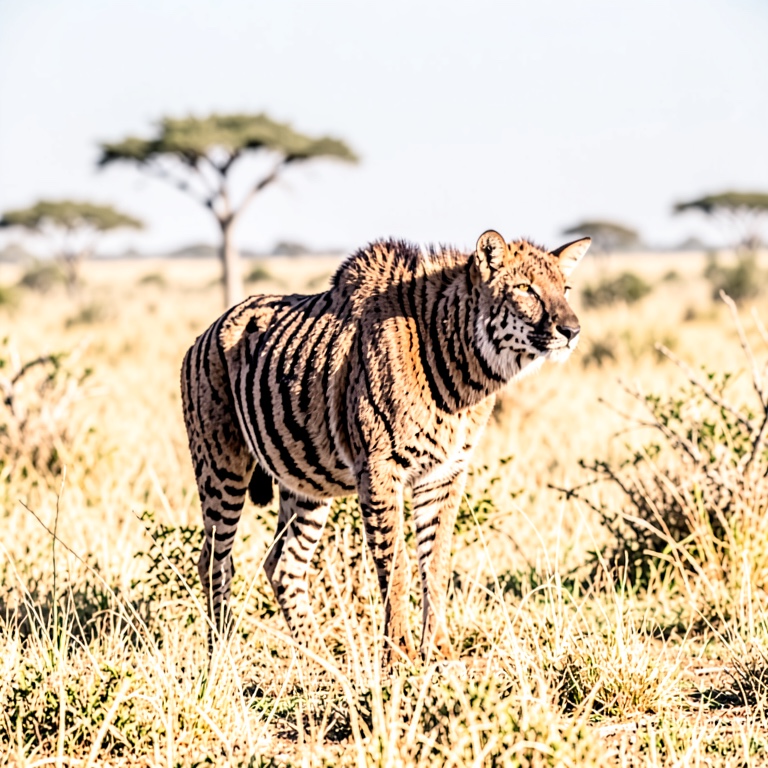} \\[0.2em]
\small 50\% giraffes & \small 75\% giraffes & \small 100\% giraffes \\
\includegraphics[width=0.26\linewidth]{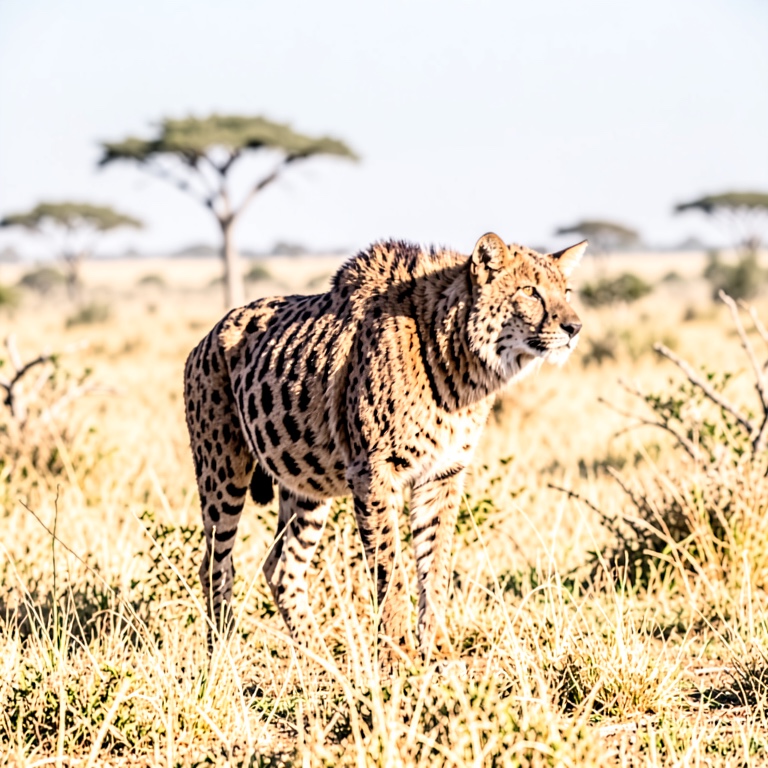} &
\includegraphics[width=0.26\linewidth]{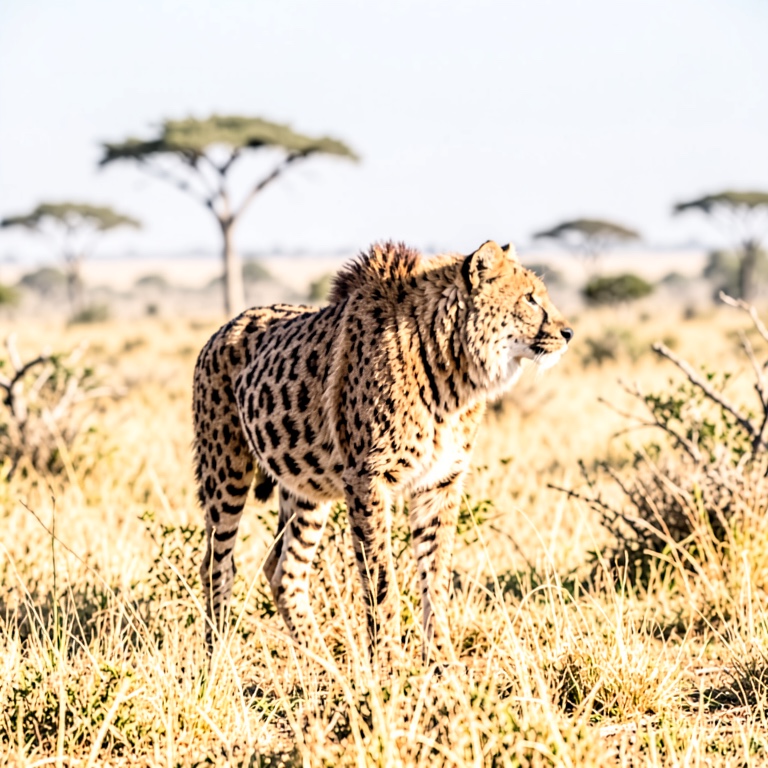} &
\includegraphics[width=0.26\linewidth]{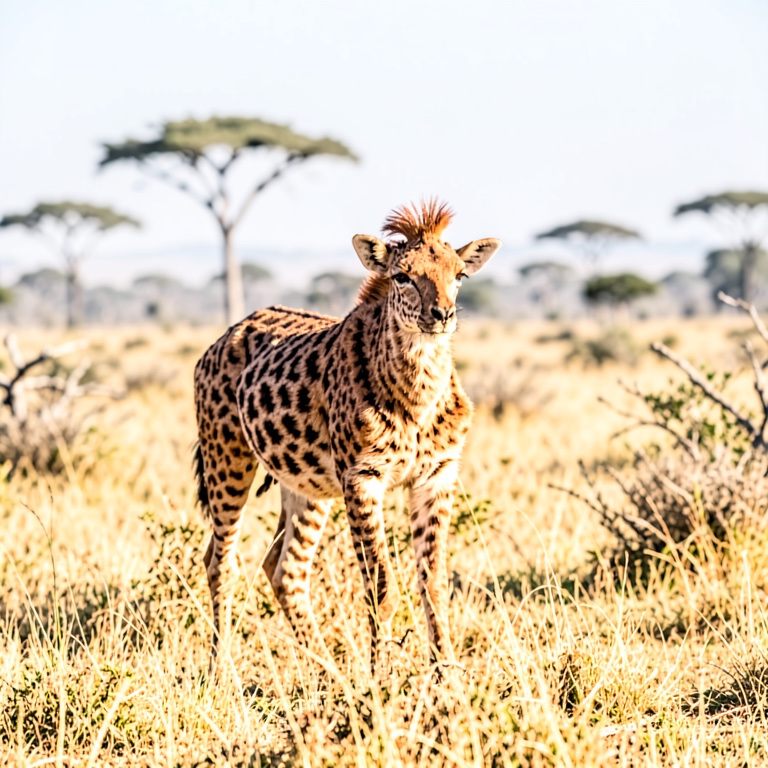}
\end{tabular}
\caption{\textbf{Mixing two banks gives graded control over object
identity.} Zebra and giraffe banks mixed for \textit{``an animal in a
savanna''}, with the prompt fixed. Labels give the share of
giraffes in the reference set, the remainder being zebras. Outputs shift
toward giraffes as their share grows.}
\label{fig:appendix_reference_composition_savanna}
\end{figure}

\begin{figure}[!htbp]
\centering
\setlength{\tabcolsep}{2pt}
\begin{tabular}{@{}ccc@{}}
\small Unguided & \small 0\% pink & \small 25\% pink \\
\includegraphics[width=0.26\linewidth]{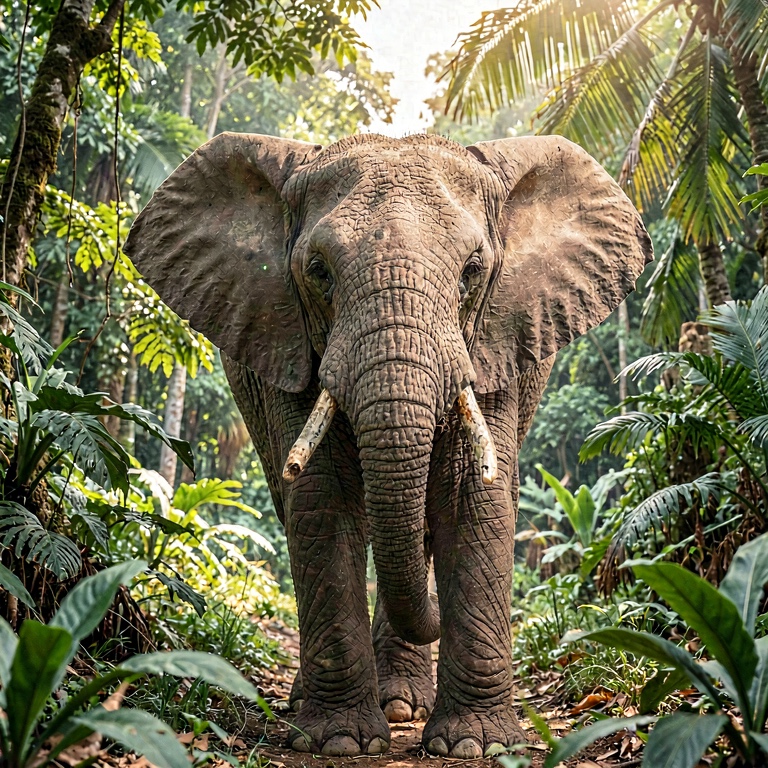} &
\includegraphics[width=0.26\linewidth]{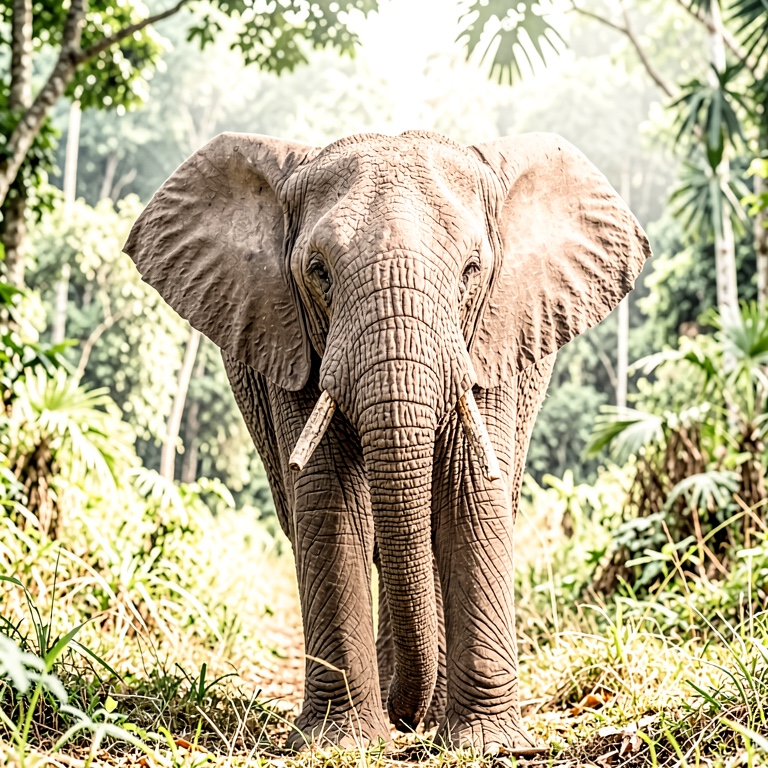} &
\includegraphics[width=0.26\linewidth]{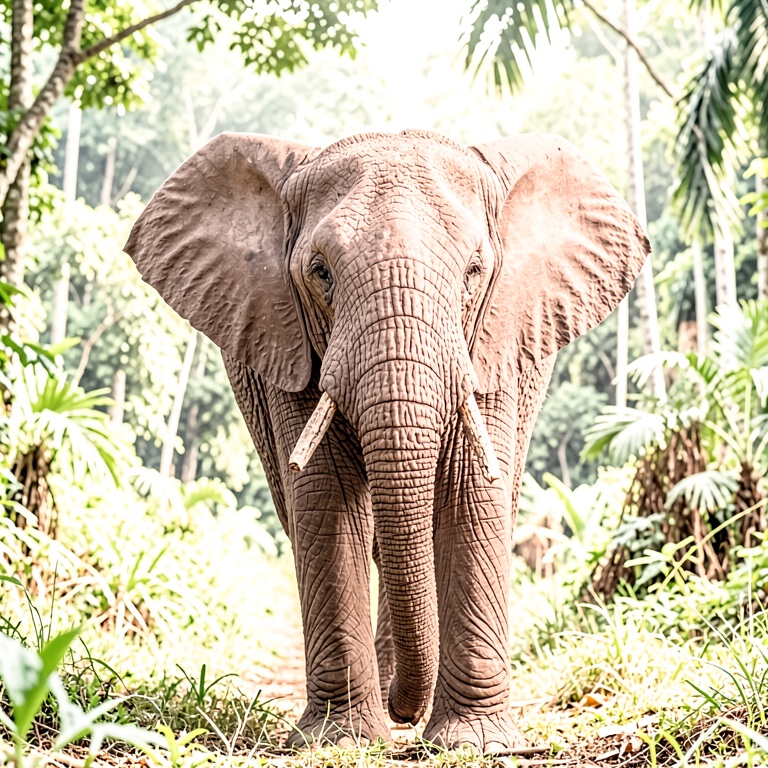} \\[0.2em]
\small 50\% pink & \small 75\% pink & \small 100\% pink \\
\includegraphics[width=0.26\linewidth]{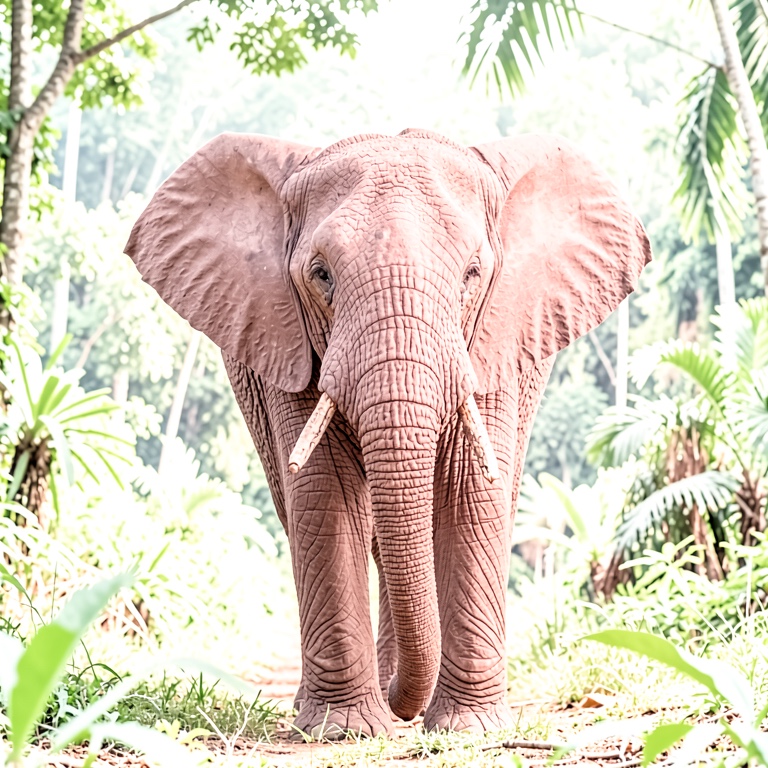} &
\includegraphics[width=0.26\linewidth]{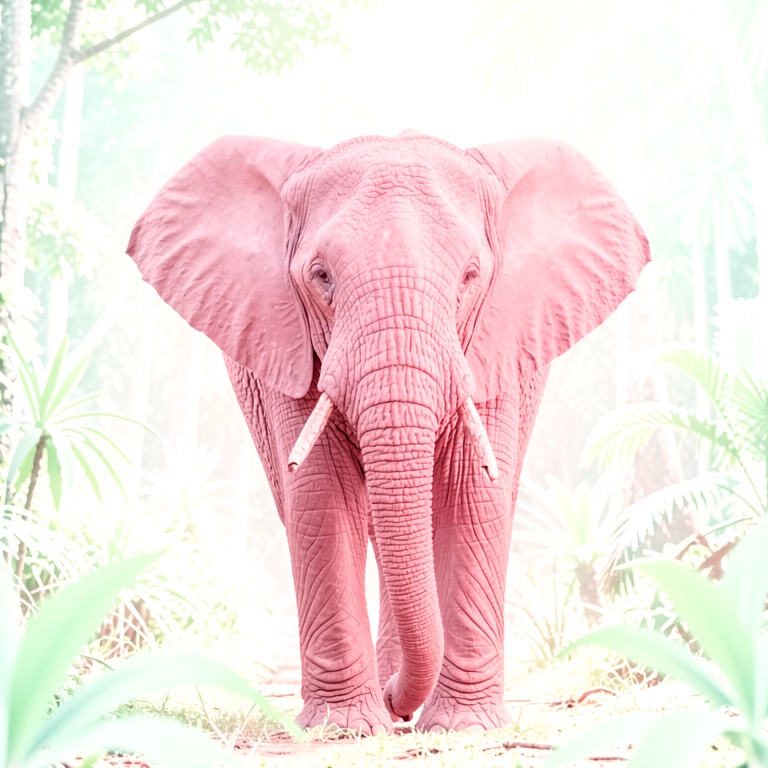} &
\includegraphics[width=0.26\linewidth]{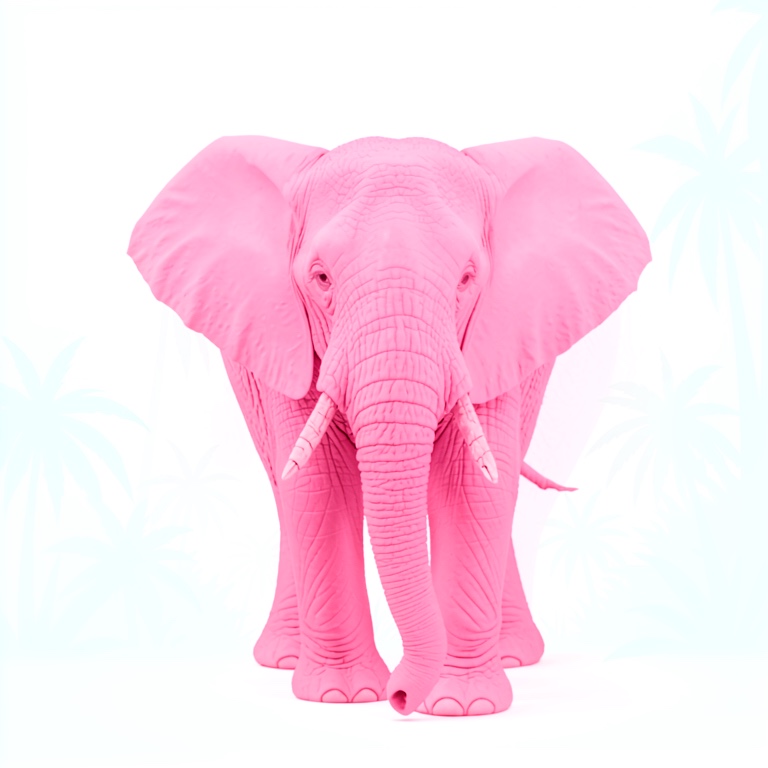}
\end{tabular}
\caption{\textbf{Mixing two banks gives graded control over color.}
Elephant and pink-elephant banks mixed for \textit{``an elephant in a
jungle''}, with the prompt fixed. Labels give
the share of pink elephants in the reference set. Outputs turn progressively
pink as that share grows.}
\label{fig:appendix_reference_composition_elephant}
\end{figure}

\paragraph{Results.}
As the share of giraffes in the reference set grows, so does the share of
giraffes among the outputs, under both classifiers, and the mean CLIP
similarity shifts from the zebra prompt to the giraffe prompt
(\Cref{fig:appendix_reference_composition_quant}). The qualitative grids
show the same progression for the savanna prompt
(\Cref{fig:appendix_reference_composition_savanna}) and for the color of an
elephant (\Cref{fig:appendix_reference_composition_elephant}).

\subsection{Q3: GenEval Comparison}
\label{app:geneval}

We describe how the GenEval banks were built and how the outputs were
scored, and then evaluate quality, diversity, and similarity to the
references beyond the alignment scores of \Cref{tab:flux_comparison}.

\subsubsection{Bank Construction}
\label{app:geneval_banks}

\begin{figure}[!htbp]
\centering
\setlength{\tabcolsep}{3pt}
\begin{tabular}{@{}c@{\hspace{4pt}}ccc@{}}
& \small \textit{bear at the right of a bench} &
\small \textit{plane above a cow} &
\small \textit{horse above a frisbee disk} \\
\rotatebox{90}{\small\hspace{1.2em}\textbf{Bank example}} &
\includegraphics[width=0.27\linewidth]{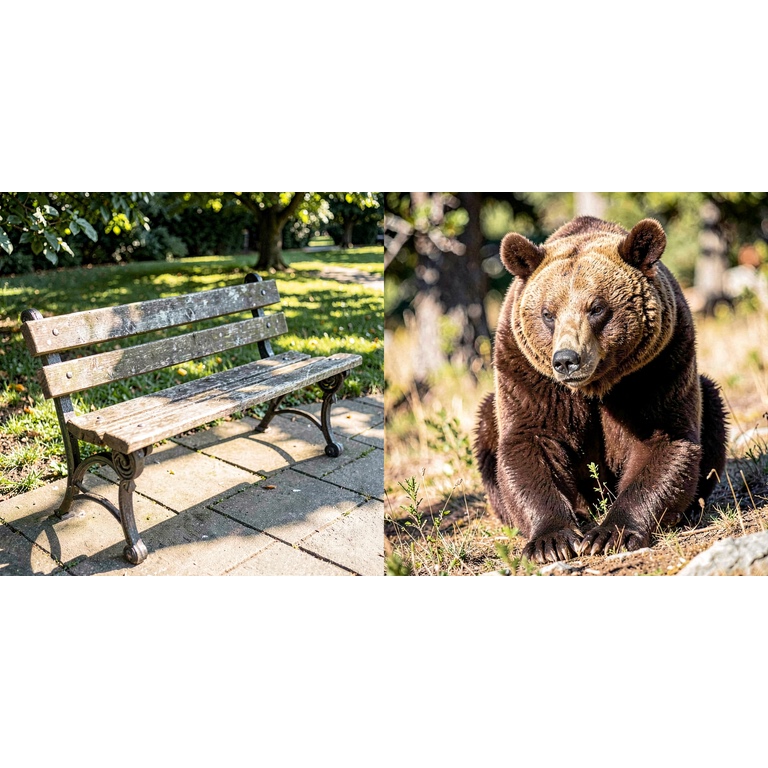} &
\includegraphics[width=0.27\linewidth]{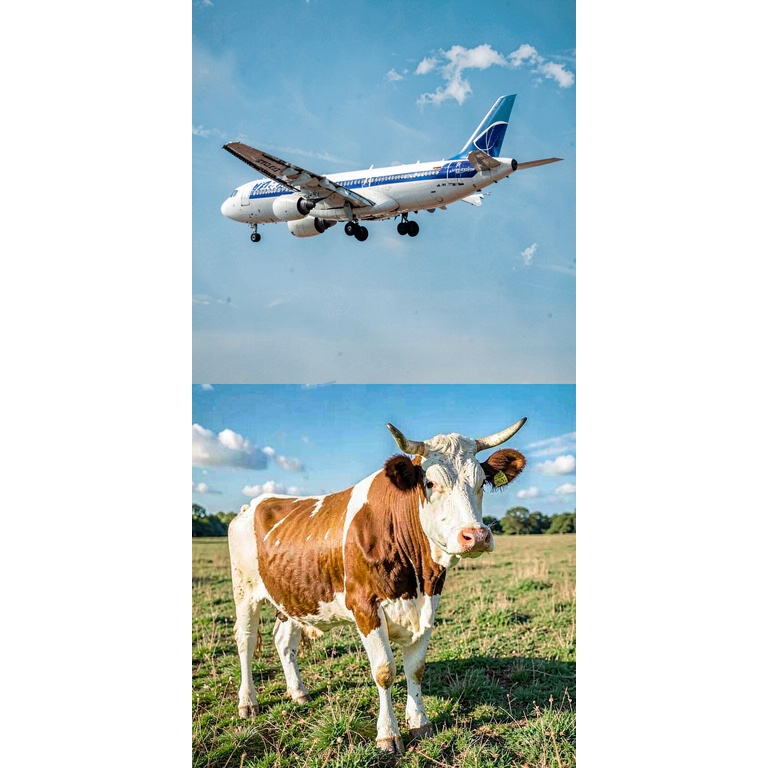} &
\includegraphics[width=0.27\linewidth]{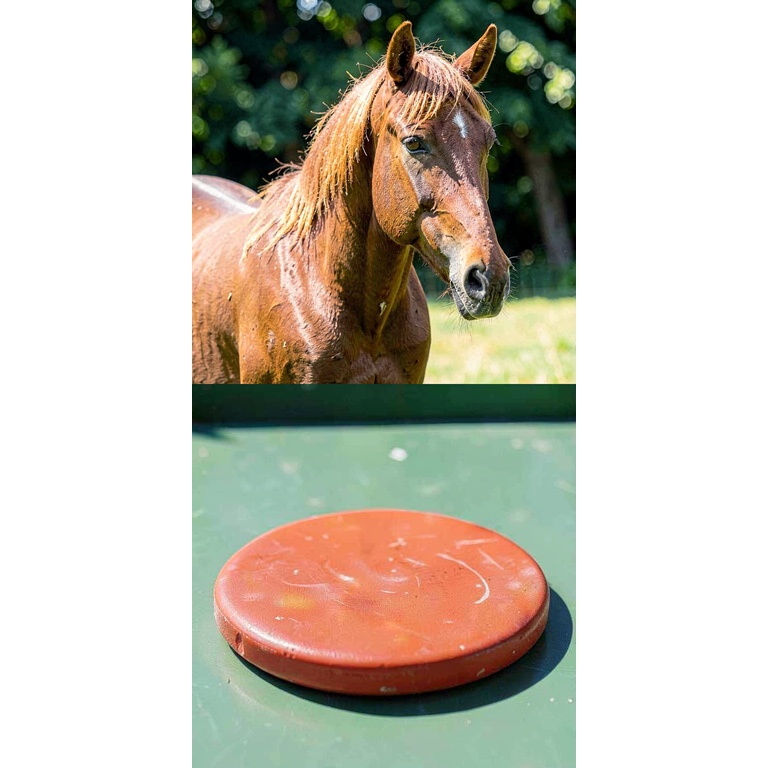} \\
\rotatebox{90}{\small\hspace{1.2em}\textbf{RMG output}} &
\includegraphics[width=0.27\linewidth]{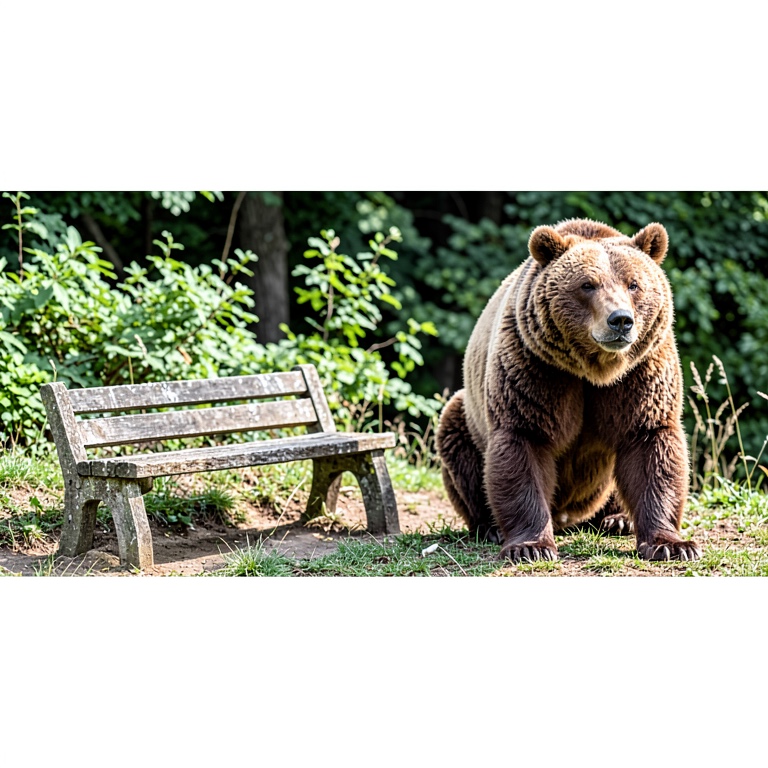} &
\includegraphics[width=0.27\linewidth]{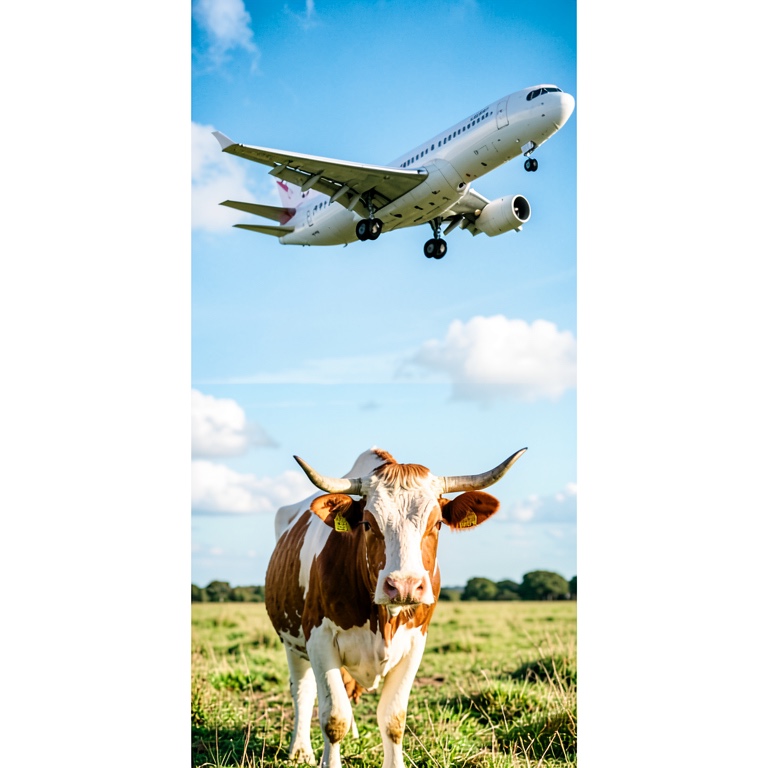} &
\includegraphics[width=0.27\linewidth]{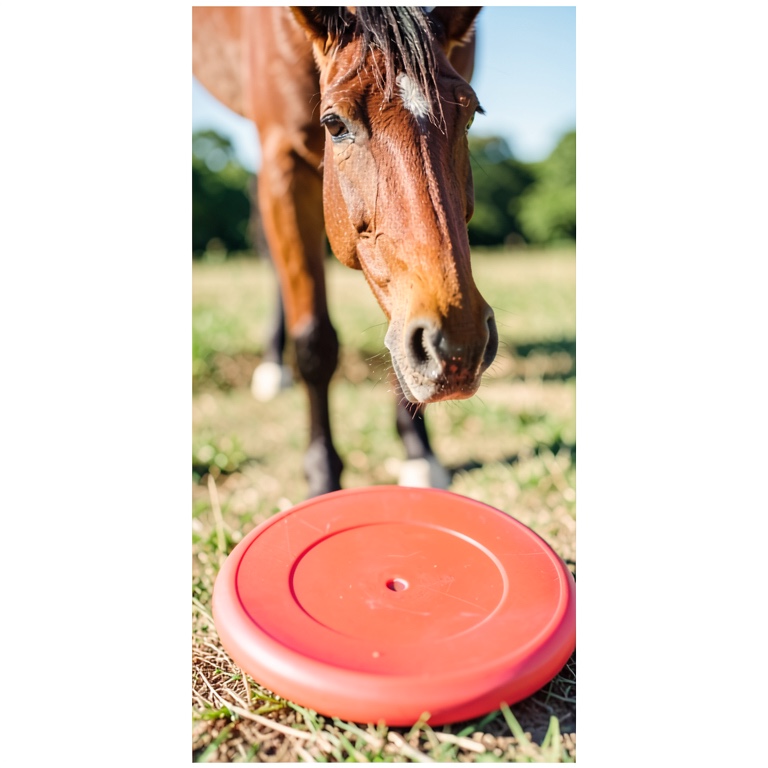}
\end{tabular}
\caption{\textbf{GenEval banks convey spatial relations through tiled
layouts.} Each column shows one image from the prompt's bank (top) and an RMG
generation (bottom). The banks encode the intended arrangement rather than
natural examples of the target scene.}
\label{fig:geneval_banks}
\end{figure}

Every GenEval prompt has its own 20-image reference bank, shared by its four
outputs rather than by the whole category. How a bank is built depends on the category.
Single-object and color banks contain direct examples of the target.
Two-object and color-attribution banks combine simpler components whose
joint effect on the posterior mean encourages the desired composition.
Position and counting banks are built by tiling single-object images into
synthetic layouts that encode the target spatial relation or count. A bank
therefore need not contain the target scene itself, and
\Cref{fig:geneval_banks} shows spatial prompts for which it supplies only
directional evidence.

\subsubsection{Evaluation Metrics and Runtime}
\label{app:geneval_evaluation_protocol}

We use all 553 GenEval prompts and keep four images per prompt for each of
the six methods in \Cref{tab:flux_comparison,tab:geneval_tradeoffs}, for
2{,}212 images per method. The search and optimization baselines spend extra
computation on candidates, which the time column reports.

\paragraph{ImageReward.}
ImageReward-v1.0~\citep{xu2023imagereward} scores each output against its
prompt. We average the four scores of each prompt and report the mean over
prompts with a 95\% bootstrap interval, obtained by resampling the 553 prompt
means with replacement 2{,}000 times and taking the 2.5th and 97.5th
percentiles.

\paragraph{Within-prompt diversity.}
We measure LPIPS~\citep{zhang2018perceptual} with the AlexNet backbone and
the learned v0.1 weights, on images resized to $256\times256$ with bicubic
interpolation and scaled to $[-1,1]$. For each prompt we average the distance
over the six pairs among its four outputs and then average over prompts, so
images from different prompts are never compared. Percentages in
\Cref{tab:geneval_tradeoffs} are relative to the unguided model. Higher LPIPS
means more perceptual variation, but artifacts raise it too, so we read it
together with ImageReward.

\paragraph{Reference similarity.}
We embed images with the SSCD copy-detection model~\citep{pizzi2022sscd},
after resizing them to $320\times320$ and normalizing them with the ImageNet
statistics, and L2-normalize the embeddings. Each output is scored by its
maximum cosine similarity to the 20 images in its prompt's bank, and we
report the mean of these scores, their 95th percentile over images, and the
fraction above 0.85. Because the baselines never see the banks, their scores
serve as controls.

\paragraph{Runtime benchmark.}
\label{app:runtime_benchmark}
The $1.02\times$ runtime reported for RMG comes from a warm-sampling
benchmark on one NVIDIA H100, with the prompt \textit{``a photo of a
bench''}, four outputs batched together, $768\times768$ resolution, 20
steps, and BF16 precision. After one warm-up run, each method is timed once.
The unguided model takes 12.34\,s for four outputs (3.09\,s per output) and
RMG takes 12.56\,s (3.14\,s per output), a ratio of $1.017\times$. Timings
include decoding and exclude bank construction, bank loading and device
transfer, and VAE encoding of the references. Since the benchmark covers a
single prompt and a single timed run, it gives no variance estimate.

\subsubsection{Quality, Diversity, and Reference Similarity}
\label{app:geneval_tradeoffs}

\Cref{tab:geneval_tradeoffs} scores the GenEval samples of
\Cref{tab:flux_comparison} with the three metrics above, ImageReward for
quality, LPIPS for within-prompt diversity, and SSCD as a check for copying.

\begin{table}[!htbp]
\centering
\caption{\textbf{RMG moves outputs toward their banks without copying them,
at some cost in diversity.} Metrics as in
\Cref{app:geneval_evaluation_protocol}, with 95\% bootstrap intervals for
ImageReward, LPIPS percentages relative to the unguided model, and the
mean, 95th percentile, and fraction above 0.85 of each output's maximum
SSCD similarity to its bank. Time is relative to the unguided model.}
\label{tab:geneval_tradeoffs}
\setlength{\tabcolsep}{3pt}
\resizebox{\linewidth}{!}{%
\begin{tabular}{lcccccc}
\toprule
Method & ImageReward $\uparrow$ & LPIPS diversity $\uparrow$ & SSCD mean & SSCD p95 & Frac.\ $>0.85$ & Time \\
\midrule
FLUX.2-klein & 1.214 [1.163, 1.265] & 0.557 (100\%) & 0.342 & 0.608 & 0.000 & $1.00\times$ \\
+ Prompt Opt. & 1.312 [1.265, 1.354] & 0.562 (101\%) & 0.331 & 0.600 & 0.000 & $7.87\times$ \\
+ Best-of-4 & 1.286 [1.240, 1.331] & 0.549 (99\%) & 0.343 & 0.610 & 0.000 & $4.07\times$ \\
+ SMC & 1.227 [1.174, 1.274] & 0.556 (100\%) & 0.342 & 0.608 & 0.000 & $6.17\times$ \\
+ ReNO & 1.250 [1.201, 1.297] & 0.457 (82\%) & 0.357 & 0.630 & 0.000 & $19.44\times$ \\
\midrule
+ RMG (ours) & 1.164 [1.108, 1.217] & 0.395 (71\%) & 0.525 & 0.711 & 0.000 & $1.02\times$ \\
\bottomrule
\end{tabular}}
\end{table}

Best-of-4, SMC, and prompt optimization keep close to 100\% of the unguided
model's diversity, as one would expect, since each of their retained images
still comes from an independent seed. The two methods that alter generation
itself are also the two that lose diversity: ReNO, which changes the
initial noise, retains 82\%, and RMG, which changes the velocity field,
retains 71\%. On quality, prompt optimization and Best-of-4 improve
ImageReward over the unguided model, whereas RMG lowers it from 1.214 to
1.164, a decrease confined to two categories (\Cref{app:quality_categories}).
On reference similarity, RMG outputs move toward their banks, with a mean
SSCD of 0.525 against 0.331 to 0.357 for the baselines, which never see the
banks. None of the 2{,}212 RMG outputs exceeds 0.85 similarity to any image
in its prompt's bank, the same holds for every baseline, and manual
inspection of the closest pairs found no evidence of direct copying. RMG
thus pulls samples toward the references, as intended, without copying them
at the default strength.

\subsubsection{ImageReward by Category}
\label{app:quality_categories}

\Cref{tab:geneval_quality_categories} breaks the ImageReward change down by
category, together with how each category's banks were built. RMG improves
the score in four of the six categories. The decrease is confined to
position and counting, whose banks were assembled by tiling single-object
images into synthetic layouts, and the white borders introduced by the
tiling were transferred along with the intended structure. We consider this
the most likely explanation (\Cref{app:limitations}).

\begin{table}[!htbp]
\centering
\small
\caption{\textbf{The ImageReward decrease is confined to the two
tiled-bank categories.} ImageReward by GenEval category for the unguided
model and RMG.
The last column gives each category's contribution to the change in the
aggregate score. Means, $\Delta$, and contributions are rounded
independently, so displayed differences can vary by 0.001.}
\label{tab:geneval_quality_categories}
\setlength{\tabcolsep}{4pt}
\begin{tabular}{llrrrr}
\toprule
Category & Bank construction & Unguided & RMG & $\Delta$ & \shortstack{Contribution\\to overall $\Delta$} \\
\midrule
Single object & Direct & 0.625 & 0.676 & +0.051 & +0.007 \\
Two objects & Composite, co-presence & 1.418 & 1.598 & +0.181 & +0.032 \\
Colors & Direct & 1.110 & 1.233 & +0.123 & +0.021 \\
Color attribution & Composite, binding & 1.552 & 1.610 & +0.058 & +0.011 \\
Position & Composite, spatial layout & 1.421 & 1.181 & $-0.239$ & $-0.043$ \\
Counting & Composite, cardinality & 0.994 & 0.454 & $-0.541$ & $-0.078$ \\
\midrule
All & & 1.214 & 1.164 & $-0.050$ & $-0.050$ \\
\bottomrule
\end{tabular}
\end{table}

\subsection{Q4: Semi-Parametric Guidance}
\label{app:spg}

We describe the architecture, training, and AFHQv2 setup of SPG, and then
study how its diversity depends on the size of the reference set and whether
it avoids transferring nuisance attributes.

\subsubsection{Architecture and Training}
\label{app:spg_training}

SPG follows \Cref{sec:spg}, and we give here the details omitted there.
Images and references are encoded with a frozen VAE encoder $\mathcal{E}$,
and the interpolants $x_t = (1-t)x_0 + tx_1$ use $x_0 \sim \mathcal{N}(0,I)$
and $t$ drawn from $[\epsilon, 1-\epsilon)$ with $\epsilon = 10^{-3}$, which
keeps the $(1-t)^{-2}$ weight of $\mathcal{L}_\mu$ finite. The anchor of
\Cref{eq:spg_anchor} is computed patchwise, with queries from the patches of
$x_t$, keys from the patches of the references, and the reference patches
themselves as values. Besides the leave-one-out structure, the reference set
is randomly masked during training, which discourages reliance on any
individual reference.

The two losses play complementary roles. $\mathcal{L}_\mu$ updates the anchor
and the refiner jointly through the full prediction of
\Cref{eq:spg_prediction}, while $\mathcal{L}_{\mathrm{ref}}$ trains the
refiner on the residual with the anchor held fixed. Since the refiner output
is added back to $\bar{x}$ in the forward pass, this residual signal moves
the prediction toward the ground truth.

\paragraph{Why the model uses the reference set.}
As \Cref{sec:spg} notes, the objective does not force the model to use
$\mathcal{R}$, and the incentive comes from the architecture. This incentive
is an inductive bias rather than a guarantee, since the refiner still sees
$x_t$ and a sufficiently flexible predictor could fall back to standard flow
matching. Whether the trained model relies on its references is therefore an
empirical question, which the noise-bank, anchor-only, and bank-swap tests
of \Cref{sec:spg} answer.

\subsubsection{AFHQv2 Setup}
\label{app:spg_setup}

We train and evaluate SPG on the full AFHQv2 training split, with the
resolution and VAE of \Cref{sec:experimental_setup}. The primary reference
bank contains all of its dog and cat images.

\paragraph{Architecture.} The reference-attention module is a single
cross-attention block with patchwise retrieval (patch size 2, decoupled
embedding), 8 heads, query--key dimension 768, and MLP ratio 1, trained with
reference-set dropout $p=0.1$. The gates $g_t$ and $\eta_t$ are learned MLPs
initialized at $0.5$. The residual refiner is a DiT-style transformer with 11
blocks, embedding dimension 768, 12 heads, patch size 2, and MLP ratio 4.

\paragraph{Optimization.} We train for 10{,}000 steps on four A100 GPUs with
batch size 64 in bf16 mixed precision, using the Lion
optimizer~\citep{chen2023symbolic} with learning rate $10^{-4}$,
$\beta_1=0.9$, $\beta_2=0.999$, and no weight decay. We keep an exponential
moving average of the weights with decay $0.9999$, clip gradients at norm
$1.0$, and weight the refiner penalty by $0.1$.

\paragraph{Evaluation.} FID and KID are computed with
the clean-FID implementation~\citep{parmar2022aliased} against real images from the
training split. Class frequencies are estimated with CLIP over 10{,}000
generated samples, using the prompts \textit{``a photo of a dog''} and
\textit{``a photo of a cat''}.

\subsubsection{Diversity and Reference-Set Size}
\label{app:spg_diversity}

\begin{figure}[!htbp]
\centering
\includegraphics[width=0.55\textwidth]{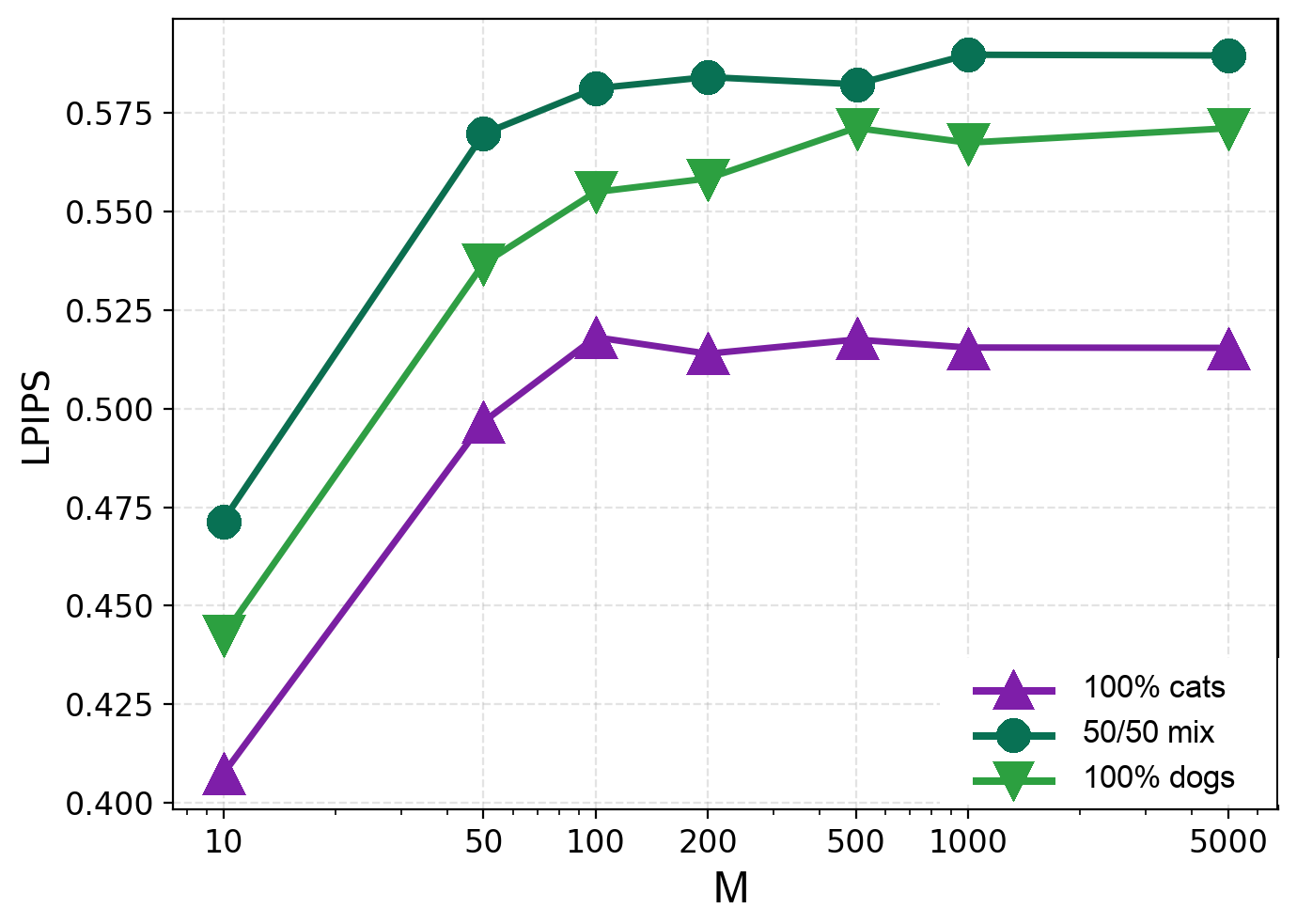}
\caption{\textbf{SPG diversity saturates at around a hundred references.}
Average pairwise LPIPS against reference-set size $M$ for cat, dog, and mixed
banks.}
\label{fig:spg_lpips_vs_n}
\end{figure}

We measure how the diversity of SPG samples depends on the number of
references available at inference time. For each reference-set size $M$ we
sample from the same trained model and compute the average pairwise LPIPS
between generated images. LPIPS alone cannot tell novel generations apart
from variation among the references, so we read it together with the
nearest-neighbor comparison in \Cref{fig:spg_qual}. Diversity rises quickly
up to roughly a hundred references and then levels off, for cat, dog, and
mixed banks alike (\Cref{fig:spg_lpips_vs_n}). Larger reference sets
therefore support more varied outputs while keeping the reference-conditioned
control reported in the main text.

\subsubsection{Suppressing Nuisance Attributes}
\label{app:spg_background_nocopy}

\begin{figure}[!htbp]
\centering
\begin{tabular}{@{}c@{}}
\small \textbf{Reference bank} \\
\includegraphics[width=0.92\linewidth]{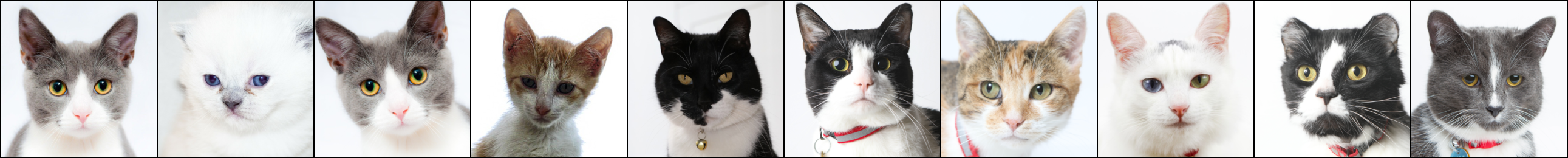} \\[0.5em]
\begin{tabular}{cc}
\small \textbf{RMG generation} & \small \textbf{SPG generation} \\
\includegraphics[width=0.26\linewidth]{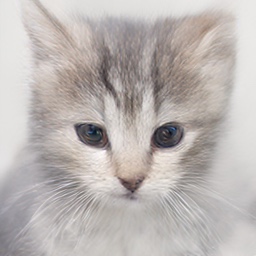} &
\includegraphics[width=0.26\linewidth]{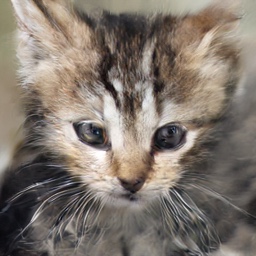}
\end{tabular}
\end{tabular}
\caption{\textbf{SPG keeps the object guidance without copying a shared
background.} Every image in this bank has a white background. RMG transfers
this nuisance property to the generated sample, whereas SPG does not.}
\label{fig:spg_background_nocopy}
\end{figure}

RMG transfers every attribute its references share (\Cref{sec:spg}). To test
whether SPG avoids this, we build a reference set in which every image has a
white background. RMG transfers the background together with the object
appearance, whereas the learned anchor and refiner of SPG keep the
object-level guidance without reproducing the background
(\Cref{fig:spg_background_nocopy}).

\section{Ablations}
\label{app:ablations}

\subsection{Guidance Schedule}
\label{app:schedule_ablation}
\label{app:schedule_form}

\begin{figure}[!htbp]
\centering
\includegraphics[width=0.5\textwidth]{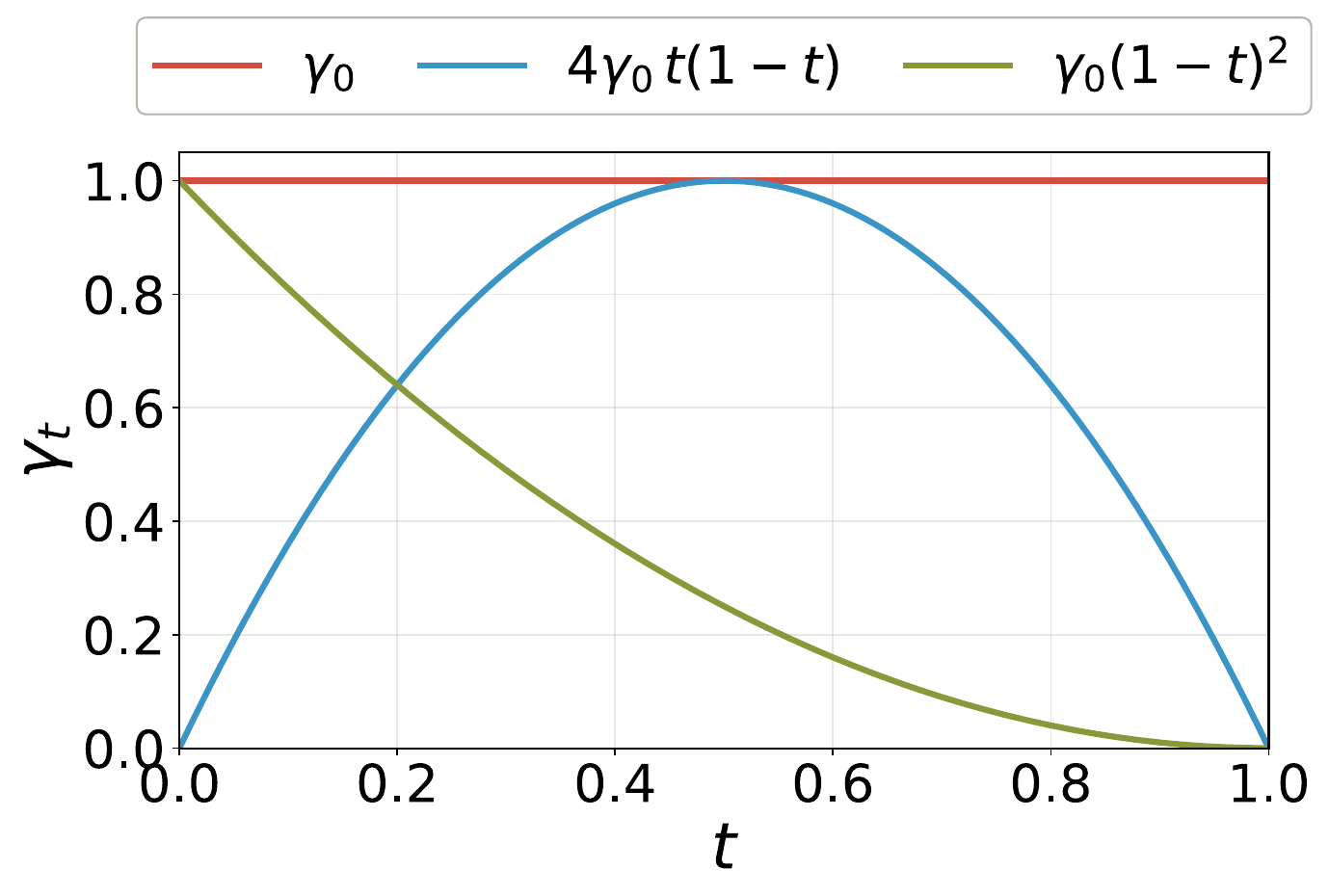}
\caption{\textbf{The three guidance schedules.} Constant, quadratic decay,
and bell schedules for $\gamma_0 = 1$, shown before the shared cutoff at
$t=0.85$.}
\label{fig:beta_schedules}
\end{figure}

The correction in \Cref{eq:practical_rmg} is scaled by $\gamma_t$, which
sets both the strength and the timing of the intervention. We ablate the
functional form of the schedule and its peak strength $\gamma_0$, comparing
the three forms in \Cref{fig:beta_schedules}. The constant schedule
$\gamma_t=\gamma_0$ applies the same strength at every step. Quadratic
decay, $\gamma_t=\gamma_0(1-t)^2$, front-loads guidance and vanishes as
$t\to1$, which cancels the divergence of the correction term. It is the
schedule used in all main experiments. The bell schedule,
$\gamma_t=4\gamma_0 t(1-t)$, suppresses guidance at both ends and
concentrates it around the midpoint, leaving the earliest steps, where the
overall structure is still forming, untouched.

All runs share the cutoff at $t \geq 0.85$ of \Cref{app:flux_sampling},
without which high-strength schedules, the constant one in particular,
become numerically unstable near the end of the trajectory. All other
settings follow \Cref{tab:hyperparams}.

\subsection{Guidance Strength \texorpdfstring{$\gamma_0$}{gamma0} and Schedule Shape}
\label{app:beta0_sweep}

For each schedule we sweep $\gamma_0 \in
\{0.01, 0.1, 0.2, 0.3, 0.4, 0.5, 1.0, 1.5, 2.0\}$ in two settings: the
pink-elephant bank with \textit{``an elephant in a jungle''} and the giraffe
bank with \textit{``an animal in a savanna''}. The geometric-mixture
derivation of \Cref{sec:rmg} uses a constant $\lambda\in[0,1]$, but the
practical rule admits larger values. We treat $\gamma_0>1$ as an
extrapolated guidance strength and use it to stress-test stability and
control, not as a mixture coefficient. The sweep is separate from the
task-specific settings in \Cref{tab:hyperparams}, where the savanna bank swap
uses $\gamma_0=1.0$.

\begin{figure}[!htbp]
\centering
\setlength{\tabcolsep}{1pt}
\renewcommand{\arraystretch}{0.8}
{\small\textit{``an elephant in a jungle''}, pink-elephant bank}\\[0.2em]
\begin{tabular}{@{}ccccc@{}}
\scriptsize $\gamma_0=0$ & \scriptsize $\gamma_0=0.01$ & \scriptsize $\gamma_0=0.1$ & \scriptsize $\gamma_0=0.2$ & \scriptsize $\gamma_0=0.3$ \\
\includegraphics[width=0.13\linewidth]{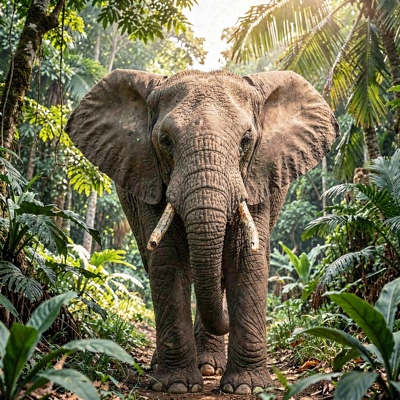} &
\includegraphics[width=0.13\linewidth]{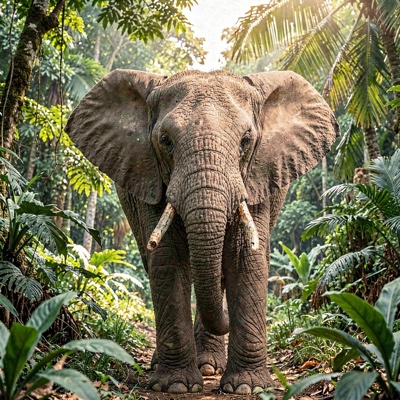} &
\includegraphics[width=0.13\linewidth]{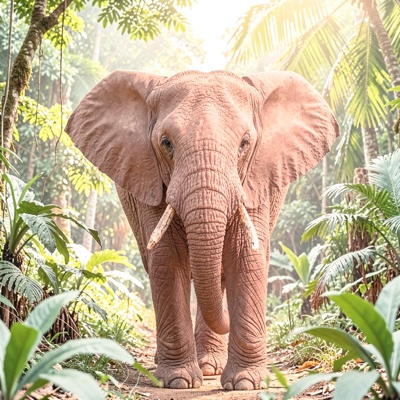} &
\includegraphics[width=0.13\linewidth]{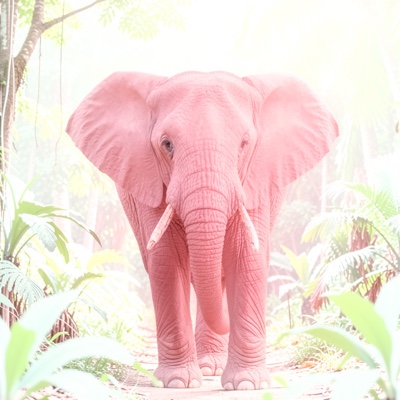} &
\includegraphics[width=0.13\linewidth]{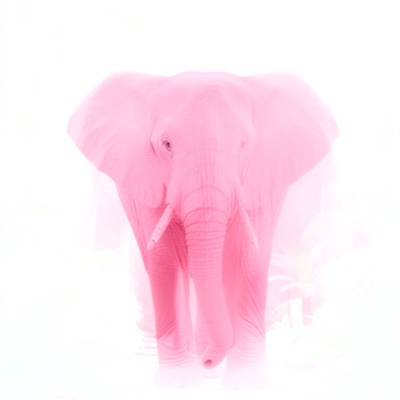} \\
\scriptsize $\gamma_0=0.4$ & \scriptsize $\gamma_0=0.5$ & \scriptsize $\gamma_0=1.0$ & \scriptsize $\gamma_0=1.5$ & \scriptsize $\gamma_0=2.0$ \\
\includegraphics[width=0.13\linewidth]{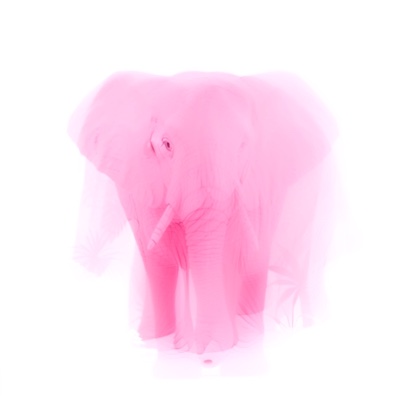} &
\includegraphics[width=0.13\linewidth]{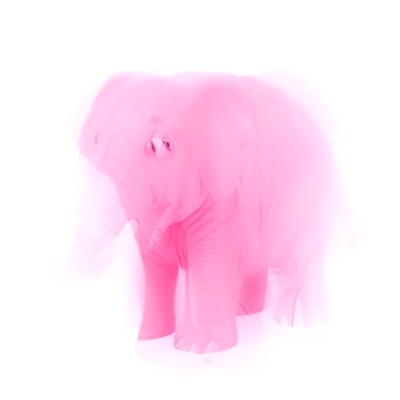} &
\includegraphics[width=0.13\linewidth]{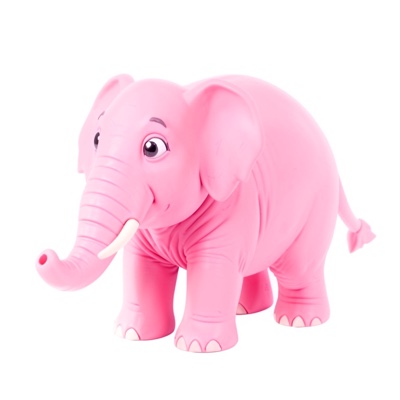} &
\includegraphics[width=0.13\linewidth]{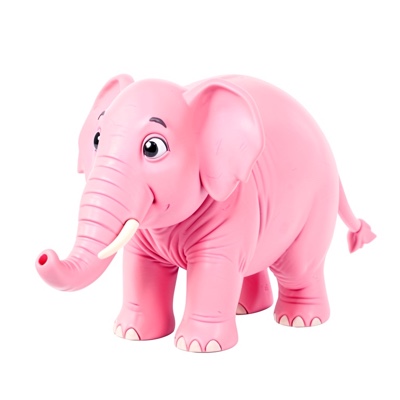} &
\includegraphics[width=0.13\linewidth]{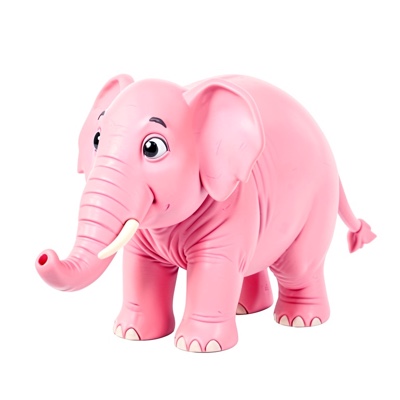}
\end{tabular}\\[0.3em]
{\small\textit{``an animal in a savanna''}, giraffe bank}\\[0.2em]
\begin{tabular}{@{}ccccc@{}}
\scriptsize $\gamma_0=0$ & \scriptsize $\gamma_0=0.01$ & \scriptsize $\gamma_0=0.1$ & \scriptsize $\gamma_0=0.2$ & \scriptsize $\gamma_0=0.3$ \\
\includegraphics[width=0.13\linewidth]{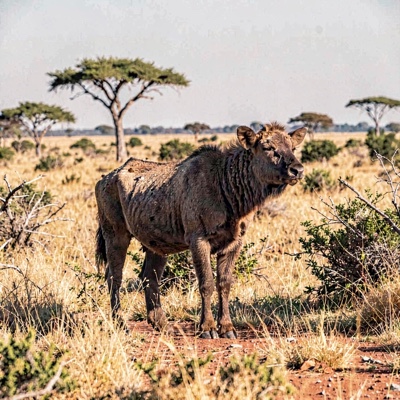} &
\includegraphics[width=0.13\linewidth]{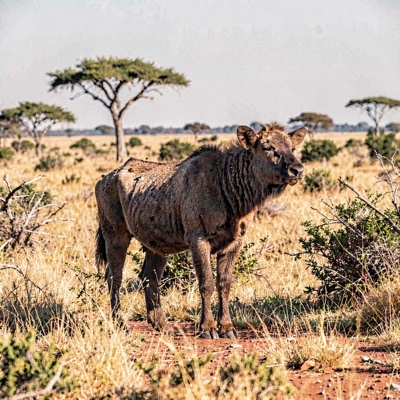} &
\includegraphics[width=0.13\linewidth]{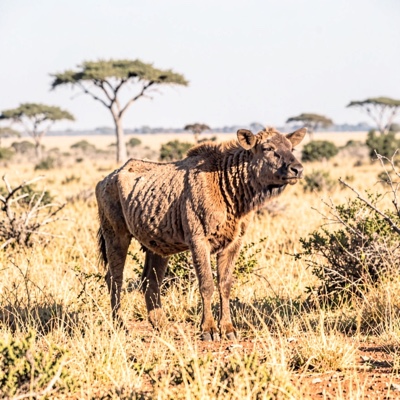} &
\includegraphics[width=0.13\linewidth]{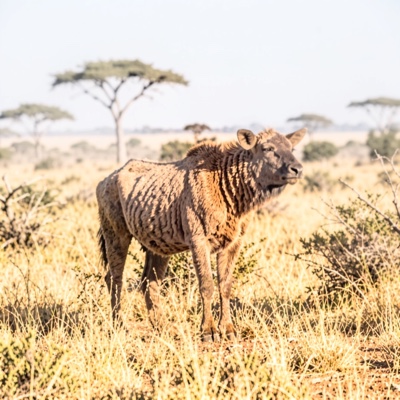} &
\includegraphics[width=0.13\linewidth]{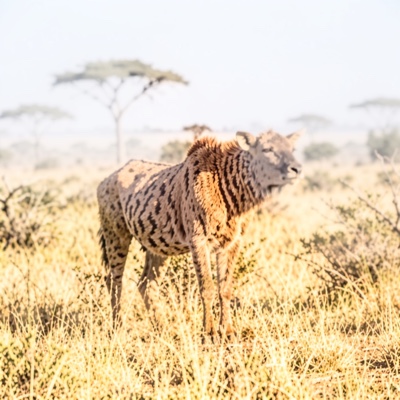} \\
\scriptsize $\gamma_0=0.4$ & \scriptsize $\gamma_0=0.5$ & \scriptsize $\gamma_0=1.0$ & \scriptsize $\gamma_0=1.5$ & \scriptsize $\gamma_0=2.0$ \\
\includegraphics[width=0.13\linewidth]{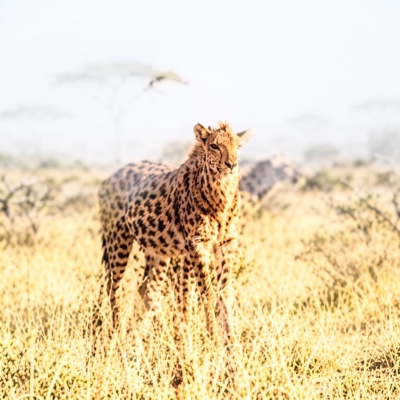} &
\includegraphics[width=0.13\linewidth]{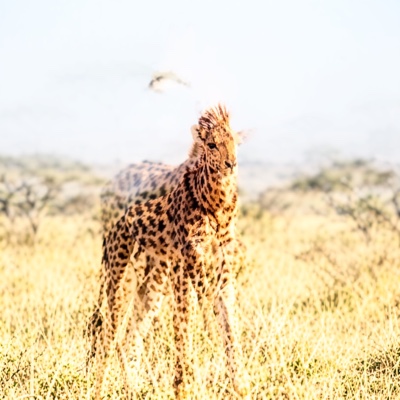} &
\includegraphics[width=0.13\linewidth]{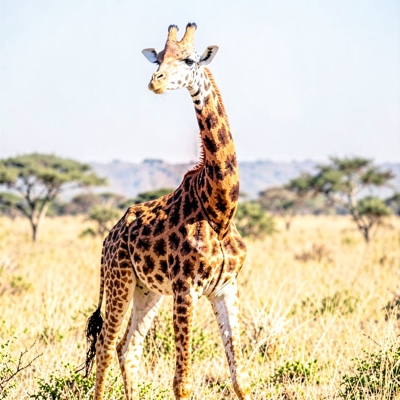} &
\includegraphics[width=0.13\linewidth]{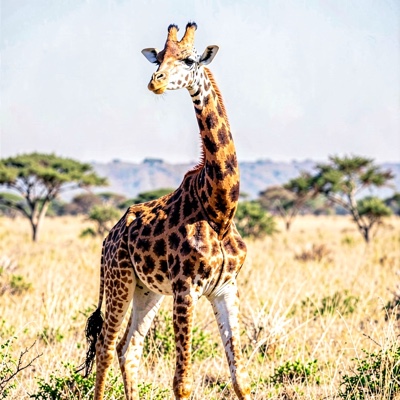} &
\includegraphics[width=0.13\linewidth]{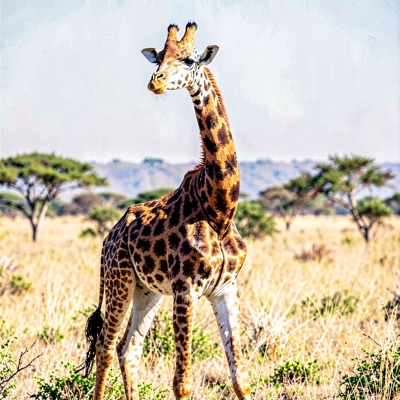}
\end{tabular}
\caption{\textbf{A constant schedule breaks down at large strengths.}
Guidance-strength sweep with $\gamma_0=0$ the unguided model. Above
$\gamma_0\approx1$, late-time artifacts appear as oversaturated colors and
structural distortion.}
\label{fig:beta_ablation_constant}
\end{figure}

\begin{figure}[!htbp]
\centering
\setlength{\tabcolsep}{1pt}
\renewcommand{\arraystretch}{0.8}
{\small\textit{``an elephant in a jungle''}, pink-elephant bank}\\[0.2em]
\begin{tabular}{@{}ccccc@{}}
\scriptsize $\gamma_0=0$ & \scriptsize $\gamma_0=0.01$ & \scriptsize $\gamma_0=0.1$ & \scriptsize $\gamma_0=0.2$ & \scriptsize $\gamma_0=0.3$ \\
\includegraphics[width=0.13\linewidth]{images/ablation_beta/elephant_pink/baseline.jpg} &
\includegraphics[width=0.13\linewidth]{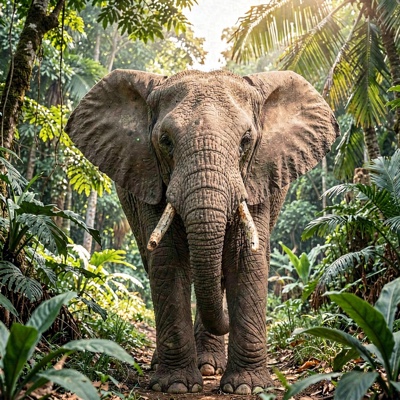} &
\includegraphics[width=0.13\linewidth]{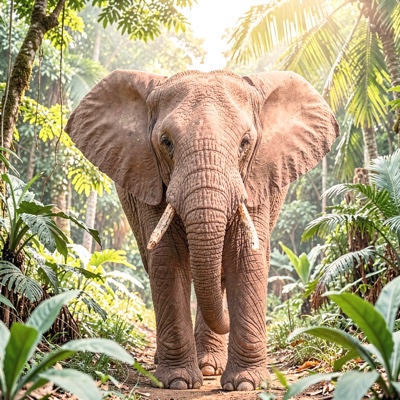} &
\includegraphics[width=0.13\linewidth]{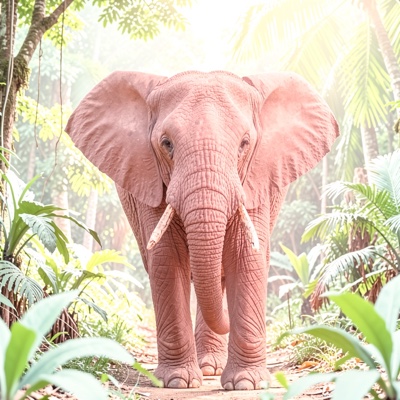} &
\includegraphics[width=0.13\linewidth]{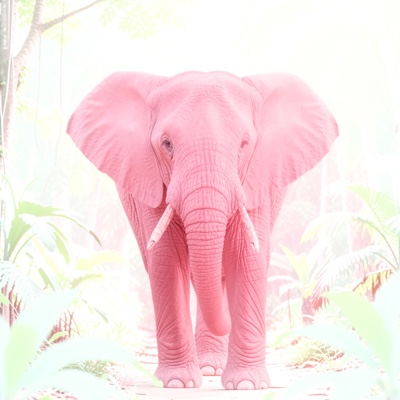} \\
\scriptsize $\gamma_0=0.4$ & \scriptsize $\gamma_0=0.5$ & \scriptsize $\gamma_0=1.0$ & \scriptsize $\gamma_0=1.5$ & \scriptsize $\gamma_0=2.0$ \\
\includegraphics[width=0.13\linewidth]{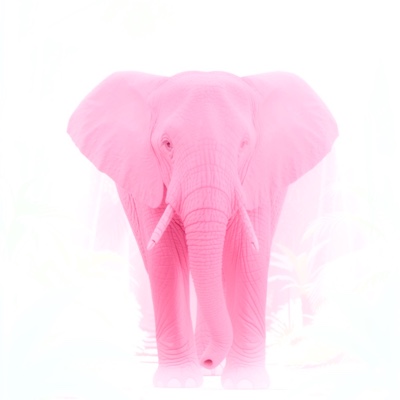} &
\includegraphics[width=0.13\linewidth]{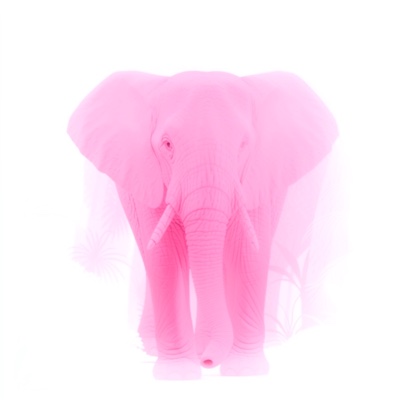} &
\includegraphics[width=0.13\linewidth]{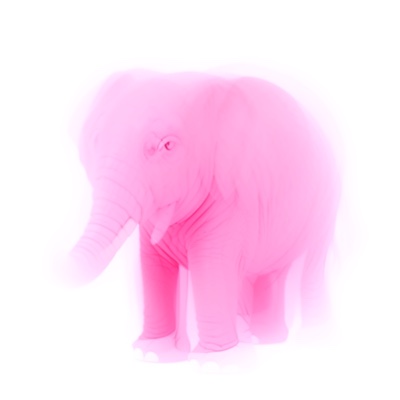} &
\includegraphics[width=0.13\linewidth]{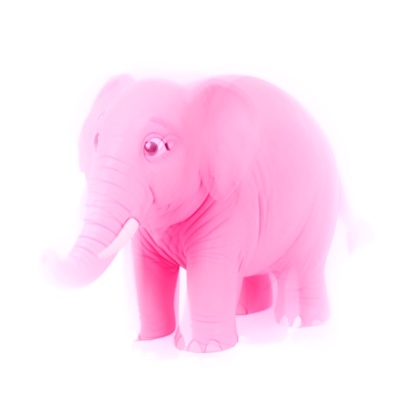} &
\includegraphics[width=0.13\linewidth]{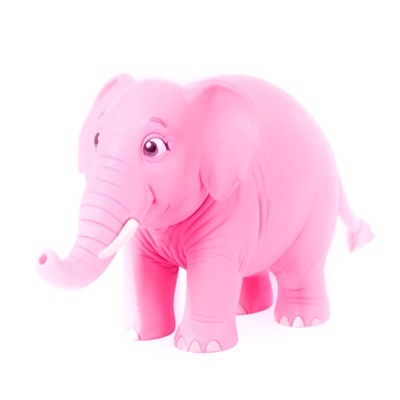}
\end{tabular}\\[0.3em]
{\small\textit{``an animal in a savanna''}, giraffe bank}\\[0.2em]
\begin{tabular}{@{}ccccc@{}}
\scriptsize $\gamma_0=0$ & \scriptsize $\gamma_0=0.01$ & \scriptsize $\gamma_0=0.1$ & \scriptsize $\gamma_0=0.2$ & \scriptsize $\gamma_0=0.3$ \\
\includegraphics[width=0.13\linewidth]{images/ablation_beta/savanna_giraffe/baseline.jpg} &
\includegraphics[width=0.13\linewidth]{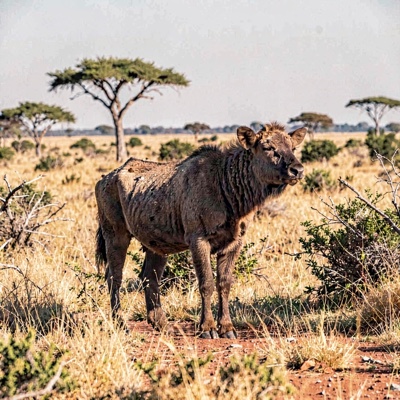} &
\includegraphics[width=0.13\linewidth]{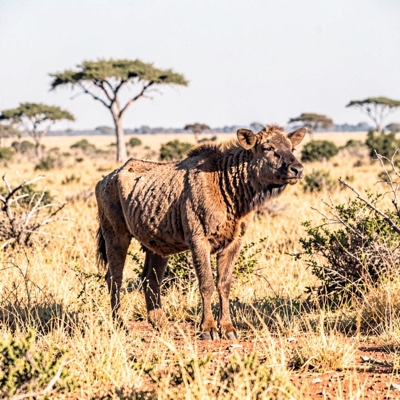} &
\includegraphics[width=0.13\linewidth]{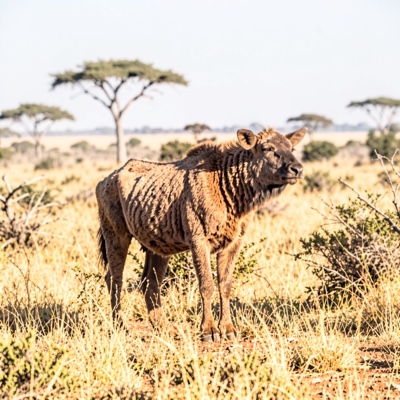} &
\includegraphics[width=0.13\linewidth]{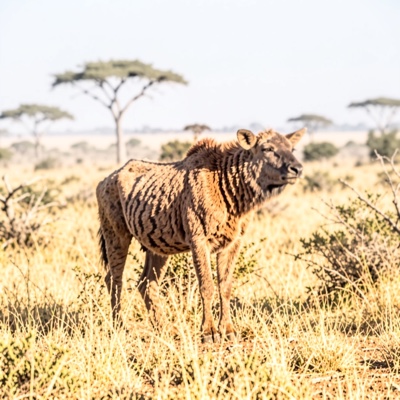} \\
\scriptsize $\gamma_0=0.4$ & \scriptsize $\gamma_0=0.5$ & \scriptsize $\gamma_0=1.0$ & \scriptsize $\gamma_0=1.5$ & \scriptsize $\gamma_0=2.0$ \\
\includegraphics[width=0.13\linewidth]{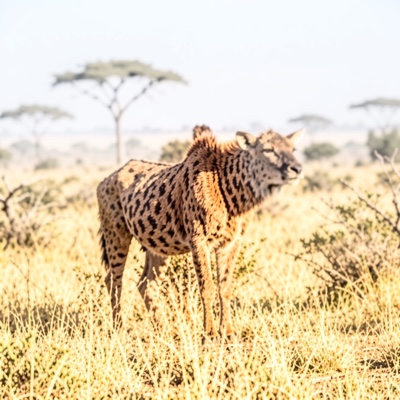} &
\includegraphics[width=0.13\linewidth]{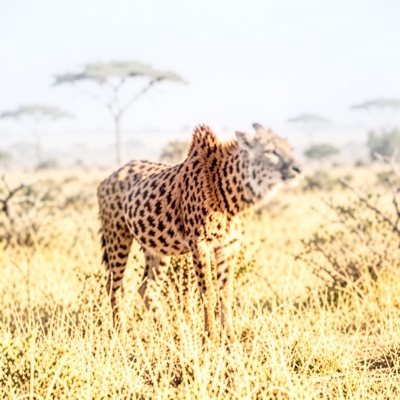} &
\includegraphics[width=0.13\linewidth]{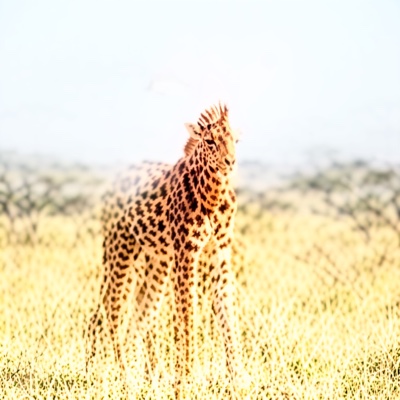} &
\includegraphics[width=0.13\linewidth]{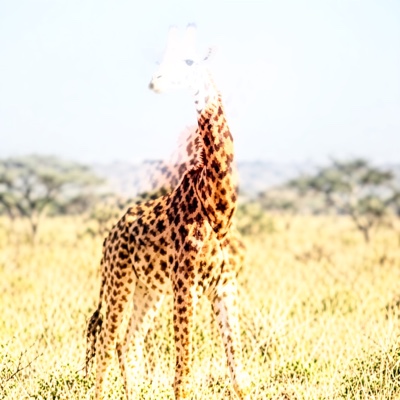} &
\includegraphics[width=0.13\linewidth]{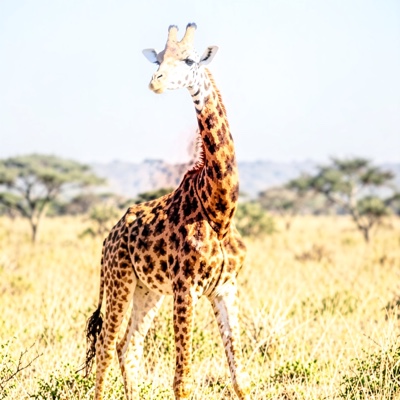}
\end{tabular}
\caption{\textbf{The bell schedule stays stable at large strengths.}
Guidance-strength sweep with $\gamma_0=0$ the unguided model. Concentrating
guidance around the midpoint delays attribute transfer slightly.}
\label{fig:beta_ablation_bell}
\end{figure}

\begin{figure}[!htbp]
\centering
\setlength{\tabcolsep}{1pt}
\renewcommand{\arraystretch}{0.8}
{\small\textit{``an elephant in a jungle''}, pink-elephant bank}\\[0.2em]
\begin{tabular}{@{}ccccc@{}}
\scriptsize $\gamma_0=0$ & \scriptsize $\gamma_0=0.01$ & \scriptsize $\gamma_0=0.1$ & \scriptsize $\gamma_0=0.2$ & \scriptsize $\gamma_0=0.3$ \\
\includegraphics[width=0.13\linewidth]{images/ablation_beta/elephant_pink/baseline.jpg} &
\includegraphics[width=0.13\linewidth]{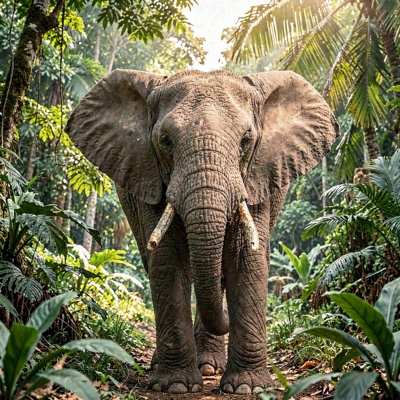} &
\includegraphics[width=0.13\linewidth]{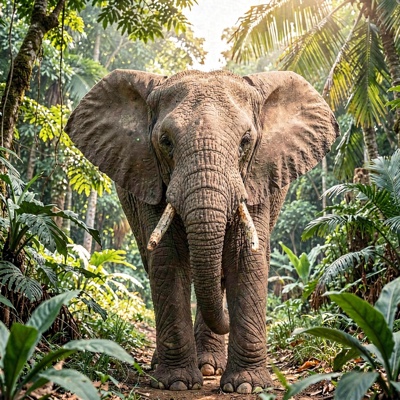} &
\includegraphics[width=0.13\linewidth]{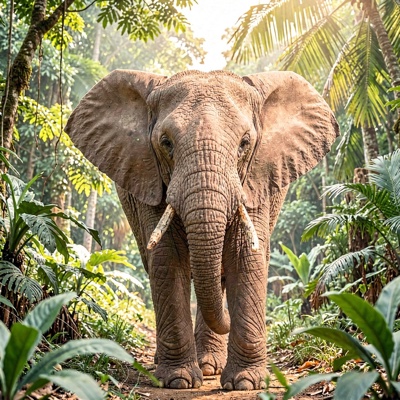} &
\includegraphics[width=0.13\linewidth]{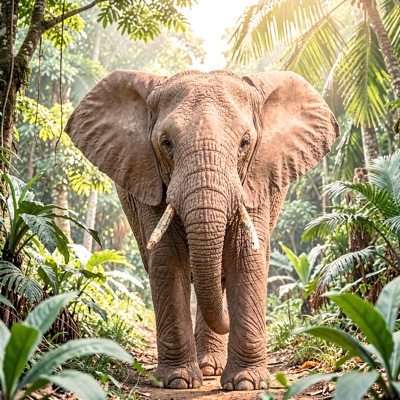} \\
\scriptsize $\gamma_0=0.4$ & \scriptsize $\gamma_0=0.5$ & \scriptsize $\gamma_0=1.0$ & \scriptsize $\gamma_0=1.5$ & \scriptsize $\gamma_0=2.0$ \\
\includegraphics[width=0.13\linewidth]{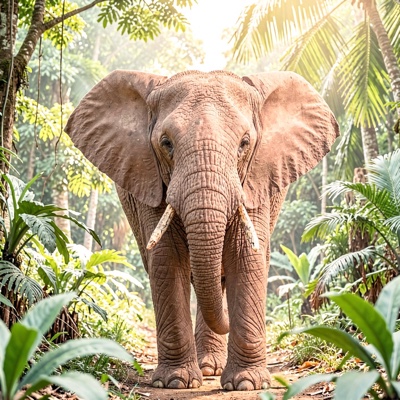} &
\includegraphics[width=0.13\linewidth]{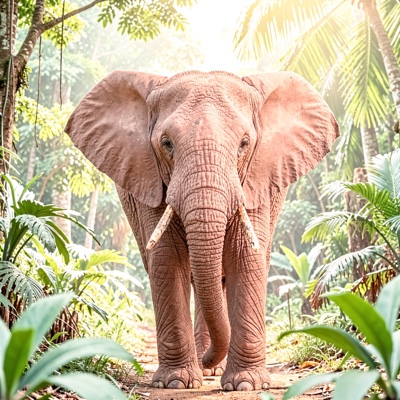} &
\includegraphics[width=0.13\linewidth]{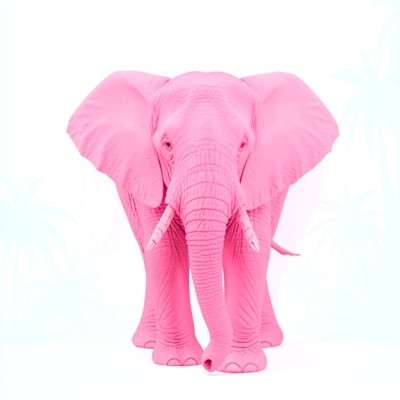} &
\includegraphics[width=0.13\linewidth]{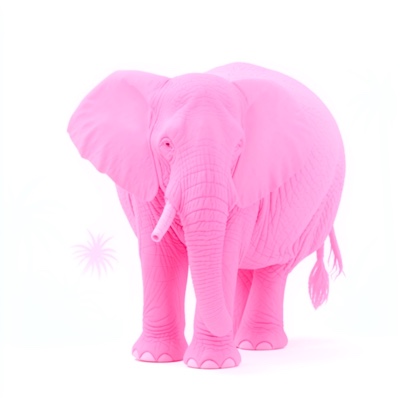} &
\includegraphics[width=0.13\linewidth]{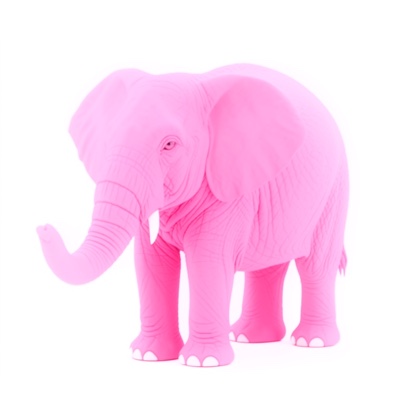}
\end{tabular}\\[0.3em]
{\small\textit{``an animal in a savanna''}, giraffe bank}\\[0.2em]
\begin{tabular}{@{}ccccc@{}}
\scriptsize $\gamma_0=0$ & \scriptsize $\gamma_0=0.01$ & \scriptsize $\gamma_0=0.1$ & \scriptsize $\gamma_0=0.2$ & \scriptsize $\gamma_0=0.3$ \\
\includegraphics[width=0.13\linewidth]{images/ablation_beta/savanna_giraffe/baseline.jpg} &
\includegraphics[width=0.13\linewidth]{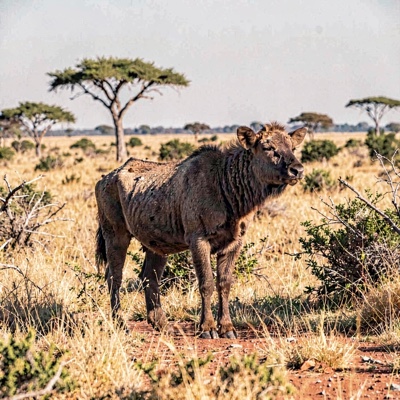} &
\includegraphics[width=0.13\linewidth]{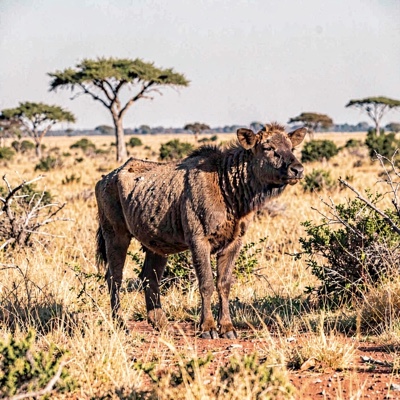} &
\includegraphics[width=0.13\linewidth]{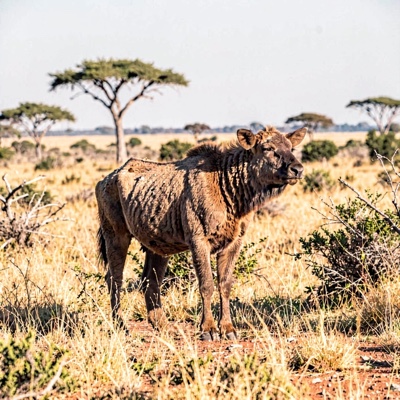} &
\includegraphics[width=0.13\linewidth]{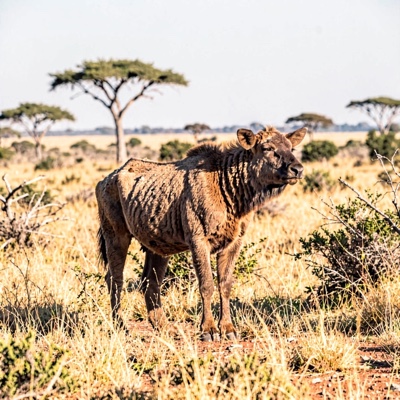} \\
\scriptsize $\gamma_0=0.4$ & \scriptsize $\gamma_0=0.5$ & \scriptsize $\gamma_0=1.0$ & \scriptsize $\gamma_0=1.5$ & \scriptsize $\gamma_0=2.0$ \\
\includegraphics[width=0.13\linewidth]{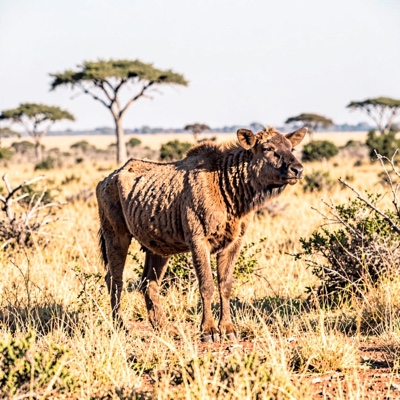} &
\includegraphics[width=0.13\linewidth]{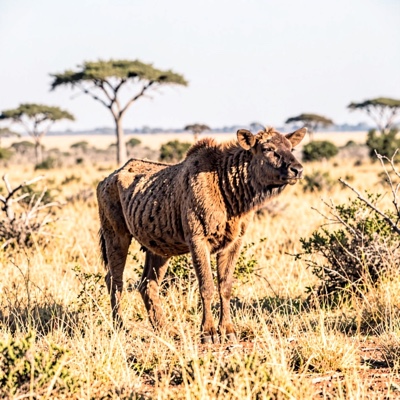} &
\includegraphics[width=0.13\linewidth]{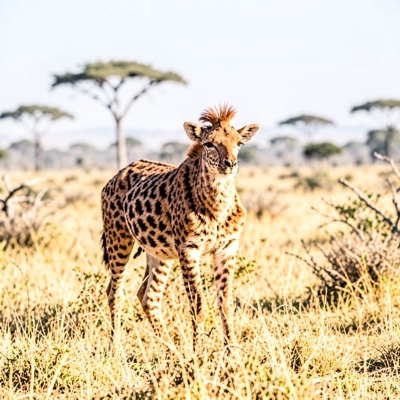} &
\includegraphics[width=0.13\linewidth]{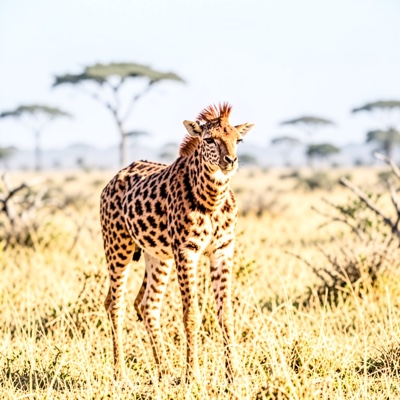} &
\includegraphics[width=0.13\linewidth]{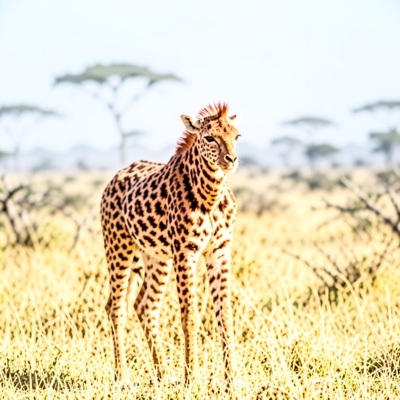}
\end{tabular}
\caption{\textbf{Quadratic decay gives the cleanest transfer across
strengths.} Guidance-strength sweep with the schedule of the main
experiments, with $\gamma_0=0$ the unguided model. Decaying to zero at $t=1$
cancels the late-time divergence.}
\label{fig:beta_ablation_quadratic}
\end{figure}

The three schedules differ mainly in how they behave at high strength. The
constant schedule transfers the target attribute cleanly at moderate
$\gamma_0$, but above $\gamma_0 \approx 1$ it introduces artifacts near
$t=1$, visible as oversaturated colors and structural distortion
(\Cref{fig:beta_ablation_constant}). The bell schedule transfers the
attribute slightly later in the sweep and remains stable at larger
$\gamma_0$ (\Cref{fig:beta_ablation_bell}). Quadratic decay gives the
cleanest transfer with the fewest late-time artifacts across the whole range
(\Cref{fig:beta_ablation_quadratic}), which is why we use it in the main
experiments.

\subsection{Reference-Set Size}
\label{app:dataset_size_ablation}

We next vary the number of references while keeping the prompt, model, and
schedule fixed. Reference sets of size
$M \in \{1, 2, 4, 8, 16, 32, 64, 128\}$ are drawn uniformly at random from a
single attribute-aligned bank, with several subsets per size, and every
subset is used with the same fixed set of seeds. Diversity is the average
pairwise LPIPS among the generated images, and the error bars show the
spread across subsets.

\begin{figure}[!htbp]
\centering
\includegraphics[width=0.5\textwidth]{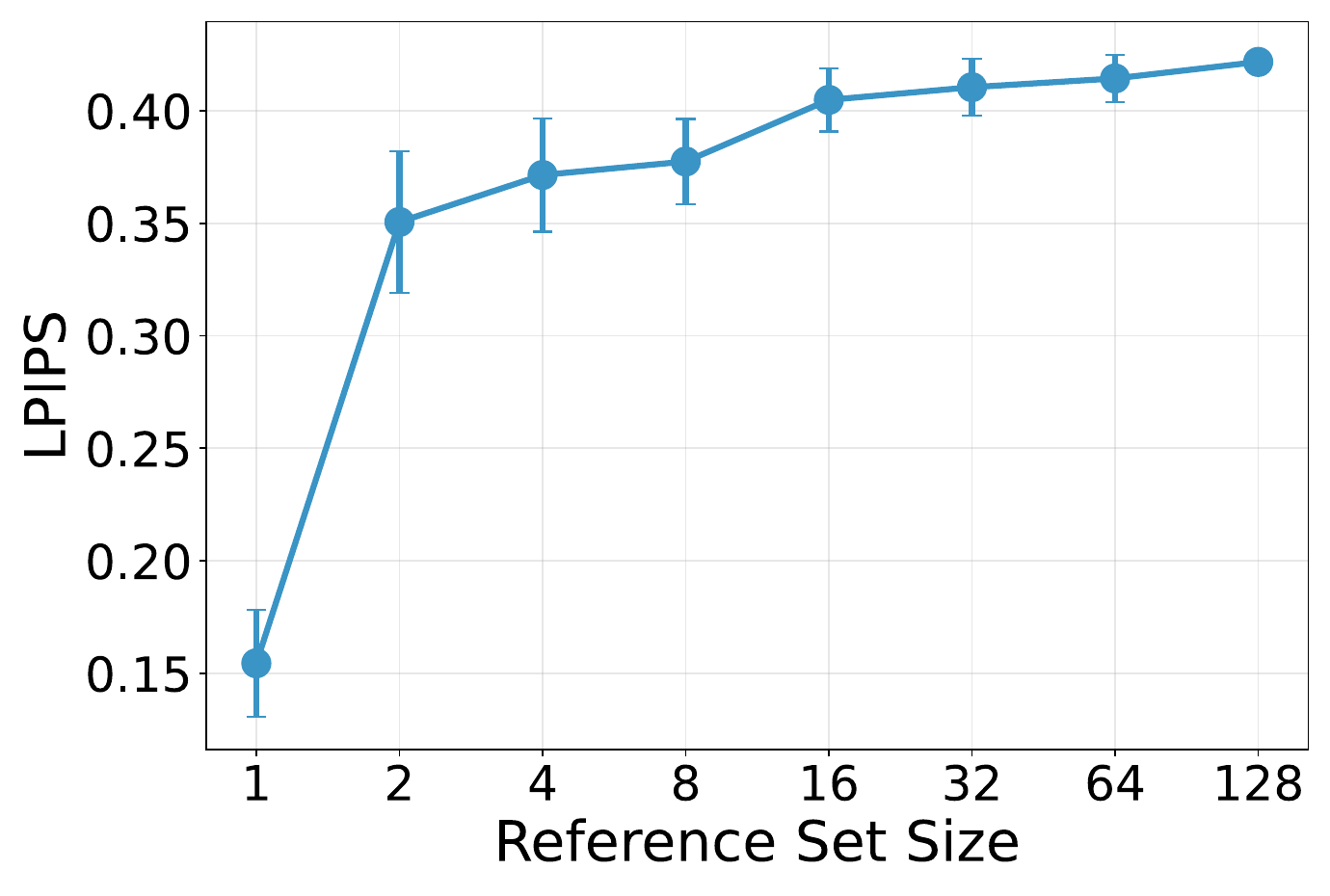}
\caption{\textbf{Diversity grows with the number of references.} LPIPS
diversity against reference-set size $M$. Error bars show the spread across
randomly sampled reference subsets.}
\label{fig:dataset_size_ablation_lpips}
\end{figure}

Diversity grows with the size of the reference set
(\Cref{fig:dataset_size_ablation_lpips}). The largest jump occurs between
one and two references, and diversity keeps rising more slowly up to
$M=128$, even though every subset represents a single semantic attribute.
Reference-mean guidance therefore does not pull generation toward a single
target but spreads it across the reference distribution. This comparison is
internal to RMG, and relative to the unguided model RMG still reduces
within-prompt diversity (\Cref{tab:geneval_tradeoffs}).

\subsection{Reference-Bank Quality}
\label{app:bank_quality}

To measure how sensitive RMG is to the quality of its bank, we fix the
prompt \textit{``an animal in a savanna''} with zebra as the target and
generate 50 samples under six conditions: an unguided control and five
20-image banks. The clean bank contains 20 zebras, and the mixed bank 11
zebras and 9 other images. The corrupted bank contains degraded versions of
the 20 clean zebra images. Of the two irrelevant banks, the first contains
other animals, two of them zebras, and the second contains only household
objects. All conditions use 50 sampling steps (rather than the 20 of the
GenEval protocol), quadratic decay with $\gamma_0=1.0$, and top-$k$
selection with $k=20$, which for $M=20$ keeps the whole bank.

\paragraph{Corruption.}
The corrupted bank applies, in order, bilinear downsampling to
$64\times64$, bilinear upsampling to the original resolution, a Gaussian
blur of radius 2, and JPEG compression at quality 25. All six conditions
share the same 50 noise seeds.

\paragraph{Metrics.}
Zebra success is judged by
Qwen2-VL-7B-Instruct~\citep{Qwen2-VL} with the question
\textit{``Is the main animal a zebra? Answer with exactly one word: yes or
no.''} The directional CLIP margin follows \Cref{app:clip_attribute_score},
using CLIP ViT-L/14~\citep{radford2021learning} with the positive prompt
\textit{``a zebra in a savanna''} and the negative prompt \textit{``a
non-zebra animal in a savanna''}. LPIPS follows
\Cref{app:geneval_evaluation_protocol} and is averaged over all
$\binom{50}{2}=1{,}225$ unordered pairs within each condition.

\begin{table}[!htbp]
\centering
\small
\caption{\textbf{Relevant references steer even as a minority, and
degraded references reduce diversity.} Bank quality for \textit{``an animal
in a savanna''} with zebra as the target, 50 outputs per condition. Each row uses
one fixed bank rather than repeated random draws. LPIPS measures variation
within a condition, not image quality.}
\label{tab:bank_quality}
\setlength{\tabcolsep}{4pt}
\begin{tabular}{lcccc}
\toprule
Bank & On-target refs. & Zebra success $\uparrow$ & CLIP margin $\uparrow$ & LPIPS diversity \\
\midrule
No bank (unguided) & -- & 0\% & $-0.026$ & 0.592 \\
Clean (zebras) & 20/20 & 100\% & +0.006 & 0.426 \\
Mixed & 11/20 & 70\% & $-0.003$ & 0.487 \\
Corrupted zebras & 20/20 & 62\% & $-0.011$ & 0.241 \\
Irrelevant (other animals) & 2/20 & 24\% & $-0.021$ & 0.526 \\
Irrelevant (household objects) & 0/20 & 0\% & $-0.024$ & 0.477 \\
\bottomrule
\end{tabular}
\end{table}

With the clean bank, zebra success rises from 0/50 to 50/50
(\Cref{tab:bank_quality}). The mixed bank still reaches 35/50, and the bank
with only two zebras reaches 12/50, so useful references are not simply
diluted in proportion to their share of the bank. Corruption acts
differently from irrelevance. The corrupted bank keeps 31/50 successes, but
its LPIPS falls to 0.241, against 0.426 for the clean bank, so the semantic
signal survives while the low-variance statistics of the degraded images are
transferred along with it. The household-object bank produces no zebras,
like the unguided model, yet it also lowers LPIPS, so even an irrelevant
bank is not inert. Since the study uses one prompt and one fixed bank per
condition, it characterizes these particular banks rather than robustness in
general.

\subsection{Number of Function Evaluations (NFE)}
\label{app:nfe_ablation}

Finally, we vary the number of function evaluations on a harder pose-control
task, a gymnastics ring leap, with the prompt
\begin{quote}
\small
\textit{``a gymnast performing a ring leap, full body visible, airborne, one
leg extended forward, the back leg bent high behind the head, arched back,
pointed toes, arms extended, dynamic sports photograph''}
\end{quote}
and a fixed reference set of ring-leap images. We sweep
$\mathrm{NFE} \in \{10, 20, 30, 50, 100, 200\}$ and
$\gamma_0 \in \{0.1, 0.2, 0.4, 0.5, 1.0\}$ under the quadratic schedule,
holding the prompt, reference set, and seeds fixed.

\begin{figure}[!htbp]
\centering
\setlength{\tabcolsep}{1.5pt}
\renewcommand{\arraystretch}{0.8}
\begin{tabular}{@{}c@{\hspace{3pt}}cccccc@{}}
& \small\textbf{NFE 10} & \small\textbf{NFE 20} & \small\textbf{NFE 30} & \small\textbf{NFE 50} & \small\textbf{NFE 100} & \small\textbf{NFE 200} \\
\rotatebox{90}{\small\hspace{0.6em}Unguided}
& \includegraphics[width=0.145\linewidth]{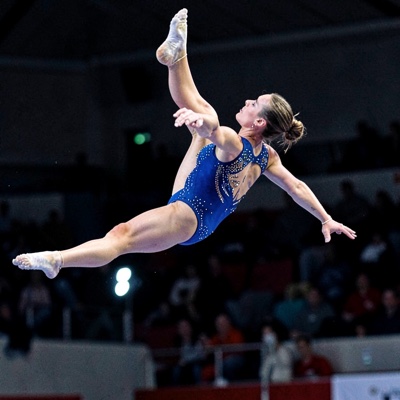} & \includegraphics[width=0.145\linewidth]{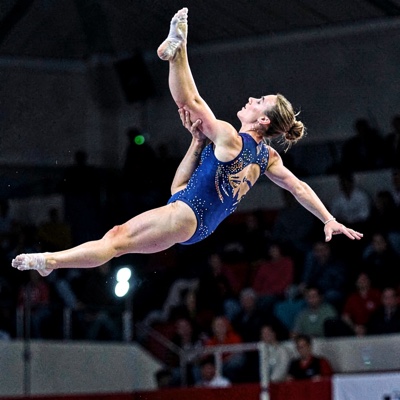}
& \includegraphics[width=0.145\linewidth]{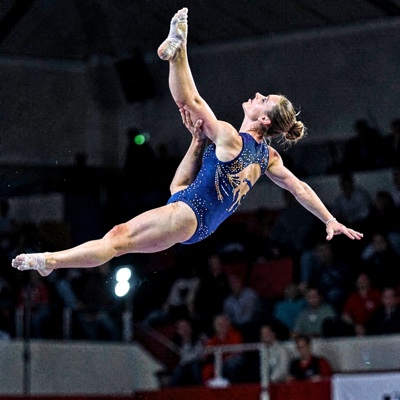} & \includegraphics[width=0.145\linewidth]{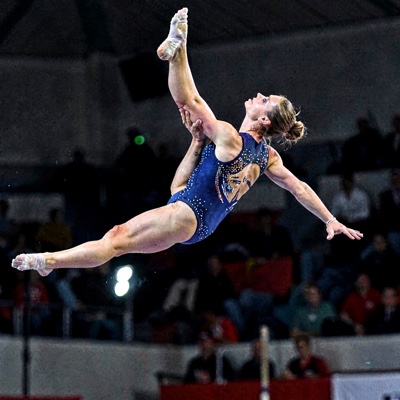}
& \includegraphics[width=0.145\linewidth]{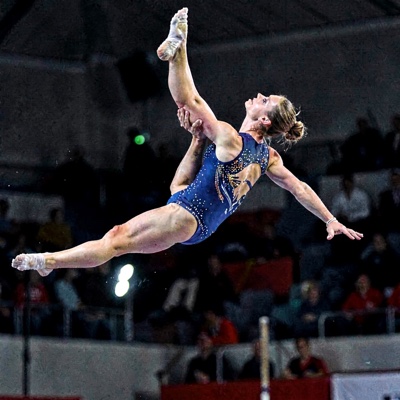} & \includegraphics[width=0.145\linewidth]{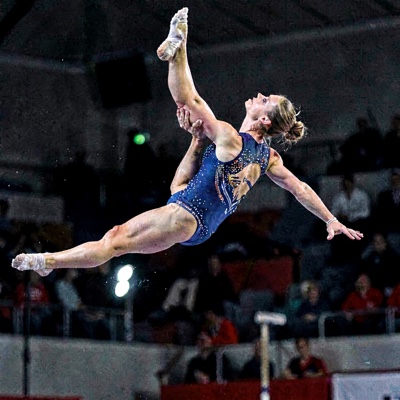} \\[2pt]
\rotatebox{90}{\small\hspace{0.4em}$\gamma_0=0.1$} %
& \includegraphics[width=0.145\linewidth]{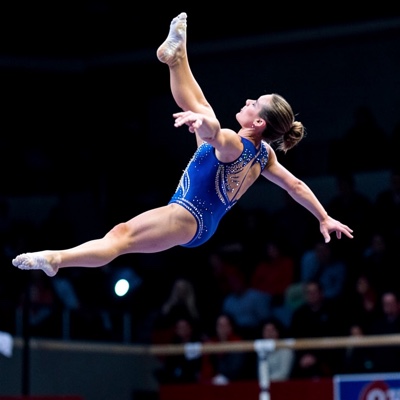} & \includegraphics[width=0.145\linewidth]{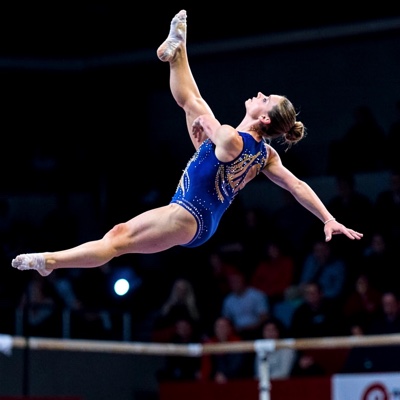}
& \includegraphics[width=0.145\linewidth]{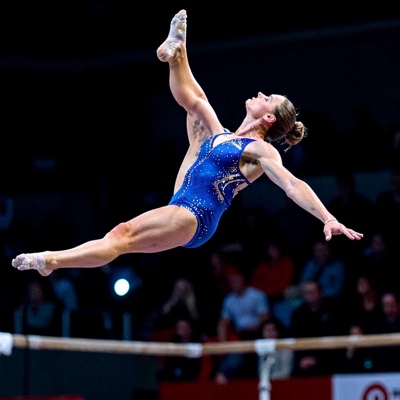} & \includegraphics[width=0.145\linewidth]{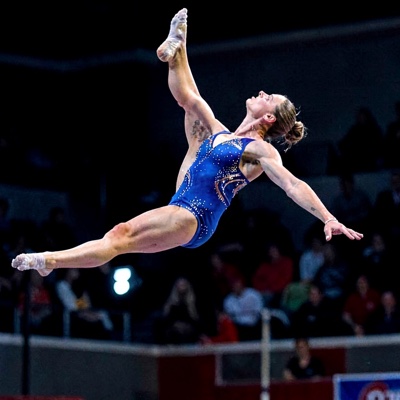}
& \includegraphics[width=0.145\linewidth]{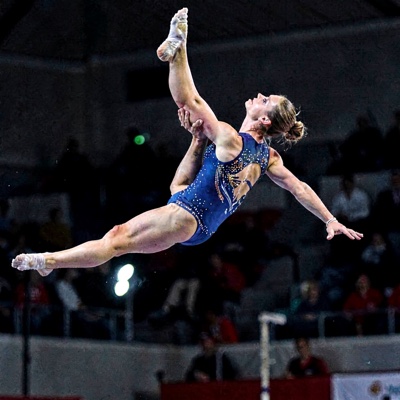} & \includegraphics[width=0.145\linewidth]{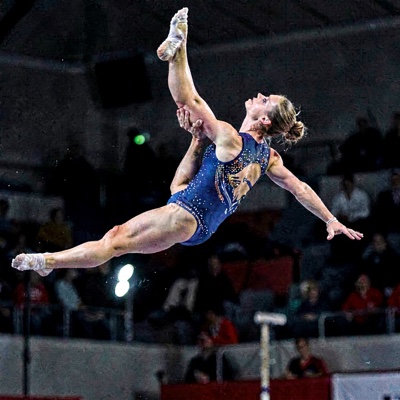} \\
\rotatebox{90}{\small\hspace{0.4em}$\gamma_0=0.2$} %
& \includegraphics[width=0.145\linewidth]{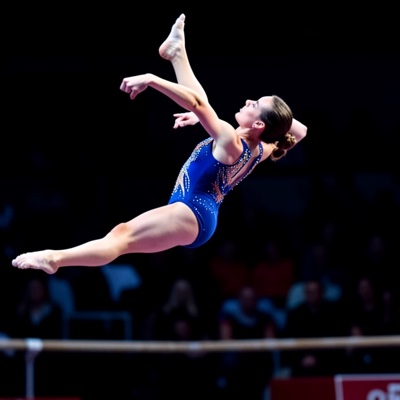} & \includegraphics[width=0.145\linewidth]{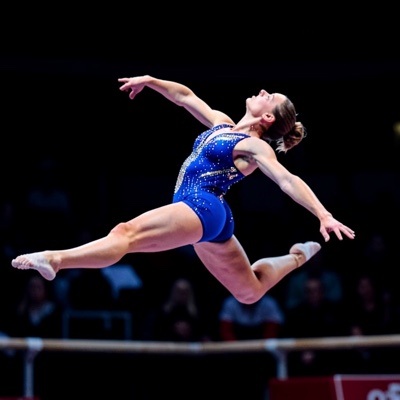}
& \includegraphics[width=0.145\linewidth]{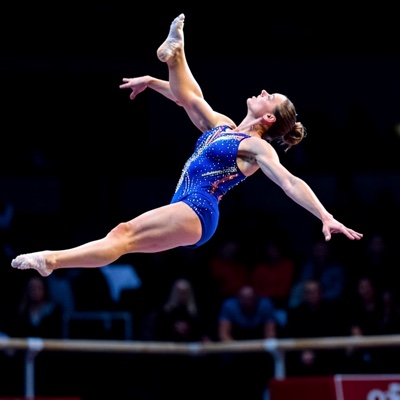} & \includegraphics[width=0.145\linewidth]{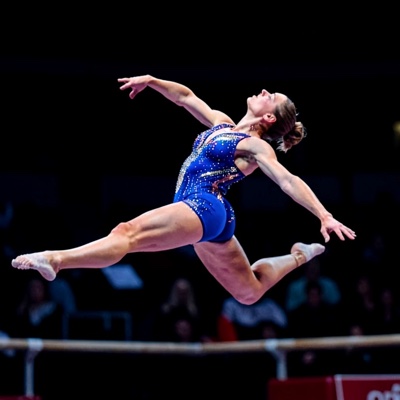}
& \includegraphics[width=0.145\linewidth]{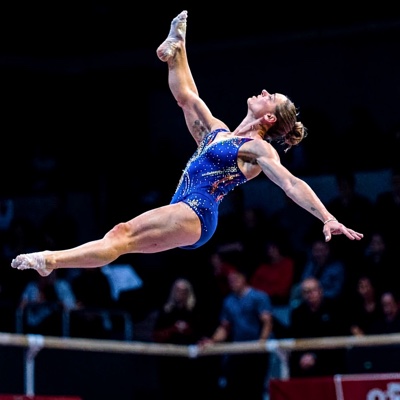} & \includegraphics[width=0.145\linewidth]{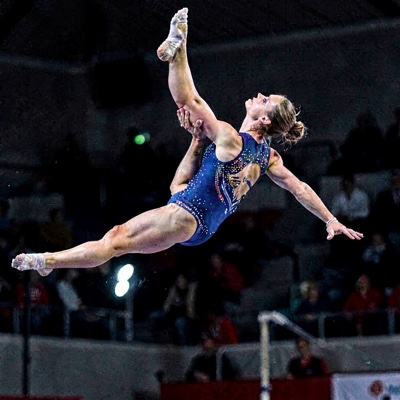} \\
\rotatebox{90}{\small\hspace{0.4em}$\gamma_0=0.4$} %
& \includegraphics[width=0.145\linewidth]{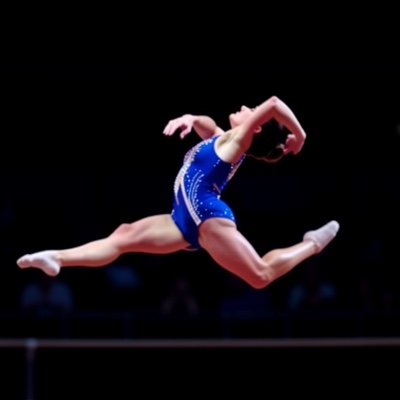} & \includegraphics[width=0.145\linewidth]{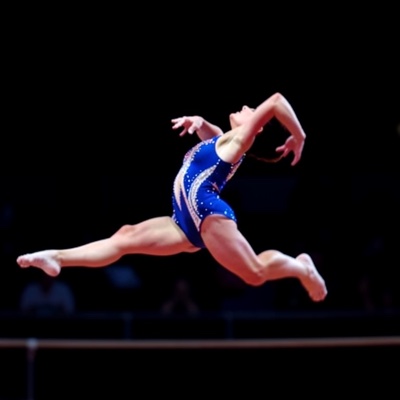}
& \includegraphics[width=0.145\linewidth]{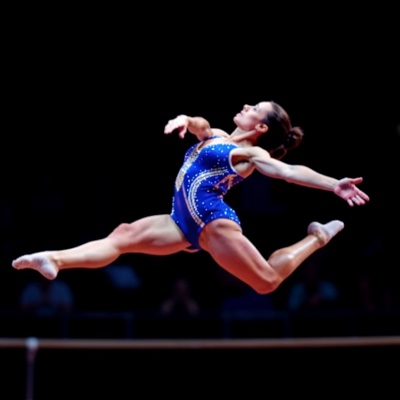} & \includegraphics[width=0.145\linewidth]{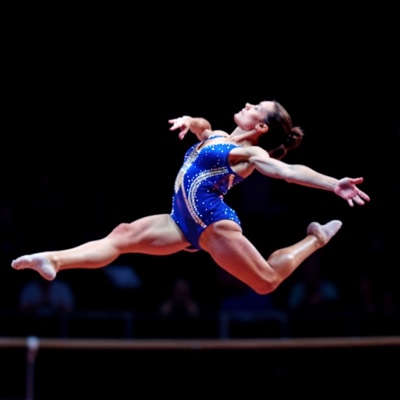}
& \includegraphics[width=0.145\linewidth]{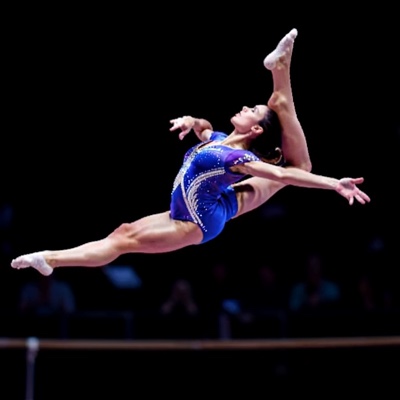} & \includegraphics[width=0.145\linewidth]{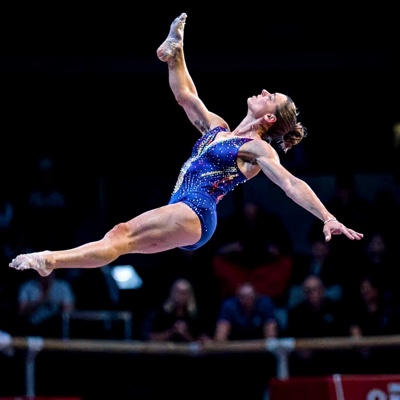} \\
\rotatebox{90}{\small\hspace{0.4em}$\gamma_0=0.5$} %
& \includegraphics[width=0.145\linewidth]{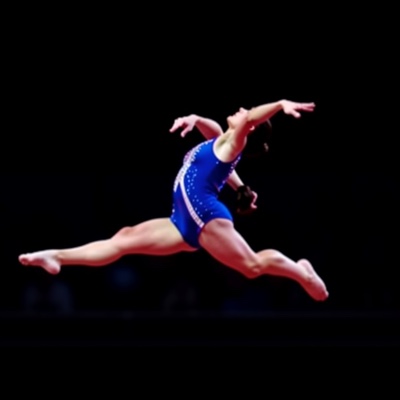} & \includegraphics[width=0.145\linewidth]{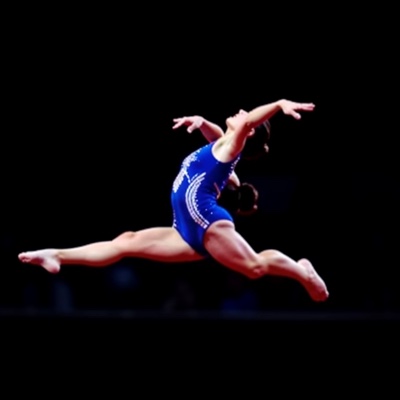}
& \includegraphics[width=0.145\linewidth]{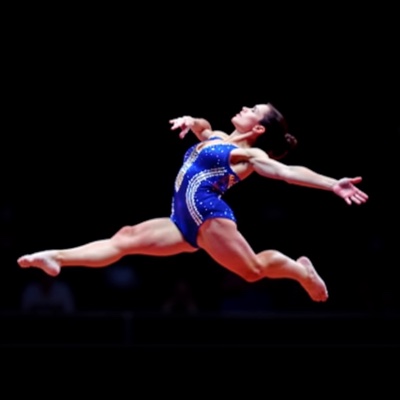} & \includegraphics[width=0.145\linewidth]{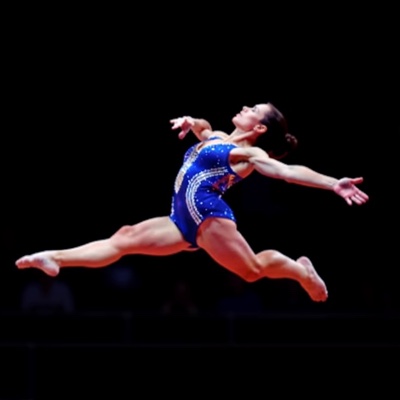}
& \includegraphics[width=0.145\linewidth]{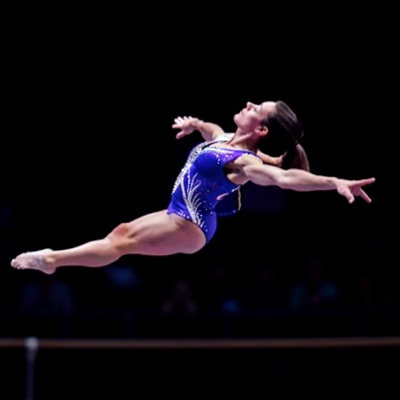} & \includegraphics[width=0.145\linewidth]{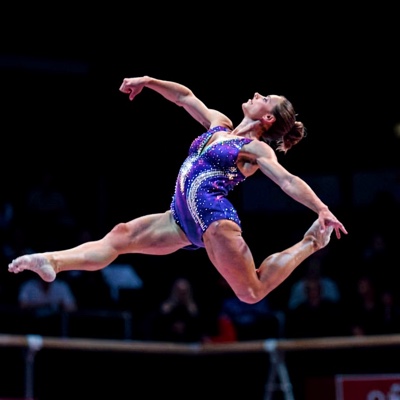} \\
\rotatebox{90}{\small\hspace{0.4em}$\gamma_0=1.0$} %
& \includegraphics[width=0.145\linewidth]{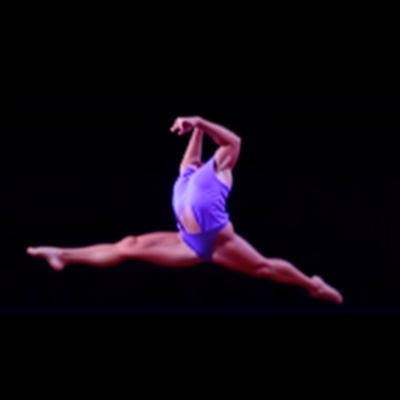} & \includegraphics[width=0.145\linewidth]{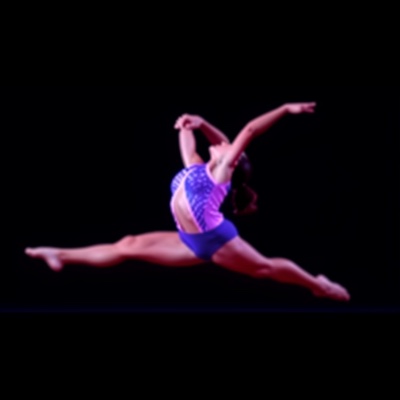}
& \includegraphics[width=0.145\linewidth]{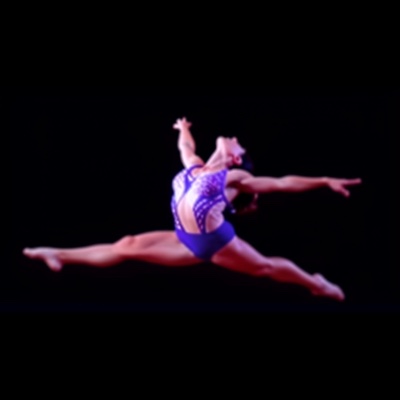} & \includegraphics[width=0.145\linewidth]{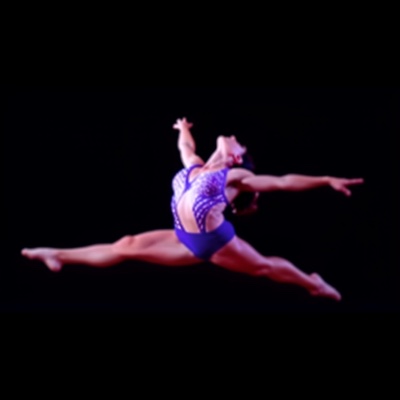}
& \includegraphics[width=0.145\linewidth]{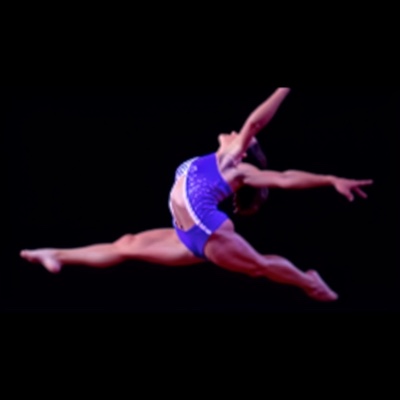} & \includegraphics[width=0.145\linewidth]{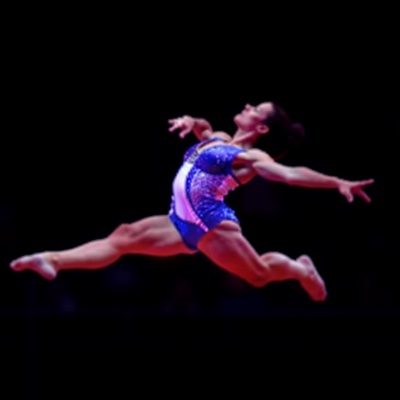} \\
\end{tabular}
\caption{\textbf{Pose control holds across solver budgets once guidance is
strong enough.} Ring-leap prompt across solver budgets and guidance
strengths. The top row is the unguided model, and each following row applies
RMG with the indicated $\gamma_0$. Columns vary the number of function
evaluations.}
\label{fig:nfe_ablation_ring_leap_grid}
\end{figure}

\Cref{fig:nfe_ablation_ring_leap_grid} shows the results. The unguided
model produces nearly the same pose at every budget. At $\gamma_0=0.1$
guidance leaves the image essentially unchanged, while from $\gamma_0=0.2$
onward the guided samples move away from the unguided pose toward the leap
of the references. At $\gamma_0=0.2$ this effect fades at the largest
budgets. At $\gamma_0=1.0$ it holds at every budget, but with a strong
color shift. The appropriate strength therefore depends somewhat on the
solver budget. Runtime grows approximately linearly with NFE.

\section{Extended Discussion of Limitations}
\label{app:limitations}

This appendix expands on the limitations summarized in \Cref{sec:limitations}.

\paragraph{Dependence on the reference bank.}
RMG inherits the quality of its reference set (\Cref{app:bank_quality}) and
transfers whatever the references have in common, including attributes the
user did not intend, such as the light backgrounds in
\Cref{fig:flux_bank_swap} (\Cref{app:spg_background_nocopy}). It needs
neither an adapter nor a reward model, but it does need a reasonably chosen
bank. All of our banks are curated, and we have not yet built banks by
retrieval or measured how results vary across repeated bank draws. Computing
posterior means over very large banks also becomes costly, although
subsampling and approximate nearest-neighbor retrieval offer practical
remedies.

\paragraph{Diversity and quality.}
On GenEval, RMG trades some within-prompt diversity and ImageReward for
alignment (\Cref{tab:geneval_tradeoffs}). The ImageReward decrease is
confined to position and counting (\Cref{tab:geneval_quality_categories}),
and we attribute it to the white borders of their tiled banks, although we
have not confirmed this by regenerating the banks without borders.
Border-free compositing, natural layout references, and weaker or adaptive
guidance are the mitigations we would try first.

\paragraph{Proximity to the references.}
No RMG output exceeds an SSCD similarity of 0.85 to its bank
(\Cref{app:geneval_tradeoffs}), but this is a threshold test with a single
detector at the default guidance setting rather than a guarantee. Schedules
that stay strong late in the trajectory bring outputs much closer to
individual references (\Cref{app:proof_thm3}).

\paragraph{Scope of the comparisons and of the theory.}
We have not compared RMG with reference-conditioned methods such as
IP-Adapter or ControlNet given the same images, which would complement the
text-only baselines of GenEval. RMG needs only a frozen VAE and an affine
interpolant, so it should apply to other latent rectified-flow models, but
we have not verified this beyond \mbox{FLUX.2-klein}. On the theoretical
side, the practical schedule and tempered weights depart from exact mixture
transport (\Cref{app:proof_thm3,app:reference_sets}), so the guidance rule is
motivated by the theory rather than derived exactly from it.

\section{Reference Banks}
\label{app:flux_banks}

\begin{figure}[H]
\centering
\begin{minipage}[t]{0.48\linewidth}
\centering
\includegraphics[width=\linewidth]{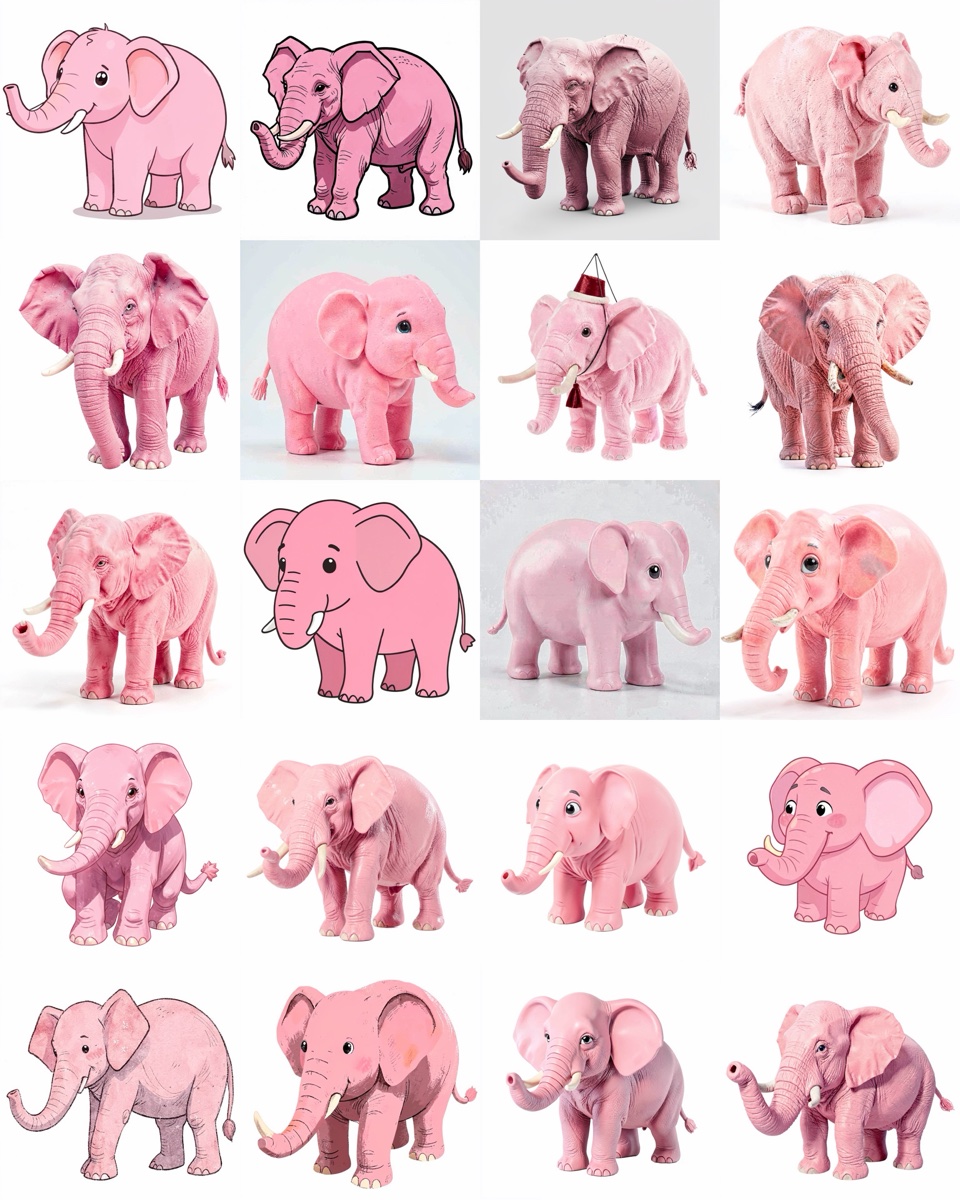}
\caption{\textbf{Pink elephants.}}
\label{fig:flux_bank00}
\end{minipage}\hfill
\begin{minipage}[t]{0.48\linewidth}
\centering
\includegraphics[width=\linewidth]{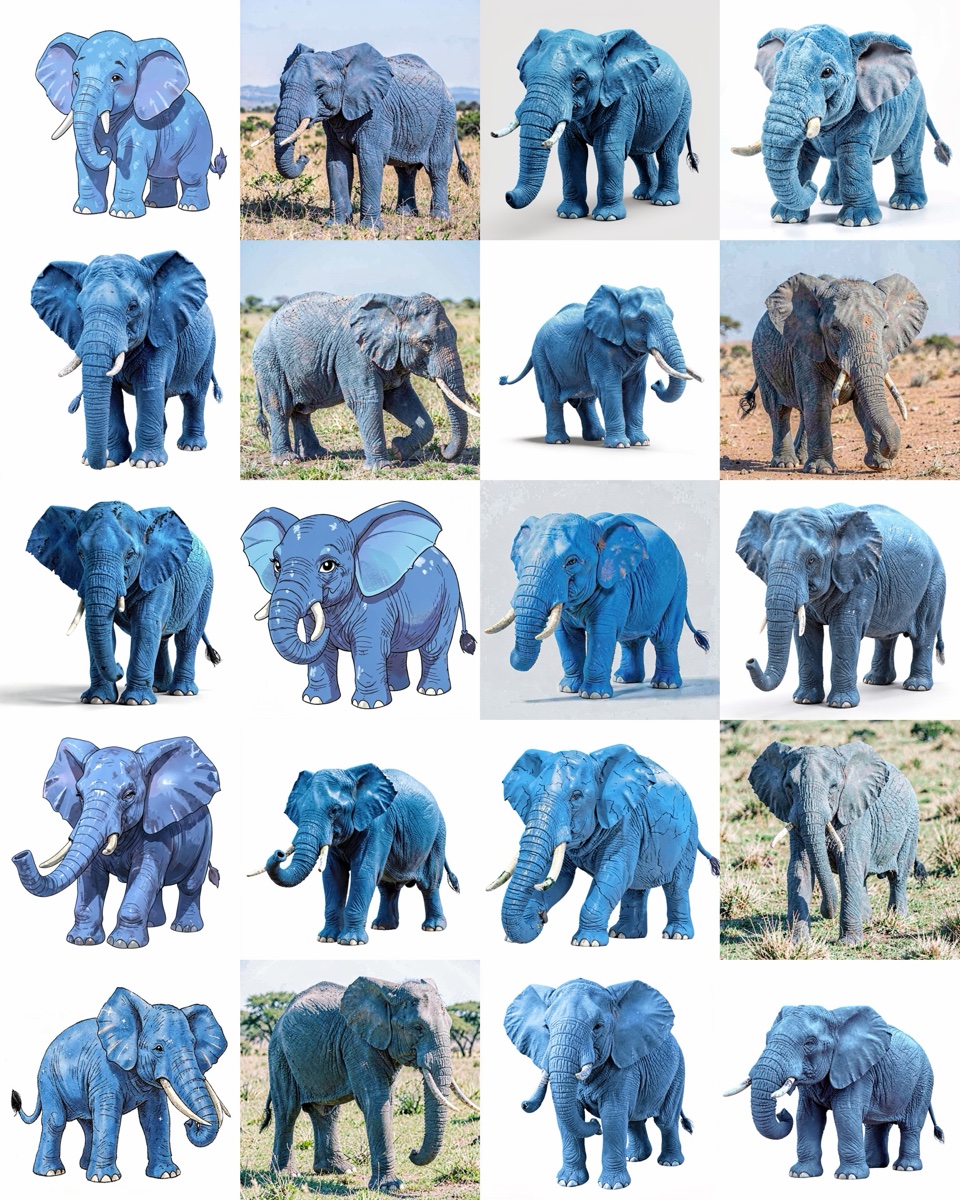}
\caption{\textbf{Blue elephants.}}
\label{fig:flux_bank01}
\end{minipage}
\end{figure}

\begin{figure}[H]
\centering
\begin{minipage}[t]{0.48\linewidth}
\centering
\includegraphics[width=\linewidth]{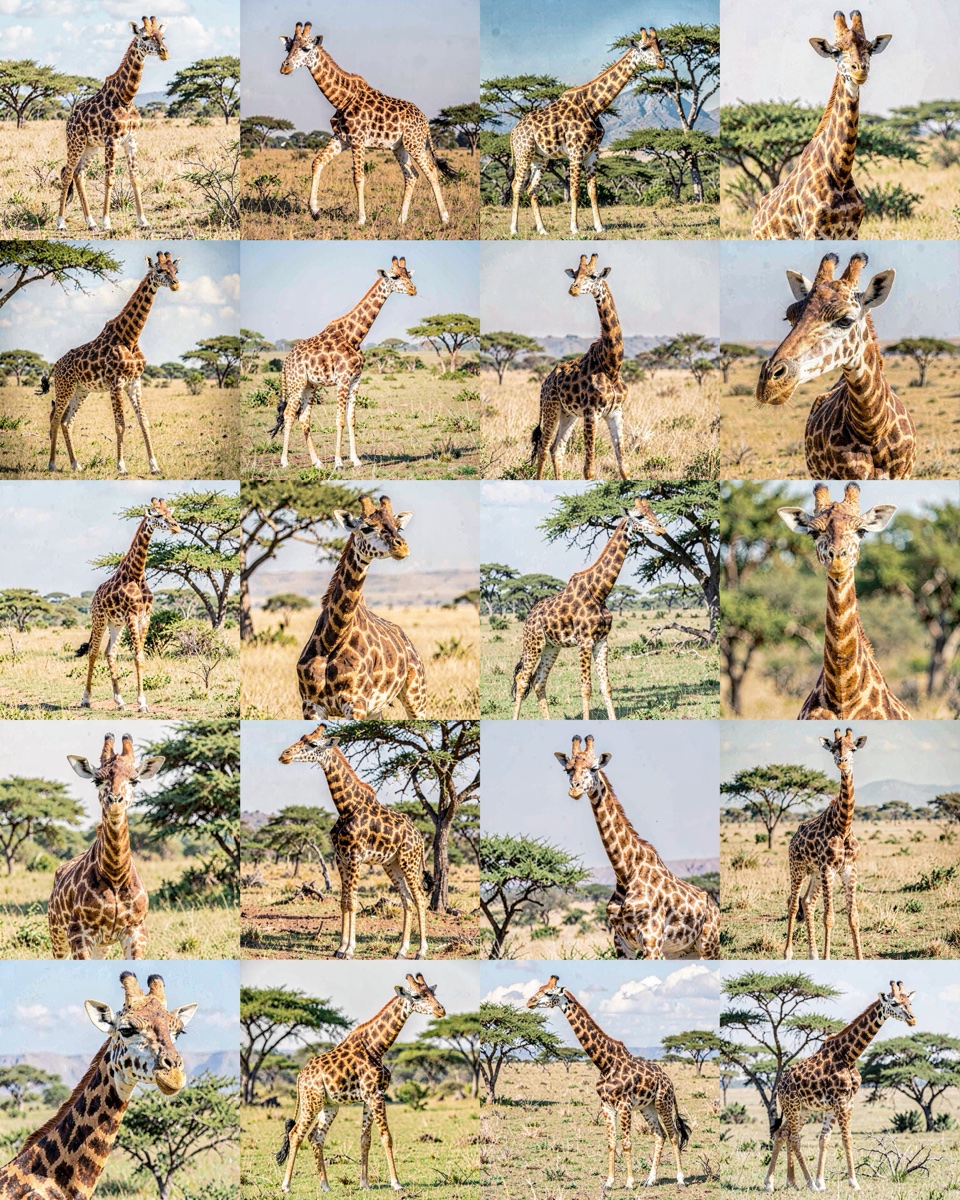}
\caption{\textbf{Giraffes.}}
\label{fig:flux_bank02}
\end{minipage}\hfill
\begin{minipage}[t]{0.48\linewidth}
\centering
\includegraphics[width=\linewidth]{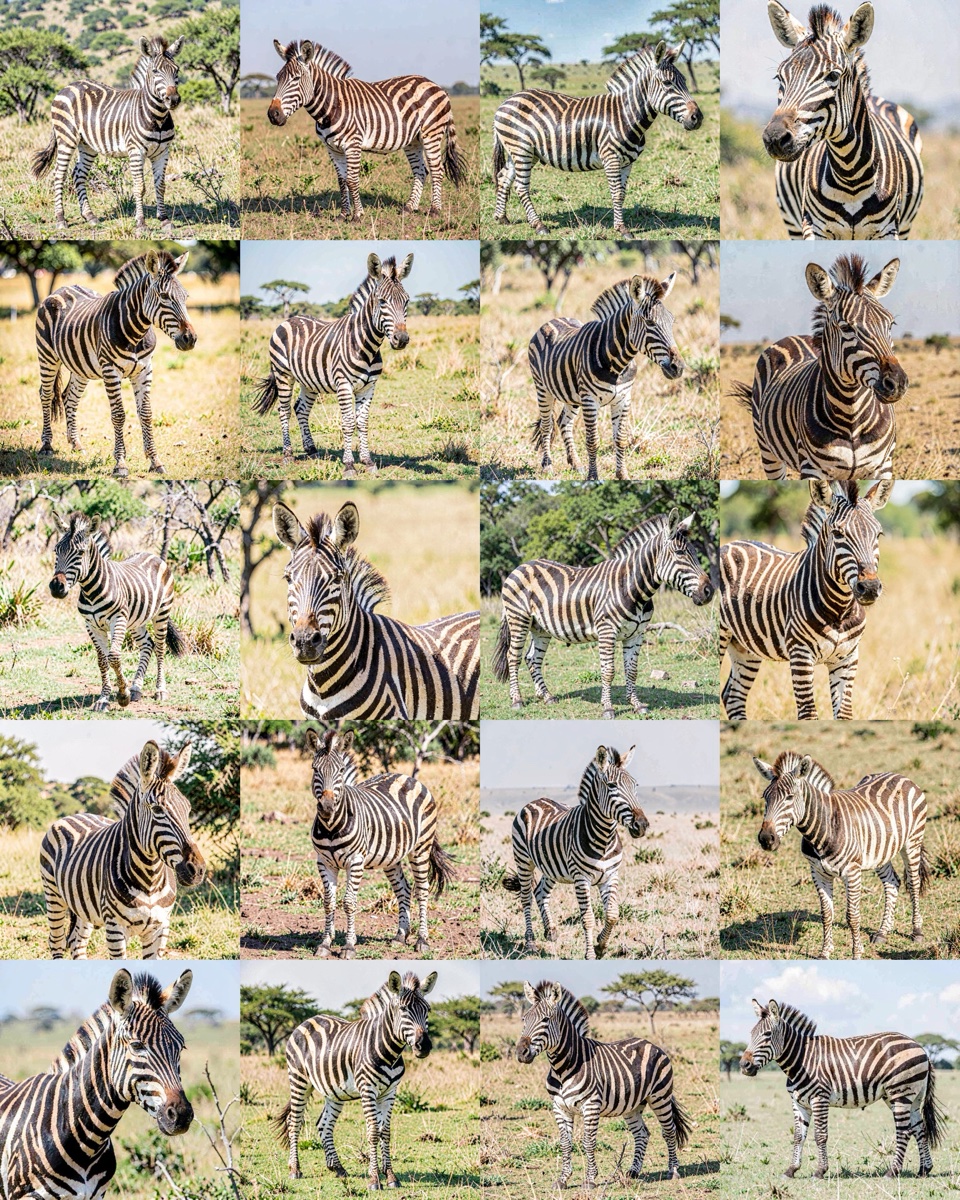}
\caption{\textbf{Zebras.}}
\label{fig:flux_bank03}
\end{minipage}
\end{figure}

\begin{figure}[H]
\centering
\begin{minipage}[t]{0.48\linewidth}
\centering
\includegraphics[width=\linewidth]{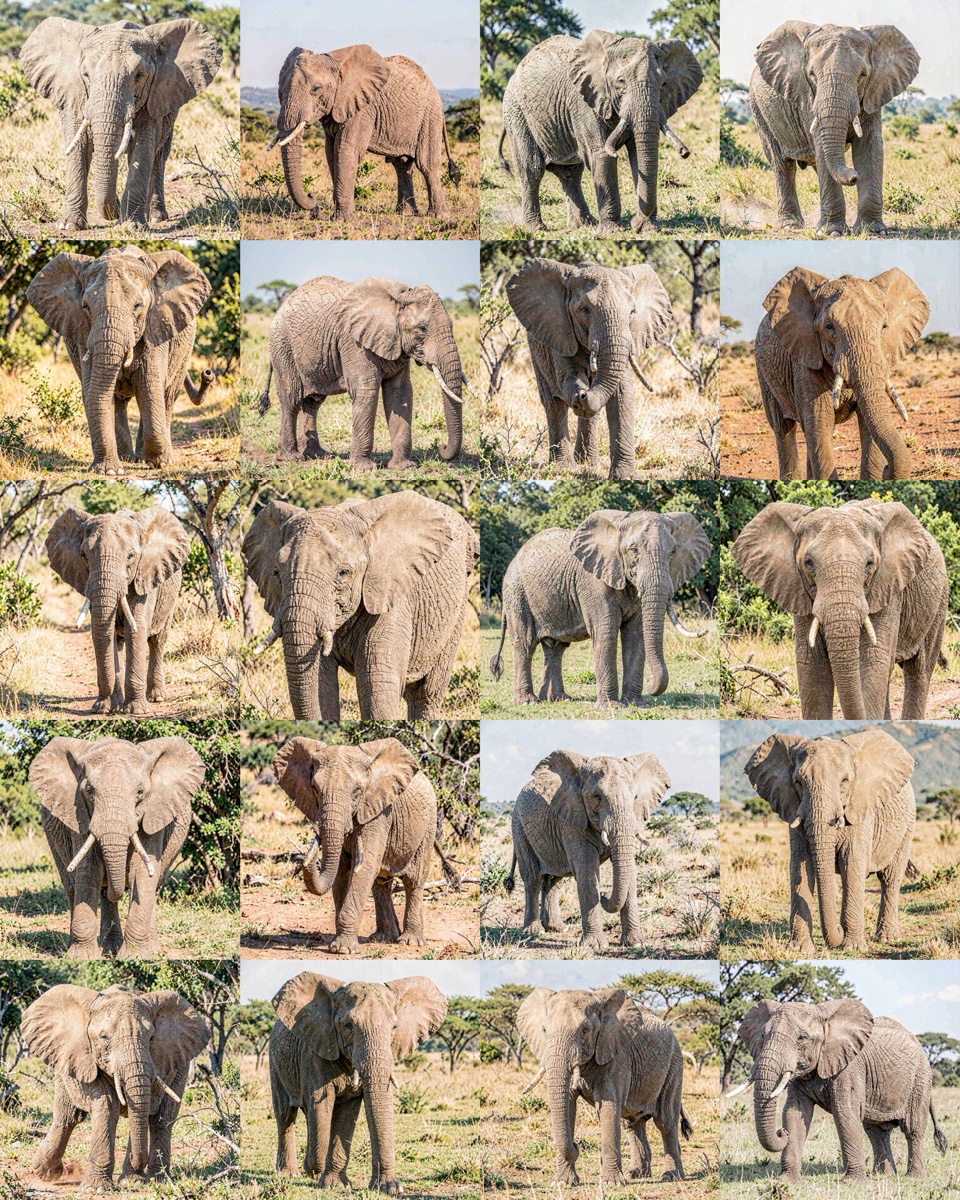}
\caption{\textbf{Elephants.}}
\label{fig:flux_bank04}
\end{minipage}\hfill
\begin{minipage}[t]{0.48\linewidth}
\centering
\includegraphics[width=\linewidth]{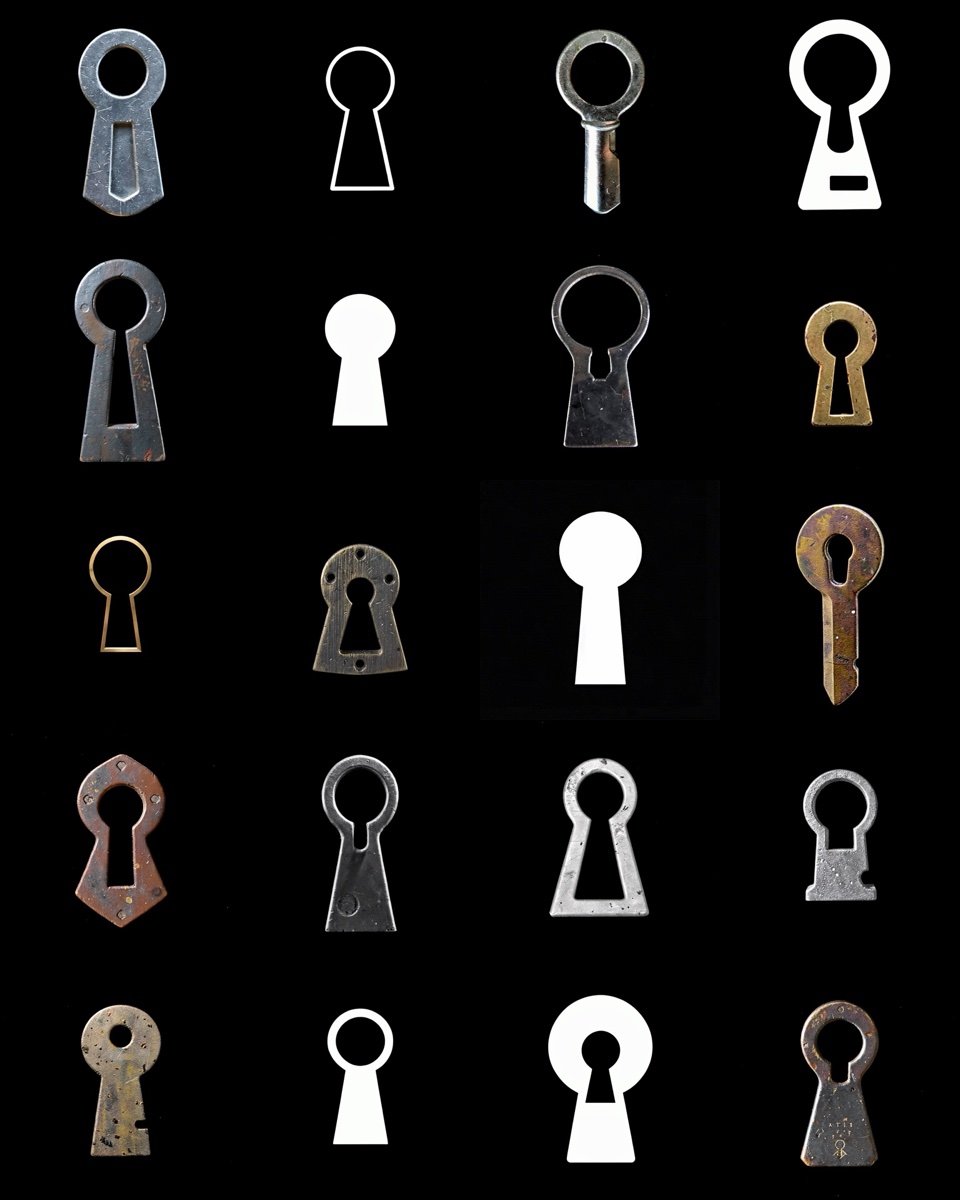}
\caption{\textbf{Keyholes.}}
\label{fig:flux_bank05}
\end{minipage}
\end{figure}

\begin{figure}[H]
\centering
\begin{minipage}[t]{0.48\linewidth}
\centering
\includegraphics[width=\linewidth]{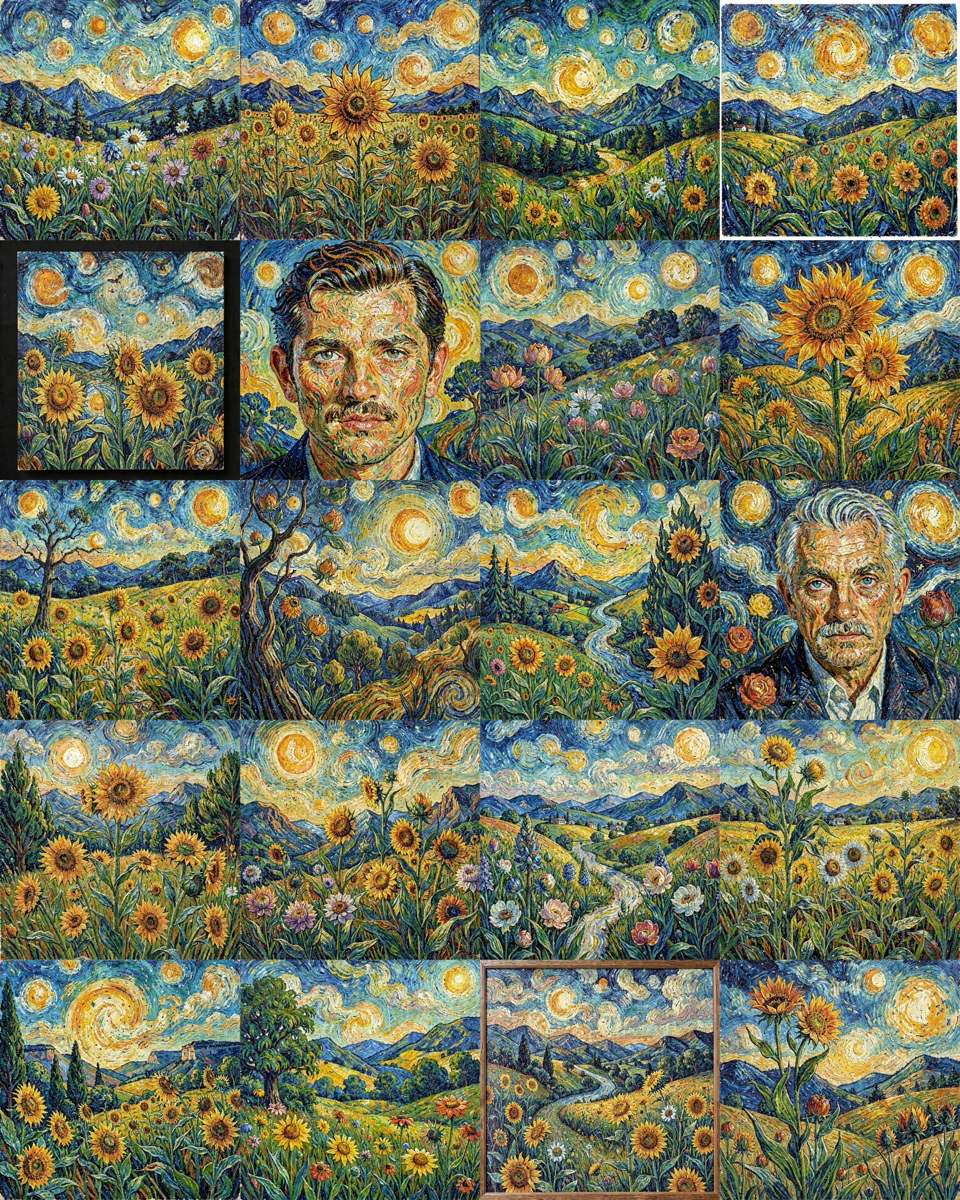}
\caption{\textbf{Van Gogh-style paintings.}}
\label{fig:flux_bank06}
\end{minipage}\hfill
\begin{minipage}[t]{0.48\linewidth}
\centering
\includegraphics[width=\linewidth]{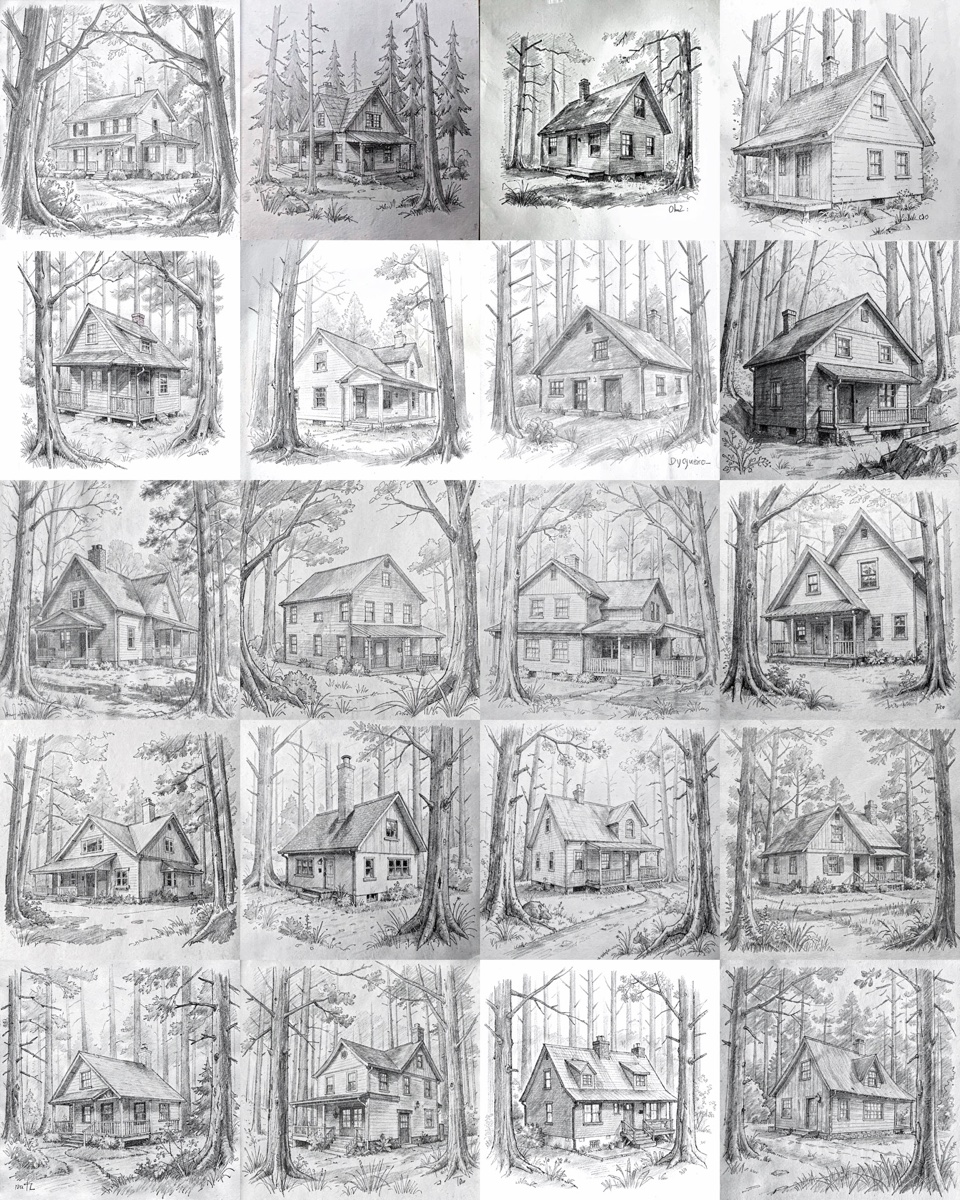}
\caption{\textbf{Pencil sketches of houses.}}
\label{fig:flux_bank07}
\end{minipage}
\end{figure}

\begin{figure}[H]
\centering
\begin{minipage}[t]{0.48\linewidth}
\centering
\includegraphics[width=\linewidth]{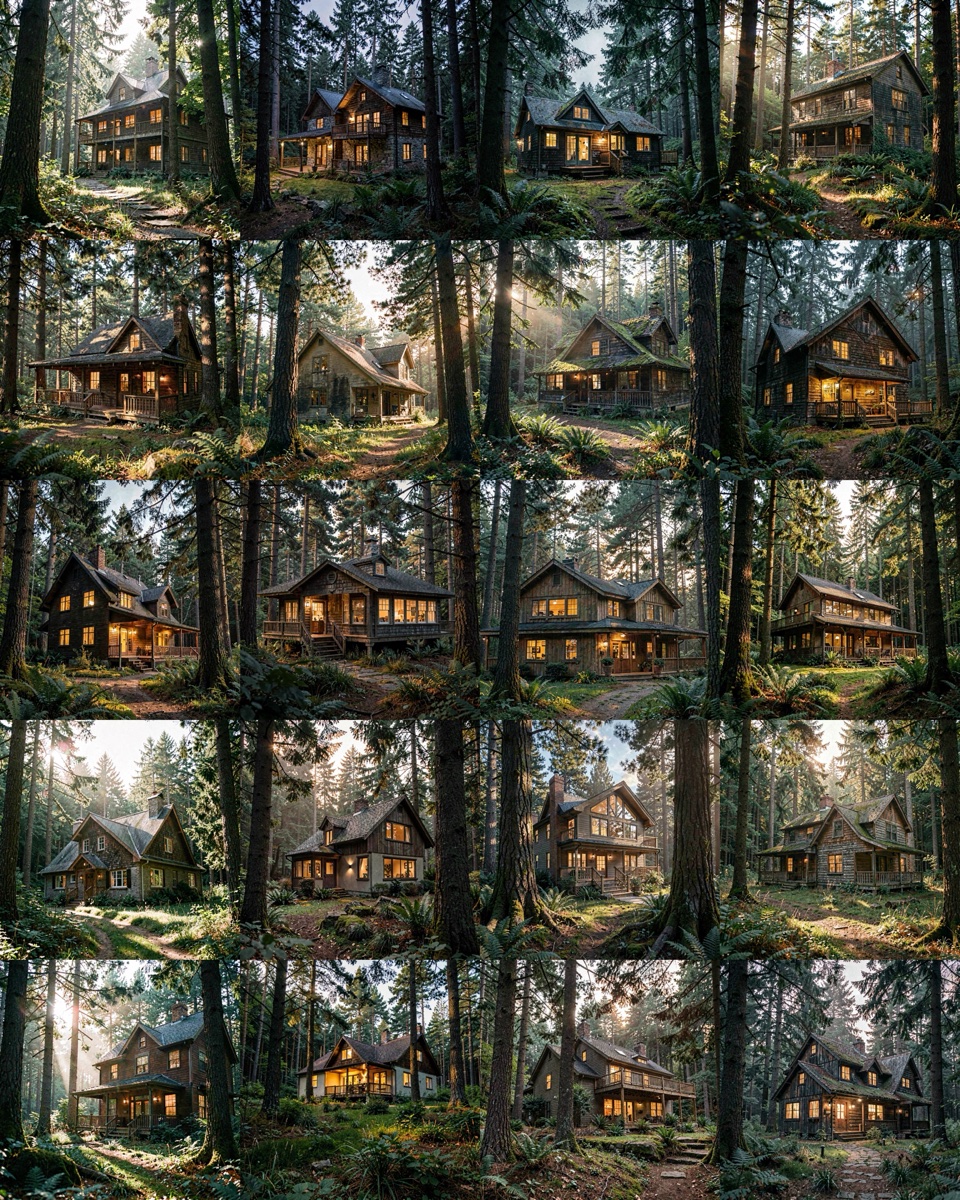}
\caption{\textbf{Cinematic photographs of houses.}}
\label{fig:flux_bank08}
\end{minipage}\hfill
\begin{minipage}[t]{0.48\linewidth}
\centering
\includegraphics[width=\linewidth]{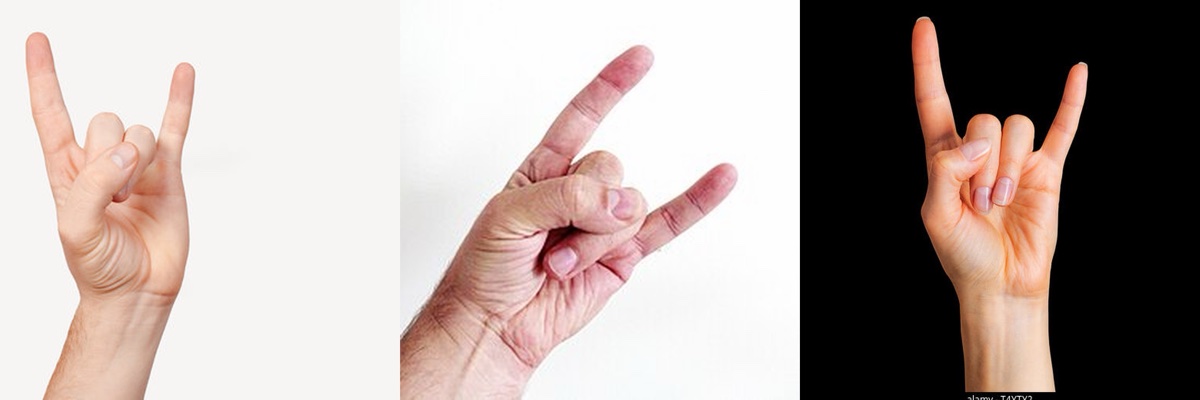}
\caption{\textbf{Hand poses for the sign-of-the-horns experiment.}}
\label{fig:flux_bank09}
\end{minipage}
\end{figure}

\Cref{fig:flux_bank00,fig:flux_bank01,fig:flux_bank02,fig:flux_bank03,fig:flux_bank04,fig:flux_bank05,fig:flux_bank06,fig:flux_bank07,fig:flux_bank08,fig:flux_bank09}
show the reference banks used in the text-to-image experiments. Each bank contains
20 images, except for the hand-pose bank, which contains three.


\FloatBarrier

\end{document}